\documentclass{article}

\usepackage[nonatbib,preprint]{neurips_data_2021}
\usepackage[backend=biber, sorting=none, sortcites=true]{biblatex}
\bibliography{egbib}

\usepackage[utf8]{inputenc} 
\usepackage[T1]{fontenc}    
\usepackage[colorlinks]{hyperref}
\usepackage{url}            
\usepackage{booktabs}       
\usepackage{amsfonts}       
\usepackage{nicefrac}       
\usepackage{microtype}      
\usepackage[dvipsnames]{xcolor, colortbl}        
\usepackage{epsfig}
\usepackage{graphicx}
\usepackage{calc}
\usepackage{amsmath}
\usepackage{multicol, booktabs}
\usepackage{pifont}
\newcommand{\cmark}{\ding{51}}
\newcommand{\xmark}{\ding{55}}
\usepackage{tikz}
\DeclareMathOperator*{\argmax}{arg\,max}
\usepackage{enumitem}
\usepackage{comment}
\usepackage{subfig}
\usepackage[nameinlink,noabbrev]{cleveref}

\usepackage{xspace}
\makeatletter
\DeclareRobustCommand\onedot{\futurelet\@let@token\@onedot}
\def\@onedot{\ifx\@let@token.\else.\null\fi\xspace}

\def\eg{\emph{e.g}\onedot} 
\def\ie{\emph{i.e}\onedot} 
\def\cf{\emph{c.f}\onedot} 
 \def\vs{\emph{vs}\onedot}
\def\wrt{w.r.t\onedot} 

\makeatother

\definecolor{Gray}{gray}{0.9}
\definecolor{DarkGray}{gray}{0.8}
\newcolumntype{a}{>{\columncolor{DarkGray}}c}
\newcolumntype{g}{>{\columncolor{Gray}}c}

\title{SegmentMeIfYouCan:\\A Benchmark for Anomaly Segmentation}

\author{%
Robin\,Chan\thanks{equal contribution} $\,^{1}\!$%
\And
Krzysztof\,Lis\footnotemark[1]$\,\,^{2}$%
\And
Svenja\,Uhlemeyer\footnotemark[1] $\,^{1}\!$%
\And
Hermann\,Blum\footnotemark[1]$\,\,^{3}$%
\And
Sina\,Honari$^{2}$%
\And
Roland\,Siegwart$^{3}$%
\And
Pascal\,Fua$^{2}$%
\And
Mathieu\,Salzmann$^{2}$%
\And
Matthias\,Rottmann$^{1}$%
}

\begin{document}

\maketitle
\footnotetext[1]{Stochastics Group, IZMD, University of Wuppertal, Wuppertal, Germany}
\footnotetext[2]{Computer Vision Laboratory, EPFL, Lausanne, Switzerland}
\footnotetext[3]{Autonomous Systems Lab, ETHZ, Zürich, Switzerland}

\begin{abstract}
State-of-the-art semantic or instance segmentation deep neural networks (DNNs) are usually trained on a closed set of semantic classes. As such, they are ill-equipped to handle previously-unseen objects. 
However, detecting and localizing such objects is crucial for safety-critical applications such as perception for automated driving, especially if they appear on the road ahead. While some methods have tackled the tasks of anomalous or out-of-distribution object segmentation, progress remains slow, in large part due to the lack of solid benchmarks; existing datasets either consist of synthetic data, or suffer from label inconsistencies. In this paper, we bridge this gap by introducing the ``SegmentMeIfYouCan'' benchmark. Our benchmark addresses two tasks: Anomalous object segmentation, which considers any previously-unseen object category; and road obstacle segmentation, which focuses on any object on the road, may it be known or unknown.
We provide two corresponding datasets together with a test suite performing an in-depth method analysis, considering both established pixel-wise performance metrics and recent component-wise ones, which are insensitive to object sizes. We empirically evaluate multiple state-of-the-art baseline methods, including several models specifically designed for anomaly / obstacle segmentation, on our datasets and on public ones, using our test suite.
The anomaly and obstacle segmentation results show that our datasets contribute to the diversity and difficulty of both data landscapes.
\end{abstract}


\section{Introduction}

\begin{figure}[t]
    \captionsetup[subfigure]{labelformat=empty}
    \centering
    \subfloat[Ours: RoadAnomaly21]{\includegraphics[width=0.24\linewidth]{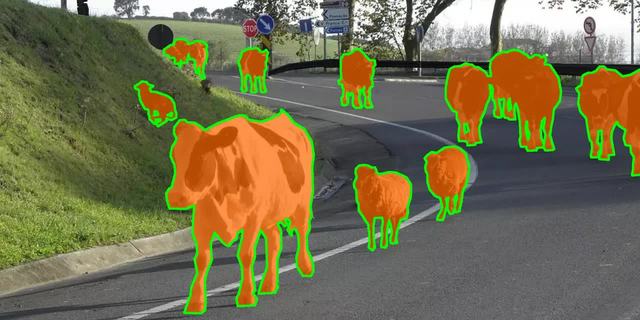}}~
    \subfloat[Fishyscapes]{\includegraphics[width=0.24\linewidth]{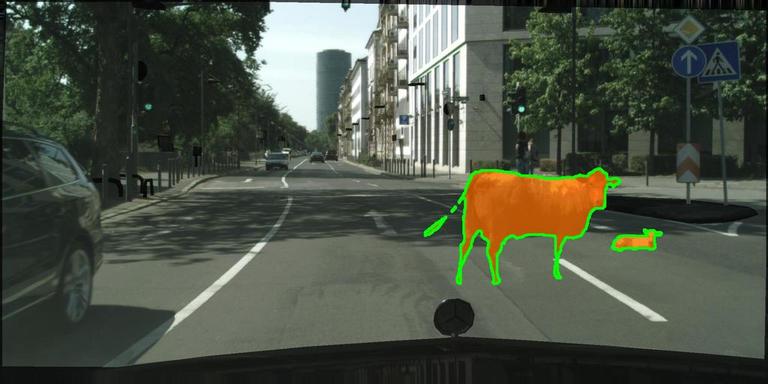}}~
    \subfloat[Ours: RoadObstacle21]{\includegraphics[width=0.24\linewidth]{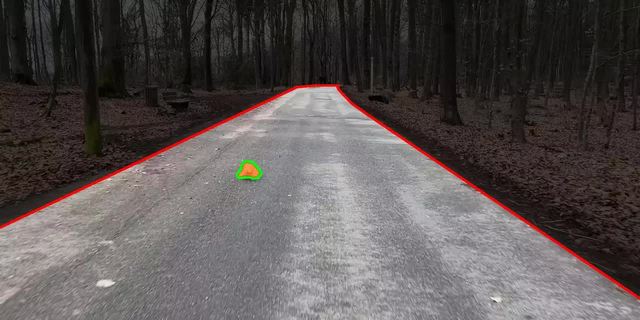}}~
    \subfloat[LostAndFound]{\includegraphics[width=0.24\linewidth]{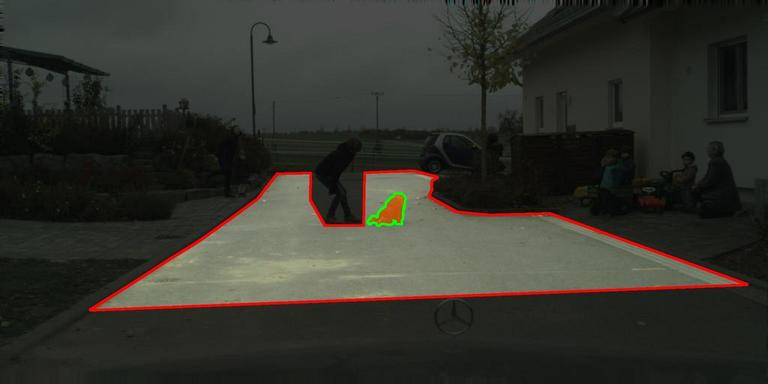}}
    \caption{Comparison of images from our and existing public datasets. Anomalies / obstacles are highlighted in orange, darkened regions are excluded from the evaluation.
    In RoadAnomaly21, anomalies may appear everywhere in the image. In contrast to Fishyscapes, where anomalous objects are synthetic, all RoadAnomaly21 images are  real. 
    In RoadObstacle21, the region of interest is restricted to the drivable area with obstacles ahead. This is comparable to LostAndFound, where the labeling, however, is not always consistent, \eg children are anomalies but other humans not.
    }
    \label{fig:page1}
\end{figure}

The advent of high-quality publicly-available datasets, such as Cityscapes \cite{Cordts2016Cityscapes}, BDD100k \cite{yu2020bdd100k}, A2D2 \cite{a2d2Dataset} and COCO \cite{Lin14COCO} has hugely contributed to the progress in semantic segmentation. However, while state-of-the-art deep neural networks (DNNs) yield outstanding performance on these datasets, they typically provide predictions for a closed set of semantic classes. Consequently, they are unable to classify an object as \emph{none of the known categories} \cite{zhang17universum}.
Instead, they tend to be overconfident in their predictions, even in the presence of previously-unseen objects~\cite{Hein_2019_CVPR}, which precludes the use of uncertainty to identify the corresponding anomalous regions.

Nevertheless, reliability in the presence of unknown objects is key to the success of applications that have to face the diversity of the real world, \eg, perception in automated driving. This has motivated the creation of 
benchmarks such as Fishyscapes \cite{Blum19fishyscapes} or CAOS~\cite{hendrycks2020scaling}. While these benchmarks have enabled interesting experiments, the limited real-world diversity in Fishyscapes, the lack of a public leader board and of a benchmark suite in CAOS, and the reliance on synthetic images in both benchmarks hinder proper evaluation of and comparisons between the state-of-the-art methods.

In this paper, motivated by the limitations of existing anomaly segmentation datasets and by the emerging body of works in this direction~\cite{Angus2019, Blum19fishyscapes, Bruegge20, chan2020entropy, Isobe17, jourdan20, Lis19, Mehrtash20, Oberdiek20}, we introduce the \emph{SegmentMeIfYouCan}\footnote{\url{https://www.segmentmeifyoucan.com/}} benchmark. It is accompanied with two datasets, consisting of diverse and manually annotated real images, a public leader board and an evaluation suite, providing in-depth analysis and comparisons, to facilitate the development of road anomaly segmentation methods.

Our benchmark encompasses two separate tasks. The first one consists of strict anomaly segmentation, where any previously-unseen object is considered as an anomaly. Furthermore, motivated by the observation that the boundary between known and unknown classes can sometimes be fuzzy, for instance for \emph{car} \vs \emph{van}, we introduce the task of obstacle segmentation, whose goal is to identify all objects on the road, may they be from known classes or from unknown ones.

For the anomaly track, we provide a dataset of 100 images with pixel-wise annotations over two classes (anomaly, not anomaly) and a void class, which, in analogy to Cityscapes, signals the pixels that are excluded from the evaluation. We consider any object that strictly cannot be seen in the Cityscapes data as anomalous, appearing anywhere in the image.
For the obstacle track, our dataset contains 327 images with analogous annotation (obstacle, not obstacle, void), and focuses only on the road as region of interest. The focus in this track is of more practical need, \eg for automated driving systems, targeting obstacles that may cause hazardous street situations, see \Cref{fig:page1}. All images of our datasets are publicly available for download\footnotemark[2], together with a benchmark suite that computes both established pixel-wise metrics and recent component-wise ones.

In the remainder of this paper, we first review existing anomaly detection datasets, methods and evaluation metrics in more detail. We then describe our new benchmark and provide extensive experiments comparing state-of-the-art road anomaly / obstacle segmentation methods on our datasets and on other related ones, showing the difficulty of the models on the proposed benchmarks.

\section{Related Work}

In this section we first review previous datasets for anomaly detection, with some of them being designed for road anomaly segmentation. Then we briefly describe some of the methods on anomaly and obstacle segmentation.

\subsection{Datasets and Benchmarks}
Existing methods for anomaly detection have often been evaluated on their ability to separate images from two different source distributions, such as separating MNIST from FashionMNIST~\cite{Choi2018-lm,Meinke20, Van_Amersfoort2020-vq}, NotMNIST~\cite{Van_Amersfoort2020-vq}, or Omniglot~\cite{lake2015human}, and separating CIFAR-10 from SVHN~\cite{Lee2018mahala, Meinke20, Van_Amersfoort2020-vq} or LSUN~\cite{Lee2018mahala,liang18odin, Meinke20}. Such experiments can be found in many works, including~\cite{Hendrycks2017msp,Choi2018-lm,Lee2018mahala,liang18odin,Van_Amersfoort2020-vq,Meinke20}.

For semantic segmentation, a similar task was therefore proposed by the WildDash benchmark~\cite{Zendel2018-ru} that analyzes semantic segmentation methods trained for driving scenes on a range of failure sources, including full-image anomalies, such as images from the beach.
In our work, by contrast, we focus on the problem of robustness to anomalies that only cover a small portion of the image, and on the methods that aim to segment such anomalies, \ie method for the task of \emph{anomaly segmentation}.

One prominent dataset tackling the task of anomaly segmentation is LostAndFound~\cite{Pinggera16LostAndFound}, which shares the same setup as Cityscapes~\cite{Cordts2016Cityscapes} but includes anomalous objects / obstacles in various street scenes in Germany.
LostAndFound contains 9 different object types as anomalies, and only has annotations for the anomaly and the road surface. Furthermore, it considers children and bicycles as anomalies, even though they are part of the Cityscapes training set, and it contains several labeling mistakes.
Although we filter and refine LostAndFound in this work\footnote{In the following, we refer to the LostAndFound subset without the images of children, bicycles and invalid annotations as ``LostAndFound-NoKnown''.}, similar to Fishyscapes \cite{Blum19fishyscapes}, the low diversity of anomalies persists.

The CAOS BDD-Anomaly benchmark~\cite{hendrycks2020scaling} suffers from a similar low-diversity issue, arising from its use of only 3 object classes sourced from the BDD100k dataset~\cite{yu2020bdd100k} as anomalies (besides including several labeling mistakes, see \Cref{sec:bdd-anomaly-supp}). Both Fishyscapes and CAOS try to mitigate this low diversity by complementing their real images with synthetic data. Such synthetic data, however, is not realistic and not representative of the situations that can arise in the real world.

In general, the above works illustrate the shortage of diverse real-world data for anomaly segmentation. Additional efforts in this regard have been made by sourcing and annotating images of animals in street scenes~\cite{Lis19}, and by leveraging multiple sensors, including mainly LiDAR, to detect obstacles on the road~\cite{Singh2020-fw}.
In any event, most of the above datasets are fully published with annotations, allowing methods to overfit to the available anomalies. Furthermore, apart from Fishyscapes, we did not find any public leader boards that allow for a trustworthy comparison of new methods. To provide a more reliable test setup, we do not share the labels and request predictions of the shared images to be submitted to our servers. Furthermore, we provide a leader board, which we publish alongside two novel real-world datasets, namely RoadAnomaly21 and RoadObstacle21. A summary of the main properties of the mentioned datasets is given in \Cref{tab:dataset_comparison}. Our main contribution in both proposed datasets is the diversity of the anomaly categories and of the scenes. 

In RoadAnomaly21, anomalies can appear anywhere in the image, which is comparable to Fishyscapes LostAndFound~\cite{Blum19fishyscapes} and CAOS BDD-Anomaly~\cite{hendrycks2020scaling}. Although the latter two datasets are larger, their images only show a limited diversity of anomaly types and scenes because they are usually frames of videos captured in single scenes. By contrast, in our dataset every image shows a unique scene, with at least one out of 26 different types of anomalous objects and each sample widely differs in size, ranging from 0.5\% to 40\% of the image. 

In RoadObstacle21, all anomalies (or obstacles) appear on the road, making this dataset comparable to LostAndFound~\cite{Pinggera16LostAndFound} and the LiDAR guided Small Obstacle dataset \cite{Singh2020-fw}. Again, the latter two datasets contain more images than ours, however, the high numbers of images result from densely sampling frames from videos. Consequently, those two datasets lack in object diversity (9 and 6 categories, respectively, compared to 31 in our dataset). Furthermore, the videos are recorded under perfect weather conditions, while RoadObstacle21 shows scenes in diverse situations, including night, dirty roads and snowy conditions.

\begin{table}[t]
    \setlength{\tabcolsep}{3.5pt}
    \begin{minipage}{\linewidth}
    \scalebox{.68}{
    \begin{tabular}{l|ccccccc}
        & anomaly & non-anomaly & diverse  & different & dataset & ground truth (gt) & mean \& std of gt size \\
        Dataset & pixels & pixels & scenes & anomalies & size & components & relative to image size\\\hline
        Fishyscapes LostAndFound val \cite{Blum19fishyscapes} & 0.23\% & 81.13\% & 12 & 7 & 373 & 165 & 0.13\% $\pm$ 0.23\% \\
        CAOS BDD-Anomaly test \cite{hendrycks2020scaling} & 0.83\%  & 81.28\%  & 810  & 3 & 810 & 1231 & 0.55\% $\pm$ 1.84\% \\
        \rowcolor{Gray} Ours: RoadAnomaly21 test & 13.83\% & 82.17\% & 100 & 26 & 100 & 262 & 4.12\% $\pm$ 7.29\% \\ \hline
        LostAndFound test (NoKnown) \cite{Pinggera16LostAndFound} & 0.12\% & 15.31\% & 13 (12) & 9 (7) & 1203 (1043) & 1864 (1709) & 0.08\% $\pm$ 0.16\% \\
        LiDAR guided Small Obstacle test \cite{Singh2020-fw} & 0.07\% & 36.09\% & 2 & 6 & 491 & 1203 & 0.03\% $\pm$ 0.07\% \\
        \rowcolor{Gray} Ours: RoadObstacle21 test & 0.12\% & 39.08\% & 8 & 31 & 327 & 388 & 0.10\% $\pm$ 0.25\% \\
    \end{tabular}}
    \end{minipage}
    \vskip2mm
    \begin{minipage}[t]{0.49\linewidth}
    \vspace{0pt}
    \scalebox{.68}{
    \begin{tabular}{l|ccccccc}
        Dataset (as above) & \parbox{\widthof{are private}}{\centering labels are private} & \parbox{\widthof{conditions}}{\centering weather conditions\strut} & geography\\\hline
        Fishyscapes val &(\cmark  in test set) & clear & DE\\
        CAOS BDD test & \xmark  & clear, snow, night, rain & US\\
        \rowcolor{Gray} Ours: RA21 test & \cmark & clear, snow & global\\ \hline
        LaF test & \xmark & clear & DE\\
        Small Obs. test & \xmark & clear & IN \\
        \rowcolor{Gray} Ours: RO21 test & \cmark & clear, snow, night & CH, DE\\
    \end{tabular}%
    }%
    \end{minipage}\hfill
    \begin{minipage}[t]{0.48\linewidth}
    \vspace{0pt}
    \vspace{-3mm}
    \caption{Main properties of comparable real-world anomaly (top three rows) and obstacle (bottom three rows) segmentation datasets.
    Our main contribution is the diversity of the anomaly (or obstacle) categories and of the scenes. Note that ``void'' pixels are not included in this table.}\label{tab:dataset_comparison}
    \end{minipage}
    \vspace{-4mm}
\end{table}

\subsection{Anomaly and Obstacle Segmentation} \label{sec:anomaly-seg}

Anomaly detection was initially tackled in the context of image classification, by developing post-processing techniques aiming to adjust the confidence values produced by a classification DNN~\cite{Hendrycks2017msp, Lee2018mahala, liang18odin, Hein_2019_CVPR, Meinke20}. Although originally designed for image-level anomaly detection, most of these methods can easily be adapted to anomaly segmentation~\cite{Angus2019, Blum19fishyscapes} by treating each individual pixel in an image as a potential anomaly.

Another relevant approach consists of estimating the uncertainty of the predictions, leveraging the intuition that anomalous image regions should correlate with high uncertainty. One way of doing so is Bayesian (deep) learning~\cite{Mackay1992, neal2012}, where the model parameters are treated as distributions. Because of the computational complexity, approximations to Bayesian inference have been developed~\cite{gal2016dropout,Atanov2019,Gustafsson2020EvaluatingSB,Lakshminarayanan17} and extended to semantic segmentation~\cite{Badrinarayanan17BayesianSegNet, Kendall17, mukhoti2019bdl}. Instead of reasoning about uncertainty, other non-Bayesian approaches tune previously-trained models to the task of anomaly detection by either modifying its architecture or exploiting additional data.
For example, in~\cite{DeVries2018}, anomaly scores are learned by adding a separate branch to the DNN. Instead of modifying the DNNs's architecture, other approaches~\cite{hendrycks19oe, Meinke20} incorporate an auxiliary ``out-of-distribution'' (OoD) dataset during training, which is disjoint from the actual training dataset. These ideas have been employed for anomaly segmentation in~\cite{Bevandic2019, chan2020entropy, jourdan20}.

A recent line of work performs anomaly segmentation via generative models that reconstruct / resynthesize the original input image. The intuition is that the reconstructed images will better preserve the appearance of regions containing known objects than those with unknown ones. Pixel-wise anomaly detection is then performed by identifying the discrepancies between the original and reconstructed image. This approach has been used not only for anomaly segmentation~\cite{dibiase2021pixelwise, Lis19, xia2020synthesize} but also specifically for road obstacle detection~\cite{Creusot15, lis2020detecting, Munawar17}.

It is important to note that there are some related works with different definitions of anomaly segmentation. For example, \cite{Bergmann2019-ho} evaluates the segmentation of industrial production anomalies like scratches, and in medical contexts anomaly segmentation can be understood as the detection of diseased parts on e.g. tomography images~\cite{SeebockOSWBKLS20} or brain MRIs~\cite{BAUR2021101952}.
What we define as anomaly segmentation will be discussed in detail in the next \Cref{sec:benchmark}.

\begin{figure*}[t]
    \centering
    \captionsetup[subfigure]{}
    \subfloat[RoadAnomaly21]{
    \begin{minipage}{0.29\textwidth}
        \centering
        \includegraphics[height=1.4cm]{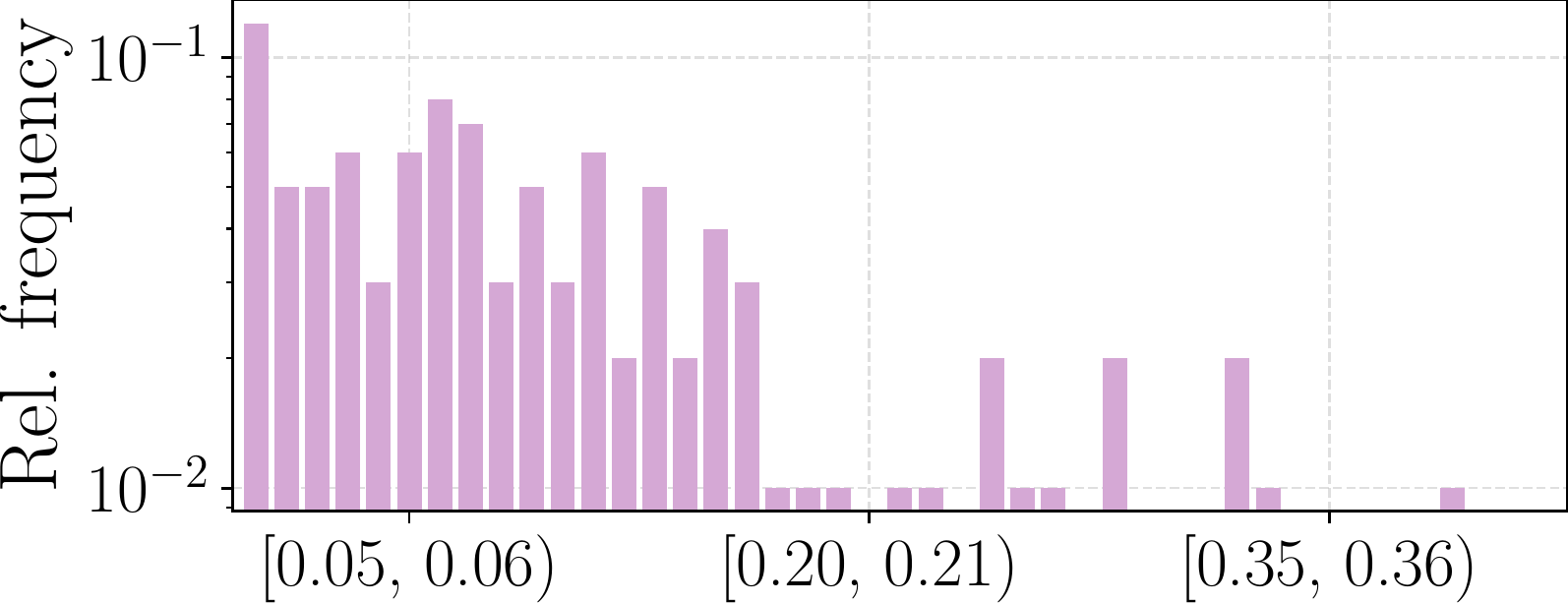} \\ ~~~\tiny{Fraction of anomaly pixels in an image}
    \end{minipage}%
    \begin{minipage}{0.21\textwidth}
        \centering
        \includegraphics[height=1.4cm]{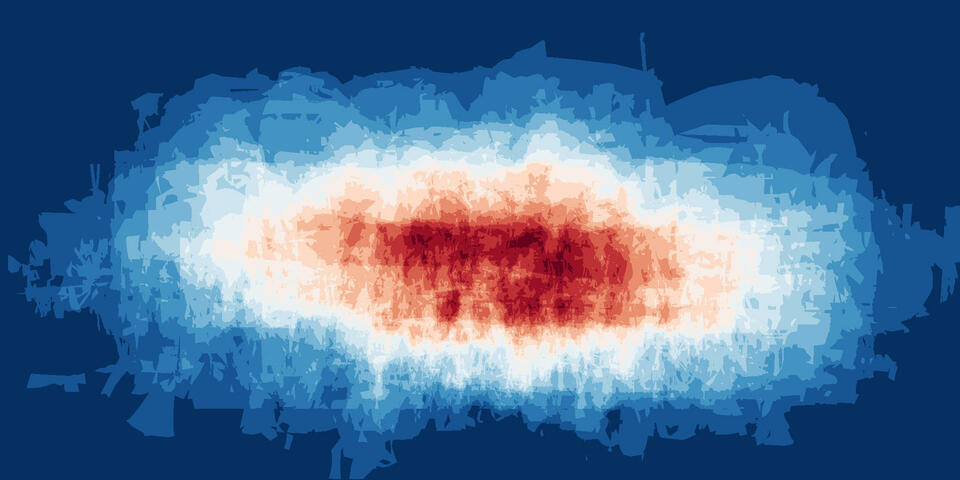}\\ ~\tiny{Anomaly pixel distribution}
    \end{minipage}}%
    \subfloat[RoadObstacle21]{
    \begin{minipage}{0.29\textwidth}
        \centering
        \includegraphics[height=1.4cm]{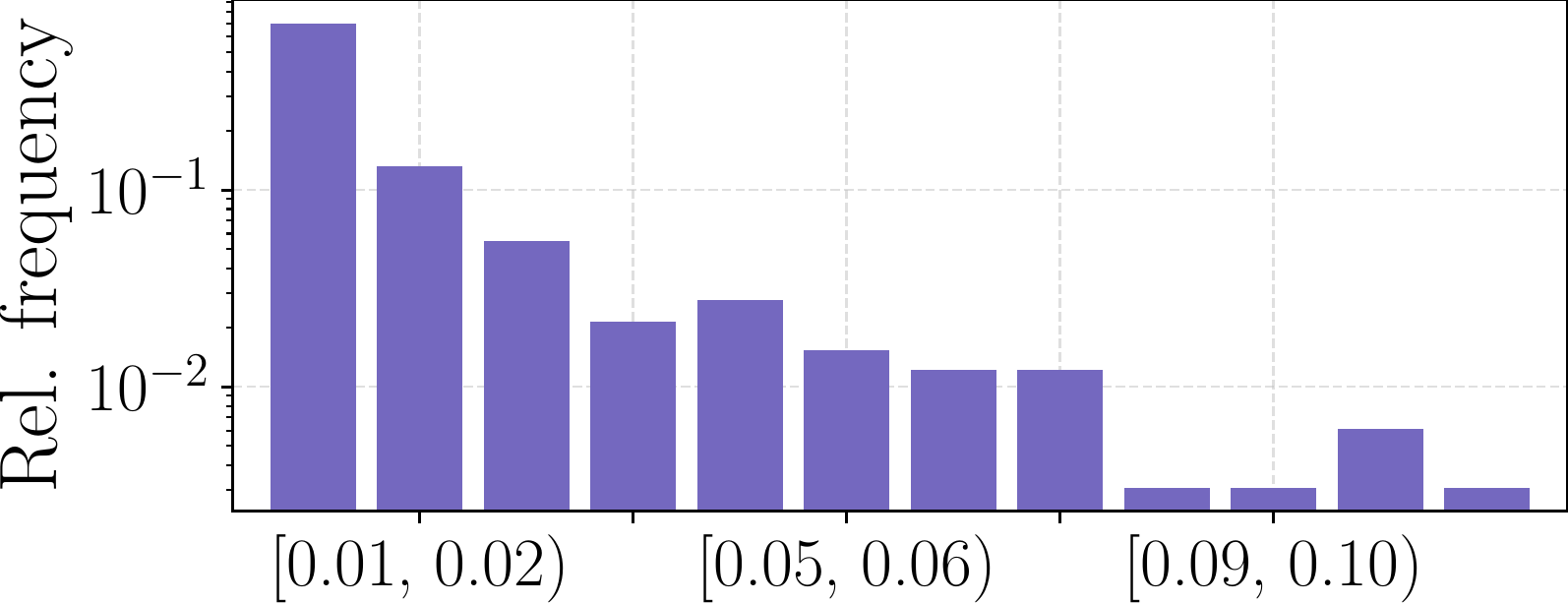} \\ ~~~\tiny{Fraction of obstacle pixels in an image}
    \end{minipage}%
    \begin{minipage}{0.21\textwidth}
        \centering
        \includegraphics[height=1.4cm]{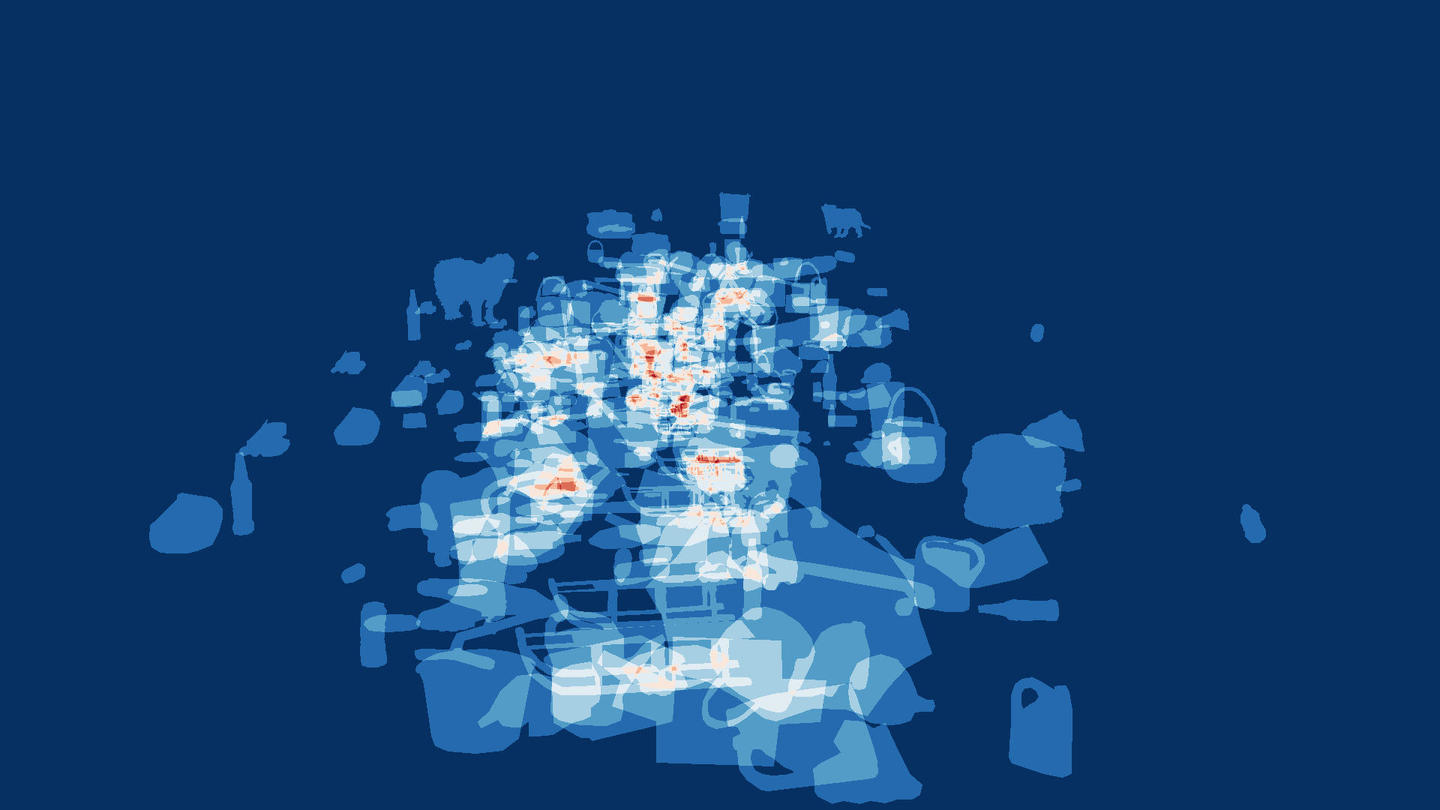}\\ ~\tiny{Obstacle pixel distribution}
    \end{minipage}}
    \caption{Relative frequency of annotated anomaly / obstacle pixels within an image over the 100 images in the RoadAnomaly21 test dataset (left) and the 327 images in the RoadObstacle21 test dataset (right), respectively.
    Here, the fraction of anomaly / obstacle pixels serves as a proxy for the size of the objects of interest within an image.
    Note that the y-axes of the histograms are $\log$ scaled.
    }
    \label{fig:hist_anoamly}
\end{figure*}

\section{Benchmark Description}
\label{sec:benchmark}

The aim of our benchmark is two-fold. On one hand, by providing diverse data with consistent annotations, we seek to facilitate progress in general semantic anomaly segmentation research. On the other hand, by focusing on road scenes, we expect our benchmark to accelerate the progress towards much needed segmentation/obstacle-detection methods for safe automated driving.

To achieve these goals, our benchmark covers two tasks. First, it tackles the general problem of anomaly segmentation, aiming to identify the image regions containing object classes that have never been seen during training, and thus for which semantic segmentation cannot be correct. 
This is necessary for any reliable decision making process and it is of great importance to many computer vision applications. Note that, in accordance to \cite{Blum19fishyscapes,hendrycks2020scaling}, we define anomaly as objects that do not fit any of the class definitions in the training data. In some works, anomaly may be used to describe visually different inputs like \eg a car in a novel color, which does not fit our definition.

This strict definition of semantic anomalies, however, can sometimes be ill-defined because (i) existing semantic segmentation datasets, such as Cityscapes~\cite{Cordts2016Cityscapes}, often contain ambiguous and ignored regions (annotated as \emph{void}), which are not strictly anomalies since they are seen during training; (ii) the boundary of some classes is fuzzy, \eg, cars \vs vans \vs rickshaws, making it unclear whether some regions should be considered as anomalous or not.
To address these issues, and to account for the fact that automated driving systems need to make sure that the road ahead is free of \emph{any} hazardous objects, we further incorporate obstacle segmentation as a second task in our benchmark, whose goal is to identify any non-drivable region on the road, may the non-drivable region correspond to a known object class or an unknown one. \looseness-1

\subsection{Benchmark Tracks and Datasets}\label{sec:datasets}
We now present the two tracks in our benchmark, corresponding to the two tasks discussed above. Each track contains its own dataset with different properties and is therefore evaluated separately in our benchmark suite. An overview comparing our datasets to related public ones is given in \Cref{tab:dataset_comparison}.

\textbf{RoadAnomaly21.}
The road anomaly track benchmarks general anomaly segmentation in full street scenes. It consists of an evaluation dataset of 100 images with pixel-level annotations. The data is an extension of the one introduced in~\cite{Lis19}, now including a broader collection of images and finer-grain labeling. In particular, we removed low quality images and ones lacking clear road scenes. Besides, we removed labeling mistakes, added the void class and included 68 newly collected images. Each image contains at least one anomalous object, \eg, an animal or an unknown vehicle.
The anomalies can appear anywhere in the image, which were collected from web resources and therefore depict a wide variety of environments.
The distribution of object sizes and location is shown in~\Cref{fig:hist_anoamly}(a).
Moreover, we provide 10 additional images with annotations such that users can check the compatibility of their methods with our benchmark implementation.

\textbf{RoadObstacle21.}
The road obstacle track focuses on safety for automated driving. The objects to segment in the evaluation data always appear on the road ahead, \ie they represent realistic and hazardous obstacles that are critical to detect.
Our dataset consists of 222 new images taken by ourselves and 105 from \cite{lis2020detecting}, summing up to a total of 327 evaluation images with pixel-level annotations.
The region of interest in these images is given by the road, which is assumed to belong to the known classes on which the algorithm was trained. The obstacles in this dataset are chosen such that they all can be understood as anomalous objects as well, \eg, stuffed toys, sleighs or tree stumps. They appear at different distances (one distance per image) and are surrounded by road pixels. This allows us to focus our evaluation on the obstacles, as other objects lie outside the region of interest. The distribution of object sizes and location is shown in~\Cref{fig:hist_anoamly}(b). Moreover, this dataset incorporates different road surfaces, lighting and weather conditions, thus encompassing a broad diversity of scenes.
An extra track of additional 85 images with scenes at night and in extreme weather, such as snowstorms, is also available. However, the latter subset is excluded from our numerical experiments due to the significant domain shift. Lastly, we provide 30 additional images with annotations such that users can check the compatibility of their methods with our benchmark implementation.\looseness-1

\textbf{Labeling Policy.}
In both datasets, the pixel-level annotations include three classes: 
\textbf{1)} anomaly / obstacle, \textbf{2)} not anomaly / not obstacle, and \textbf{3)} void.

In RoadAnomaly21, the 19 Cityscapes evaluation classes~\cite{Cordts2016Cityscapes}, on which most semantic segmentation DNNs are trained, serve as basis to judge whether an object is considered anomalous or not. Everything that fits in the class definitions of Cityscapes is thus labeled as \emph{not anomaly}. This track focuses on the detection of objects which are semantically different from those in the Cityscapes training data.
Therefore, if image regions cannot be clearly assigned to any of the Cityscapes classes, they are labeled as \emph{anomaly}. The objects, which are not the main anomalies of interest in the context of street scenes, are labeled as \emph{void} and excluded from our evaluation.
The latter class include, for instance, mountains or water in the image background, and street lights. In ambiguous cases, which \eg can arise from a strong domain shift to Cityscapes, we assign the void class as well to properly evaluate semantic anomaly segmentation.

In RoadObstacle21, the task is defined as distinguishing between drivable area and non-drivable area. The goal is to make sure that the road ahead of the ego-car is free of any hazard, irrespective of the object category of potential obstacles. Therefore, the drivable area is labeled as \emph{not obstacle}. This class particularly also includes regions on the road, which visually differ from the rest of the road. Moreover, every object, which is visually enclosed in the drivable area, is labeled as \emph{obstacle}. All image regions outside the road are assigned to the \emph{void} class and ignored in the evaluation.

As a quality assessment for both tracks, each labeled image was reviewed by at least three people in order to guarantee the highest quality of labels.

\subsection{Performance Metrics}\label{sec:metrics}

For the sake of brevity, in what follows we refer to both anomalies and obstacles as \emph{anomalies}.

\textbf{Pixel level.}
Let $\mathcal{Z}$ denote the set of image pixel locations. 
A model with a binary classifier providing anomaly scores $s(x)\in\mathbb{R}^{|\mathcal{Z}|}$ for an image $x\in\mathcal{X}$ (from a dataset $\mathcal{X} \subseteq [0,1]^{N\times |\mathcal{Z}| \times 3}$ of $N$ images) discriminates between the two classes \emph{anomaly} and \emph{non-anomaly}. We evaluate the separability of the pixel-wise anomaly scores via the area under the precision-recall curve (AuPRC), where precision and recall are considered as functions of some threshold $\delta\in\mathbb{R}$ applied to $s(x) ~\forall~ x\in \mathcal{X}$. The AuPRC puts emphasis on detecting the minority class, making it particularly well suited as our main pixel-wise evaluation metric since the pixel-wise class distributions of RoadAnomaly21 and RoadObstacle21 are considerably unbalanced, \cf \Cref{tab:dataset_comparison}.

To consider the safety point of view, we also include the false positive rate at 95\% true positive rate (FPR$_{95}$) in our evaluation. The FPR$_{95}$ metric indicates how many false positive predictions must be made to reach the desired true positive rate. Note that, any prediction which is contained in a ground-truth labeled region of the class void is not counted as false positive, \cf \Cref{sec:datasets}. In particular for the RoadObstacle21 dataset the evaluation is therefore restricted to the road area.

\textbf{Component level.} From a practitioner's perspective, it is very important to detect all anomalous regions in the scene, regardless of their size, \ie, the number of pixels they cover. However, pixel-level metrics may neglect small anomalies. While one could thus focus on object detection metrics, the notion of individual objects is in fact not relevant for anomaly (region) detection. To satisfy these requirements, we therefore consider performance metrics acting at the component level.

The main metrics for component-wise evaluation are the numbers of \emph{true-positives} (TP), \emph{false-negatives} (FN) and \emph{false-positives} (FP). Considering anomalies as the positive class, we use a component-wise localization and classification quality measure to define the TP, FN and FP components.
Specifically, we define this measure as an adjusted version of the component-wise intersection over union (sIoU), introduced in \cite{Rottmann18metaseg}. In particular, while in \cite{Rottmann18metaseg} the sIoU is computed for predicted components, we consider the sIoU for ground-truth components to compute TP and FN. To compute FP, we employ the positive predictive value (PPV, or component-wise precision) for predicted components as quality measure. We discuss the definitions of these quantities in more detail below.

Let $\mathcal{Z}_{c}$ be the set of pixel locations labeled with class $c=\textrm{``anomaly''}$ in the dataset $\mathcal{X}$. We consider a connected component of pixels (where the 8 pixels surrounding pixel $z$ in image $x\in\mathcal{X}$ are taken to be its neighbors) that share the same class label as a \emph{component}. Then, let us denote by $\mathcal{K} \subseteq \mathcal{P}(\mathcal{Z}_{c})$, with $\mathcal{P}(\mathcal{S})$ the power set of a set $\mathcal{S}$, the set of anomaly components according to the ground truth, and by $\hat{\mathcal{K}} \subseteq \mathcal{P}(\mathcal{Z}_{c})$ the set of components predicted to be anomalous by some machine learning model.

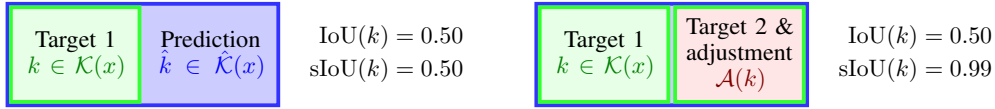
\begin{figure}
    \centering
    \scalebox{0.9}{
    
    \begin{minipage}{0.49\linewidth}
    \begin{tikzpicture}

    \draw [color=blue!80, fill=blue!20, ultra thick] (-2,0.75) rectangle (2,-0.75);
    \draw [color=green!80, fill=white!90!green, ultra thick] (-1.95,0.7) rectangle (-0.05,-0.7);
    \node[text width=2cm, align=center] at (-1,0) {Target 1 \\ \textcolor{black!50!green}{$k\in\mathcal{K}(x)$}};
    \node[text width=3cm, align=center] at (1,0) {Prediction \\ \textcolor{blue}{$\hat{k}\in\mathcal{\hat{K}}(x)$}};

    \node at (3.6,0) {$\begin{aligned}%
    \mathrm{IoU}(k) &= 0.50 \\
    \mathrm{sIoU}(k) &= 0.50
    \end{aligned}$};
    \end{tikzpicture}
    \vspace{1pt}
    \end{minipage}
~~~~~~~~~
    \begin{minipage}{0.49\linewidth}
    \begin{tikzpicture}

    \draw [color=blue!80, fill=blue!20, ultra thick] (-2,0.75) rectangle (2,-0.75);
    \draw [color=green!80, fill=white!90!green, ultra thick] (-1.95,0.7) rectangle (-0.05,-0.7);
    \draw [color=green!80, fill=white!90!red, ultra thick] (1.95,0.7) rectangle (0.05,-0.7);
    \node[text width=2cm, align=center] at (-1,0) {Target 1 \\ \textcolor{black!50!green}{$k\in\mathcal{K}(x)$}};
    \node[text width=2cm, align=center] at (1,0) {Target 2 \& \\ adjustment \textcolor{black!50!red}{$\mathcal{A}(k)$}};

    \node at (3.6,0) {$\begin{aligned}
    \mathrm{IoU}(k) &= 0.50 \\
    \mathrm{sIoU}(k) &= 0.99
    \end{aligned}$};
    \end{tikzpicture}
    \vspace{1pt}
    \end{minipage}

    }
    \vspace{-3mm}
    \caption{Illustration of the ordinary component-wise intersection over union ($\mathrm{IoU}$) and the adjusted one ($\mathrm{sIoU}$).
    In both examples above, the prediction $\hat{k}$ (blue rectangle) is the same but covers different targets (green rectangles). On the left, both IoU and sIoU yield the same score. On the right, IoU punishes the prediction as it does not cover each object precisely. By contrast, sIoU checks how much the predictions cover the ground-truth regions, independently of whether prediction/ground truth belongs to a single or multiple objects. In automated driving, it is more important to detect all anomalous regions (whether they belong to single or multiple objects), rather than to detect each object precisely. Since two targets are separated by at least one pixel, $\mathrm{IoU}=\mathrm{sIoU}=1$ if and only if the prediction covers one target perfectly. \looseness-1 $\phantom{\hat{k}}$}
    \label{fig:adj_iou}
\end{figure}

Formally, the sIoU is a mapping $\mathrm{sIoU} : \mathcal{K} \to [0,1]$. For $k \in \mathcal{K}$, 
it is defined as
\begin{equation}\label{eq:adj_iou_gt}
    \mathrm{sIoU}(k)  := \frac{|k \cap \hat{K}(k)|}{  | (k \cup \hat{K}(k) ) \setminus \mathcal{A}(k)|  }
~~~~~\text{ with }~~~~
    \hat{K}(k) = \!\!\!\!\!\! \bigcup_{\hat{k} \in \hat{\mathcal{K}}, \hat{k} \cap k \neq \emptyset }\!\!\! \hat{k} ~~~
\end{equation}
and $\mathcal{A}(k) = \{ z\in k': k' \in \mathcal{K} \setminus \{k\} \}$.
With the adjustment $\mathcal{A}(k)$, the pixels are excluded from the union if and only if they correctly intersect with another ground-truth component $k' \in \mathcal{K}(x)$, which is not equal to $k$. This may happen when one predicted component covers multiple ground-truth components, as illustrated in \Cref{fig:adj_iou}.
Given some threshold $\tau\in[0,1)$, we then call a target $k\in\mathcal{K}$ $\mathrm{TP}$ if $\mathrm{sIoU}(k)>\tau$, and $\mathrm{FN}$ otherwise.
We refer to \Cref{sec:siou-example} for qualitative examples of the difference between IoU and sIoU.

For the other error type, \ie, FP, we compute the  PPV (or precision) for $\hat{k} \in \hat{\mathcal{K}}$, which is defined as
\begin{equation}\label{eq:prec_pred}
    \mathrm{PPV}(\hat{k}) := \frac{|\hat{k}\cap K(\hat{k})|}{|\hat{k}|} \, ,
\end{equation}
We then call a predicted component $\hat{k}\in\hat{\mathcal{K}}$ $\mathrm{FP}$ if $\mathrm{PPV}(\hat{k})\leq\tau$.

As an overall metric, we additionally include the component-wise $F_1$-score defined as
\begin{equation}\label{eq:comp-f1}
    F_1(\tau) := \frac{2 \cdot \mathrm{TP}(\tau)}{2 \cdot \mathrm{TP}(\tau) + \mathrm{FN}(\tau) + \mathrm{FP}(\tau)}  \in [0,1]\;,
\end{equation}
which summarizes the $\mathrm{TP}$, $\mathrm{FN}$ and $\mathrm{FP}$ quantities (that depend on $\tau$). The component-level metrics allow one to evaluate localization of objects irrespective of their size and hence big objects will not dominate these metrics. In addition, while object detection metrics punish predictions that cover multiple ground-truth objects or vice-versa, our component-level metric does not do so, \cf \Cref{fig:adj_iou}.

\subsection{Evaluated Methods}\label{sec:methods}

Several anomaly segmentation methods have already been evaluated on our benchmark and constitute our initial leader board. We evaluate at least one method per type discussed in \Cref{sec:anomaly-seg}, namely
\begin{itemize}[noitemsep]
    \item \emph{Methods originating from image classification}: \textbf{maximum softmax probability}~\cite{Hendrycks2017msp}, \textbf{ODIN}~\cite{liang18odin}, \textbf{Mahalanobis distance}~\cite{Lee2018mahala};
    \item \emph{Bayesian model uncertainty}: \textbf{Monte Carlo (MC) dropout}~\cite{mukhoti2019bdl}, \textbf{ensemble}~\cite{Lakshminarayanan17};
    \item \emph{Learning to identify anomalies}: \textbf{learned embedding density}~\cite{Blum19fishyscapes}, \textbf{void classifier}~\cite{Blum19fishyscapes}, \textbf{maximized softmax entropy}~\cite{chan2020entropy};
    \item \emph{Reconstruction via generative models}: \textbf{image resynthesis}~\cite{Lis19}, \textbf{SynBoost}~\cite{dibiase2021pixelwise} and \textbf{road inpainting} (obstacle track only)~\cite{lis2020detecting}.
\end{itemize}
All methods have an underlying semantic segmentation DNN trained on Cityscapes and provide pixel-wise anomaly scores.
A semantic segmentation DNN trained on Cityscapes is also our recommendation as underlying model, however, we leave it up to the participants which network and training data they use. Furthermore, some evaluated methods additionally employ out-of-distribution (OoD) data to tune the anomaly detector. For our set of methods, this would be any data with labels semantically different from the Cityscapes train classes. OoD data is also allowed to be used to alleviate the effects of a potential domain shift.
For additional details on the methods, we refer the reader to \Cref{sec:methods-supp}.

\begin{table*}[t]
    \setlength{\tabcolsep}{2pt}
    \begin{center}
    \scalebox{0.66}{
    \begin{tabular}{l||c||acc||cc|ccc|ccc|ccc|g}
    \toprule
    & & \multicolumn{3}{c||}{Pixel-level} & \multicolumn{12}{c}{Component-level}  \\
    \midrule
    \midrule
    & requires & \multicolumn{3}{c||}{Anomaly scores} & $k\in\mathcal{K}$ & $\hat{k}\in\mathcal{\hat{K}}$ & \multicolumn{3}{c|}{$ \tau=0.25$} & \multicolumn{3}{c|}{$ \tau=0.50$} & \multicolumn{3}{c|}{$\tau=0.75$} & \multicolumn{1}{c}{} \\
    Method & OoD data & \multicolumn{1}{c}{AuPRC $\uparrow$} & FPR$_{95}$ $\downarrow$ & $F_1^*\uparrow$ & $\overline{\mathrm{sIoU}}$ $\uparrow$ & $\overline{\mathrm{PPV}}$ $\uparrow$ & FN $\downarrow$ & FP $\downarrow$ & $F_1\uparrow$ & FN $\downarrow$ & FP $\downarrow$ & $F_1\uparrow$ & FN $\downarrow$ & FP $\downarrow$ & $F_1\uparrow$ & \multicolumn{1}{c}{$\overline{F_1}\uparrow$}\\
    \midrule
    Maximum softmax \cite{Hendrycks2017msp} & \xmark &              28.0 &                    72.0 &                  34.2 &                  15.5 &                    15.3 &                    204 &                    681 &                    11.6 &                    233 &                    714 &                     5.8 &                    256 &                    744 &                     1.2 &                       5.9 \\
    ODIN \cite{liang18odin} & \xmark &              33.1 &                    71.7 &                  39.1 &                  19.6 &                    17.9 &                    181 &                    924 &                    12.8 &                    226 &                    985 &                     5.6 &                    254 &                   1043 &                     1.2 &                       6.0 \\
    Mahalanobis \cite{Lee2018mahala} & \xmark &              20.0 &                    87.0 &                  31.9 &                  14.8 &                    10.2 &                    206 &                   1433 &                     6.4 &                    241 &                   1478 &                     2.4 &                    257 &                   1512 &                     0.6 &                       2.9 \\
    MC dropout \cite{mukhoti2019bdl} & \xmark &             28.9 &                    69.5 &                  39.0 &                  20.5 &                    17.3 &                    175 &                   1320 &                    10.4 &                    225 &                   1391 &                     4.4 &                    252 &                   1459 &                     1.2 &                       4.9 \\
    Ensemble \cite{Lakshminarayanan17}  & \xmark &                                       17.7 &                                        91.1 &                                      27.8 &                                            16.4 &                                              20.8 &                                           197 &                                          1454 &                                           7.3 &                                           233 &                                          1511 &                                           3.2 &                                           254 &                                          1553 &                                           0.9 &                                             3.4 \\
    Void classifier \cite{Blum19fishyscapes} & \cmark &              36.8 &                    63.5 &                  44.3 &                  21.1 &                    22.1 &                    181 &                    797 &                    14.2 &                    219 &                    845 &                     7.5 &                    253 &                    879 &                     1.6 &                       7.6 \\
    Embedding density \cite{Blum19fishyscapes} & \xmark &              37.5 &                    70.8 &                  48.7 &                  33.8 &                    20.5 &                    107 &                   1437 &                    16.7 &                    176 &                   1485 &                     9.4 &                    250 &                   1592 &                     1.3 &                       9.2 \\
    Image resynthesis \cite{Lis19} & \xmark &              52.3 &                    25.9 &                  60.5 &                  39.5 &                    11.0 &                     95 &                   1187 &                    20.7 &                    153 &                   1225 &                    13.7 &                    230 &                   1294 &                     4.0 &                      12.9 \\
    SynBoost \cite{dibiase2021pixelwise} & \cmark &              56.4 &                    61.9 &                  58.0 &                  35.0 &                    18.3 &                    109 &                   1062 &                    20.7 &                    178 &                   1114 &                    11.5 &                    247 &                   1216 &                     2.0 &                      11.5 \\
    Maximized entropy \cite{chan2020entropy} & \cmark  &              \textbf{85.5} &                    \textbf{15.0} &                  \textbf{77.4} &                  \textbf{49.2} &                    \textbf{39.5} &                     \textbf{85} &                    \textbf{413} &                    \textbf{41.5} &                    \textbf{115} &                    \textbf{421} &                    \textbf{35.4} &                    \textbf{163} &                    \textbf{439} &                    \textbf{24.8} &                      \textbf{34.5} \\
    \bottomrule
    \end{tabular}
    }
    \end{center}
    \caption{Benchmark results for our RoadAnomaly21 dataset. This dataset contains 262 ground-truth components in total. The main performance metrics are highlighted with gray columns.}
    \label{tab:table1}
\end{table*}

\begin{table*}[t]
    \setlength{\tabcolsep}{2pt}
    \begin{center}
    \scalebox{0.66}{
    \begin{tabular}{l||c||acc||cc|ccc|ccc|ccc|g}
    \toprule
    & & \multicolumn{3}{c||}{Pixel-level} & \multicolumn{12}{c}{Component-level}  \\
    \midrule
    \midrule
    & requires & \multicolumn{3}{c||}{Anomaly (obstacle) scores} & $k\in\mathcal{K}$ & $\hat{k}\in\mathcal{\hat{K}}$ & \multicolumn{3}{c|}{$\tau=0.25$} & \multicolumn{3}{c|}{$\tau=0.50$} & \multicolumn{3}{c|}{$\tau=0.75$} & \multicolumn{1}{c}{} \\
    Method & OoD data & \multicolumn{1}{c}{AuPRC $\uparrow$} & FPR$_{95}$ $\downarrow$ & $F_1^*\uparrow$ & $\overline{\mathrm{sIoU}}$ $\uparrow$ & $\overline{\mathrm{PPV}}$ $\uparrow$ & FN $\downarrow$ & FP $\downarrow$ & $F_1\uparrow$ & FN $\downarrow$ & FP $\downarrow$ & $F_1\uparrow$ & FN $\downarrow$ & FP $\downarrow$ & $F_1\uparrow$ & \multicolumn{1}{c}{$\overline{F_1}\uparrow$}\\
    \midrule
    Maximum softmax \cite{Hendrycks2017msp} & \xmark & 15.7 & 16.6 & 22.5 & 19.7 & 15.9 & 255 & 1494 & 13.2 & 326 & 1503 & 6.3 & 372 & 1517 & 1.7 & 6.9 \\
    ODIN \cite{liang18odin} & \xmark & 21.2 & 15.4 & 29.2 & 20.7 & 18.5 & 260 & 1072 & 16.1 & 312 & 1079 & 9.9 & 362 & 1093 & 3.5 & 10.0 \\
    Mahalanobis \cite{Lee2018mahala} & \xmark & 20.9 & 13.1 & 25.8 & 14.0 & 21.8 & 293 & 1101 & 12.0 & 352 & 1104 & 4.7 & 385 & 1116 & 0.4 & 5.5 \\
    MC dropout \cite{mukhoti2019bdl} & \xmark & 3.7 & 50.6 & 8.0 & 6.3 & 5.8 & 351 & 2782 & 2.3 & 375 & 2784 & 0.8 & 386 & 2790 & 0.1 & 1.0 \\
    Ensemble \cite{Lakshminarayanan17} & \xmark & 1.1 & 77.2 & 3.1 & 8.6 & 4.7 & 335 & 3758 & 2.5 & 365 & 3768 & 1.1 & 382 & 3782 & 0.3 & 1.3 \\
    Void classifier \cite{Blum19fishyscapes} & \cmark & 9.2 & 41.5 & 23.4 & 6.3 & 20.3 & 350 & 350 & 9.8 & 365 & 350 & 6.0 & 381 & 353 & 1.9 & 5.9 \\ 
    Embedding density \cite{Blum19fishyscapes} & \xmark &  0.8 & 46.4 & 2.0 & 35.6 & 2.9 & 145 & 10972 & 4.2 & 244 & 11037 & 2.5 & 370 & 11191 & 0.3 & 2.4 \\
    Image resynthesis \cite{Lis19} & \xmark & 37.2 & 4.7 & 38.8 & 16.6 & 20.5 & 286 & 743 & 16.5 & 334 & 773 & 8.9 & 374 & 824 & 2.3 & 9.5 \\
    Road inpainting \cite{lis2020detecting} & \xmark &  52.6 & 47.1 & 67.5 & \textbf{57.6} & 39.5 & \textbf{79} & 580 & 48.4 & \textbf{131} & 586 & 41.8 & \textbf{240} & 611 & 25.8 & 40.2 \\
    SynBoost \cite{dibiase2021pixelwise} & \cmark & 70.3 & 3.1 & 70.1 & 44.3 & 41.8 & 133 & 352 & 51.3 & 185 & 363 & 42.6 & 286 & 414 & 22.6 & 40.4 \\
    Maximized entropy \cite{chan2020entropy} & \cmark &  \textbf{85.1} & \textbf{0.8} & \textbf{79.6} & 47.9 & \textbf{62.6} & 136 & \textbf{151} & \textbf{63.7} & 177 & \textbf{158} & \textbf{55.7} & 247 & \textbf{174} & \textbf{40.1} & \textbf{54.2}   \\
    \bottomrule
    \end{tabular}
    }
    \end{center}
    \caption{Benchmark results for our RoadObstacle21 dataset. This dataset contains 388 ground-truth components in total. The main performance metrics are highlighted with gray columns.}
    \label{tab:table2}
\end{table*}

\section{Numerical Experiments}
\begin{figure*}[t]
    \centering
    \captionsetup[subfigure]{labelformat=empty, position=top}
    \subfloat[Image \& annotation]{\includegraphics[width=0.199\textwidth]{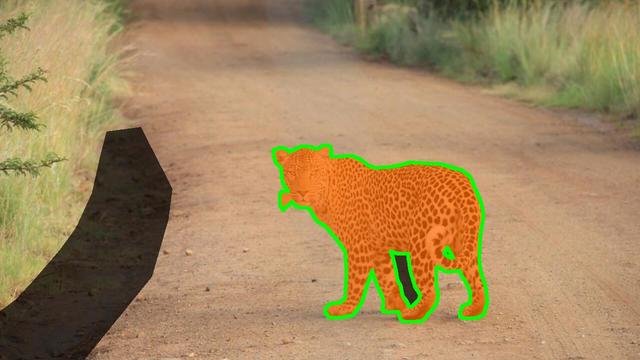}}
    \subfloat[Mahalanobis]{\includegraphics[width=0.199\textwidth]{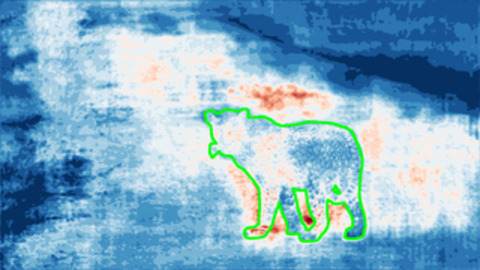}}
    \subfloat[MC dropout]{\includegraphics[width=0.199\textwidth]{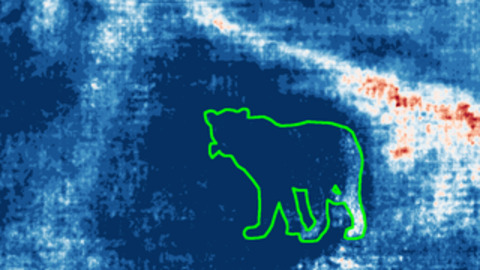}}
    \subfloat[SynBoost]{\includegraphics[width=0.199\textwidth]{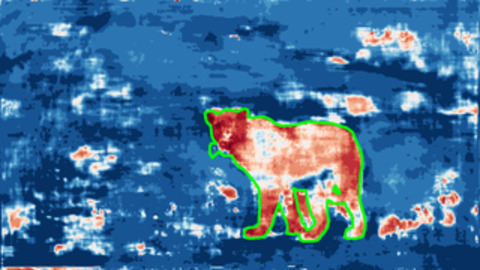}}
    \subfloat[Maximized entropy]{\includegraphics[width=0.199\textwidth]{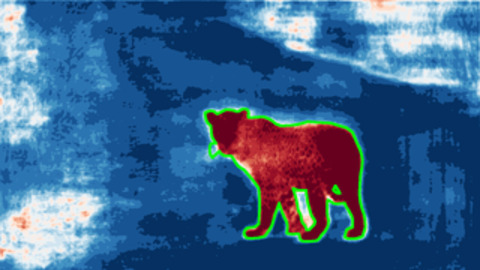}}\\
    \vspace{-1.07\baselineskip}
    \captionsetup[subfigure]{labelformat=empty, position=bottom}
    \subfloat[]{\includegraphics[width=0.199\textwidth]{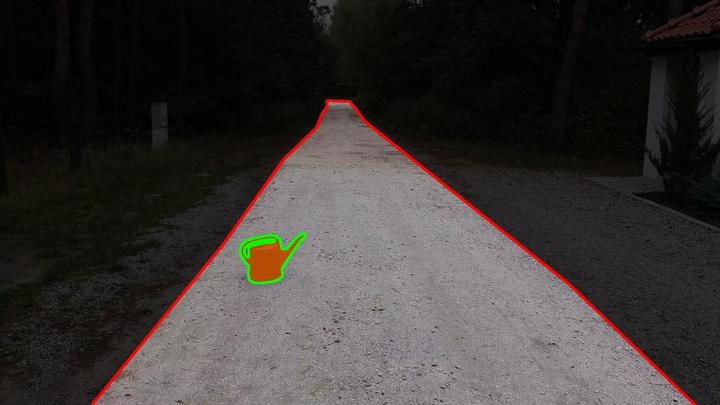}}
    \subfloat[]{\includegraphics[width=0.199\textwidth]{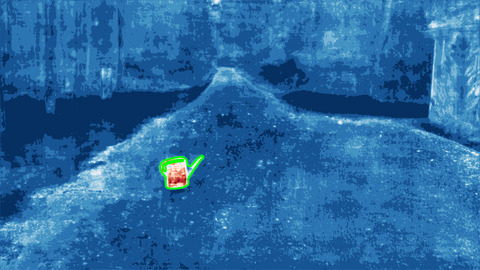}}
    \subfloat[]{\includegraphics[width=0.199\textwidth]{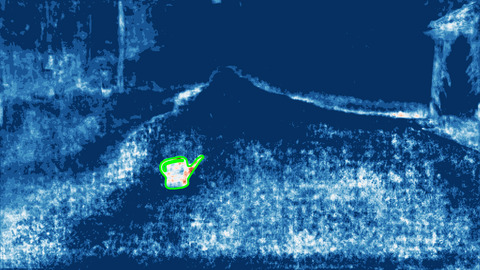}}
    \subfloat[]{\includegraphics[width=0.199\textwidth]{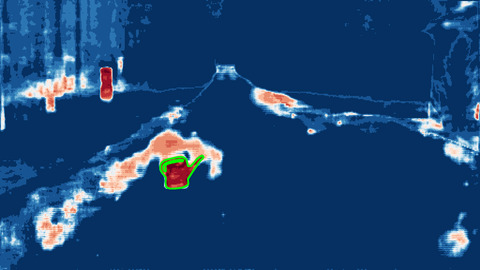}}
    \subfloat[]{\includegraphics[width=0.199\textwidth]{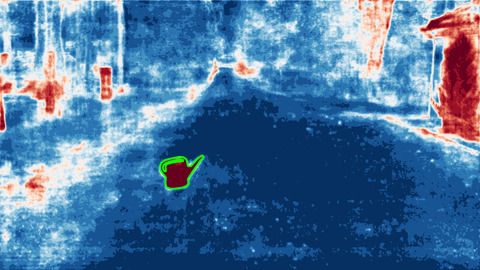}}
    \vspace{-5mm}
    \caption{Qualitative comparison of the anomaly scores produced by the methods introduced in \Cref{sec:methods} for one example image of RoadAnomaly21 (top row) and one example image of RoadObstacle21 (bottom row). Here, red indicates higher anomaly / obstacle scores and blue lower ones. The ground-truth anomaly / obstacle component is indicated by green contours.
    }
    \label{fig:obstacle-comparison}
\end{figure*}

\begin{table*}[t]
    \setlength{\tabcolsep}{2pt}
    \begin{center}
    \scalebox{0.66}{
    \begin{tabular}{l|c|ag|ac|ccg|ag|ac|cc|g}
    \toprule
    & & \multicolumn{2}{c|}{RoadAnomaly} & \multicolumn{5}{c|}{Fishyscapes LostAndFound Validation} & \multicolumn{2}{c|}{RoadObstacle} & \multicolumn{5}{c}{LostAndFound test-NoKnown}\\
    \midrule
    & & \multicolumn{2}{c|}{} & \multicolumn{2}{c|}{Pixel-level} & \multicolumn{3}{c|}{Component-level} & \multicolumn{2}{c|}{} & \multicolumn{2}{c|}{Pixel-level} & \multicolumn{3}{c}{Component-level} \\
    \midrule
    \midrule
    & OoD & \multicolumn{2}{c|}{} & \multicolumn{2}{c|}{Anomaly scores} & $k\in\mathcal{K}$ & $\hat{k}\in\mathcal{\hat{K}}$ & \multicolumn{1}{c|}{} & \multicolumn{2}{c|}{} & \multicolumn{2}{c|}{Anomaly scores} & $k\in\mathcal{K}$ & $\hat{k}\in\mathcal{\hat{K}}$ & \multicolumn{1}{|c}{} \\
    Method & data & \multicolumn{1}{c}{AuPRC $\uparrow$} & \multicolumn{1}{c|}{$\overline{F_1} \uparrow$} & \multicolumn{1}{c}{AuPRC$\uparrow$}  & FPR$_{95}$ $\downarrow$ & $\overline{\mathrm{sIoU}}$ $\uparrow$ & $\overline{\mathrm{PPV}}$ $\uparrow$ & \multicolumn{1}{c|}{$\overline{F_1}\uparrow$} & \multicolumn{1}{c}{AuPRC $\uparrow$} & \multicolumn{1}{c|}{$\overline{F_1} \uparrow$} & \multicolumn{1}{c}{AuPRC $\uparrow$}  & FPR$_{95}$ $\downarrow$ & $\overline{\mathrm{sIoU}}$ $\uparrow$ & $\overline{\mathrm{PPV}}$ $\uparrow$ & \multicolumn{1}{|c}{$\overline{F_1}\uparrow$}\\
    \midrule
    Maximum softmax \cite{Hendrycks2017msp} & \xmark & 28.0 & 5.9 & 5.6 & 40.5 & 3.5 & 9.5 & 1.8 & 15.7 &  6.9 & 30.1 & 33.2 & 14.2 & 62.2 & 13.4 \\
    ODIN \cite{liang18odin} & \xmark  & 33.1 & 6.0 & 15.5 & 38.4 & 9.9 & 21.9 & 9.7 & 21.2 & 10.0 & 51.0 & 30.7 & 38.9 & 48.0 & 38.1 \\
    Mahalanobis \cite{Lee2018mahala} & \xmark & 20.0 & 2.9 & 32.9 & \textbf{8.7} & 19.6 & 29.4 & 19.2 & 20.9 & 5.5 & 55.0 & 12.9 & 33.8 & 31.7 & 24.6 \\
    MC dropout \cite{mukhoti2019bdl} & \xmark & 28.9 & 4.9 & 14.4 & 47.8 & 4.8 & 18.1 & 4.3 & 3.7 & 1.0 & 36.2 & 36.0 & 17.0 & 34.7 & 14.7 \\
    Ensemble \cite{Lakshminarayanan17} & \xmark & 17.7 & 3.4 & 0.3  & 90.4  & 3.1 & 1.1 & 0.4 & 1.1 & 1.3 & 2.9 & 82.0 & 6.7 & 7.6 & 2.7  \\
    Void classifier \cite{Blum19fishyscapes} & \cmark & 36.8 & 7.6 & 11.7 & 15.3 & 9.2 & 39.1 & 14.9 & 9.2 & 5.9 & 4.4 & 47.0 & 0.7 & 35.1 & 1.1 \\ 
    Embedding density \cite{Blum19fishyscapes} & \xmark & 37.5 & 9.2 & 8.9 & 42.2 & 5.9 & 10.8 & 4.9 & 0.8 & 2.4 & 61.7 & 10.4 & 37.8 & 35.2 & 30.8 \\
    Image resynthesis \cite{Lis19} & \xmark & 52.3 & 12.9 & 5.1 & 29.8 & 5.1 & 12.6 & 4.1 & 37.2 & 9.5 & 57.1 & 8.8 & 27.2 & 30.7 & 21.5 \\
    Road inpainting \cite{lis2020detecting} & \xmark & - & - &
    - & - & - & - & - & 52.6 & 40.2 & \textbf{83.0} & 35.7 & \textbf{49.2} & 60.7 & \textbf{56.9} \\
    SynBoost \cite{dibiase2021pixelwise} & \cmark & 56.4 & 11.5 & \textbf{64.9} & 30.9 & \textbf{27.9} & \textbf{48.6} & \textbf{38.0} & 70.3 & 40.4 & 81.8 & \textbf{4.6} & 37.2 & \textbf{72.3} & 53.0\\
    Maximized entropy \cite{chan2020entropy} & \cmark & \textbf{85.5} & \textbf{34.5} & 44.3 & 37.7 & 21.1 & \textbf{48.6} & 30.0 & \textbf{85.1} & \textbf{54.2} & 77.9 & 9.7 & 45.9 & 63.1 & 55.0\\
    \bottomrule
    \end{tabular}
    }
    \end{center}
    \caption{
    Benchmark results for Fishyscapes LostAndFound validation and LostAndFound test-NoKnown, containing 165 and 1709 ground-truth components in total, respectively. In this table the pixel-wise AuPRC and the component-wise $\overline{F_1}$ from RoadAnomaly21 and RoadObstacle21, \cf \Cref{tab:table1} and \Cref{tab:table2}, are additionally included for cross evaluation (gray columns).
    }
    \label{tab:LAF}
\end{table*}

In our benchmark suite we integrate a default method to generate the anomaly segmentation from pixel-wise anomaly scores. We choose the threshold $\delta^*$, at which one pixel is classified as anomaly, by means of the optimal pixel-wise $F_1$-score, that we denote with $F_1^*$. Then, $\delta^*$ is computed as 
\begin{equation} \label{eq:optim-f1-pixel}
    \delta^* = \argmax\limits_{\delta\in\mathbb{R}} ~2\cdot\textrm{precision}(\delta)\cdot\textrm{recall}(\delta) ~ / ~ (\textrm{precision}(\delta)+\textrm{recall}(\delta))\;,
\end{equation}
subject to $\textrm{precision}(\delta)+\textrm{recall}(\delta) > 0 ~ \forall ~ \delta$. In \Cref{sec:param-study-supp} we provide a study where $\delta^*$ is varied.

Moreover, for the anomaly track, components smaller than 500 pixels are discarded from the predicted segmentation, and for the obstacle track, components smaller than 50 pixels are discarded. These sizes are chosen based on the smallest ground-truth components. All methods presented in \Cref{sec:methods} produce anomaly scores for which we apply the default segmentation method. We emphasize that using our proposed default method for anomaly segmentation masks is completely optional. We provide results without filtering by predicted component sizes in \Cref{sec:param-study-supp}. We allow and encourage competitors in the benchmark to submit their own anomaly segmentation masks generated via more sophisticated image operations.

In our evaluation, we additionally include the average sIoU per component $\overline{\mathrm{sIoU}}$, which can be computed by averaging sIoU over all ground-truth components $k\in\mathcal{K}$.
Analogously, we also include the average PPV per component $\overline{\mathrm{PPV}}$ for all predicted components $\hat{k}\in\mathcal{\hat{K}}$.
As the number of component-wise TP, FN and FP depends on some threshold $\tau$ for sIoU and PPV, respectively (see \Cref{sec:metrics}), we average these quantities over different thresholds $\tau \in \mathcal{T}=\{ 0.25,0.30,\ldots, 0.75\}$, similarly to \cite{Lin14COCO}, yielding the averaged component-wise $F_1$ score $\overline{F_1}=\frac{1}{|\mathcal{T}|} \sum_{\tau \in \mathcal{T}} F_1(\tau)$.

\paragraph{Discussion of the Results.} \label{sec:discussion}
Our benchmark results for RoadAnomaly21 and RoadObstacle21 are summarized in \Cref{tab:table1} and \Cref{tab:table2}, respectively. In general, we observe that methods originally designed for image classification, including maximum softmax, ODIN, and Mahalanobis, do not generalize well to anomaly and obstacle segmentation. For methods based on statistics of the Cityscapes dataset, such as Mahalanobis as well as learned embedding density, anomaly detection is typically degraded by the presence of a domain shift. This results in a poor performance, particularly in RoadObstacle21, where various road surfaces can be observed. Interestingly, learned embedding density, MC dropout and the void classifier yield worse performance than maximum softmax on RoadObstacle21, whereas we observe the opposite on RoadAnomaly21. 

The detection methods based on generative models, namely image resynthesis and SynBoost, appear to be better suited to both anomaly and obstacle segmentation at pixel as well as component level, clearly being superior to all the approaches discussed previously. This observation also holds for road inpainting in the obstacle track. These autoencoder-based methods are nonetheless limited by their discrepancy module, and they are outperformed in our experiments by maximized softmax entropy, which peaks at an AuPRC of 86\% and a component-wise $\overline{F_1}$ of 49\%. This highlights the importance of anomaly and obstacle proxy data. Illustrative example score maps produced by the discussed methods are shown in \Cref{fig:obstacle-comparison}.

In summary, the component-level evaluation highlights the methods' weaknesses even more clearly than the pixel-wise evaluation, the latter giving a stronger weight to larger anomalies and obstacles. All methods indeed tend to face difficulties in the presence of smaller anomalies and obstacles, as we demonstrate in more detail in \Cref{sec:eval-comp-size-supp}. In addition, we observe a much lower component-wise $\overline{F_1}$ score than a pixel-wise $F_1^*$, demonstrating the importance of evaluating at component level. The results \wrt the different categories of methods are challenging for models, hence leaving room for improvement.

Our benchmark suite enables a unified evaluation across different datasets whenever ground truth is available. 
In \Cref{tab:LAF} we summarize our results for Fishyscapes LostAndFound \cite{Blum19fishyscapes}, a validation set of 100 LostAndFound images \cite{Pinggera16LostAndFound} with refined labels fitting the anomaly track, and the LostAndFound test split, with original labels fitting the obstacle track.
Note that, for the LostAndFound test split, we filtered out all images that contain humans and bicycles labeled as obstacles (therefore called LostAndFound test-NoKnown) because we applied anomaly segmentation methods out of the box to the task of obstacle segmentation, and these methods focus on previously-unseen objects. 

In comparison to our datasets, for both LostAndFound datasets we observe a less pronounced gap, in terms of both main performance metrics, the pixel-level AuPRC and component-level $\overline{F_1}$ scores, between the methods orignially designed for image classification, especially ODIN and Mahalanobis, and those specifically designed for anomaly segmentation, especially road inpainting and maximized entropy. This signals that both of our datasets contribute new challenges for anomaly and obstacle segmentation.
In \Cref{sec:eval-comp-size-supp} and \Cref{sec:eval-cate-supp} we provide further and more fragmented results in terms of both objects sizes and object categories.

Finally, we also applied our benchmark suite to the LiDAR guided Small obstacle Segmentation dataset~\cite{Singh2020-fw}. Our main findings are that our whole set of methods yields weak performance on that dataset.
The main purpose of this dataset is the detection of small obstacles from multiple sensors including LiDAR. Hence, the conditions for the other sensor modalities are purposely challenging (\eg, low illumination), making this dataset less suitable to camera-only methods. We present the corresponding results in \Cref{sec:eval-small-obstacle}.

\section{Conclusion}

In this work, we have introduced a unified and publicly available benchmark suite that evaluates a method's performance for anomaly segmentation with established pixel level as well as recent component level metrics. Our benchmark suite is applicable in a plug and play fashion to any dataset for anomaly segmentation that comes with ground truth, such as LostAndFound and Fishyscapes LostAndFound, allowing for a better comparison of new methods. Moreover, our benchmark is accompanied with two publicly available datasets, RoadAnomaly21 for anomaly segmentation and RoadObstacle21 for obstacle segmentation. 

These two datasets challenge two important abilities of computer vision systems: On one hand their ability to detect and localize unknown objects; on the other hand their ability to reliably detect and localize obstacles on the road, may they be known or unknown. Our datasets consist of real images with pixel-level annotations and depict street scenes with higher variability in object types and object sizes than existing datasets. Our experiments have demonstrated that both of our datasets show a distinct separation in terms of performance between the methods that are specifically designed for anomaly / obstacle segmentation and those that are not. However, there remains much room for performance improvement, particularly in terms of component-wise metrics, which stresses the need for future research in the direction of anomaly segmentation. 

The images of the datasets and the software are available at \href{https://www.segmentmeifyoucan.com/}{{https://www.segmentmeifyoucan.com/}}.

\section*{Broader Impact}
This benchmark advances research towards the safe deployment of autonomous vehicles. This ultimately will have many consequences, \eg, reducing the number of jobs in the transport sector. More immediately, the benchmark measures the reliability of algorithms and therefore may be misunderstood as giving safety guarantees. This benchmark however only works for the specified training regime \ie it cannot certify fitness for real-world deployment and should not be misunderstood as such. In particular, while our datasets greatly contribute to the diversity of anomalies, the scale of the datasets is still not even close to sufficient in order to represent every possible type of an anomaly. Furthermore, although we do not publicly provide test labels, there remains a risk, common to any other benchmark, of the community designing methods that overfit on our benchmark tasks.

\section*{Acknowledgement}
Robin Chan and Svenja Uhlemeyer acknowledge funding by the German Federal Ministry for Economic Affairs and Energy, within the projects ``KI Absicherung - Safe AI for Automated Driving'', grant no.\ 19A19005R, and ``KI Delta Learning - Scalable AI for Automated Driving'', grant no.\ 19A19013Q, respectively. We thank the consortiums for the successful cooperation.
We would also like to thank the ``BUW-KI'' team who substantially contributed to collecting and labeling of data.

\newpage
\printbibliography


\newpage
\section*{Appendix}
\appendix

\section{NeurIPS Questionaire}

\begin{enumerate}

\item Submission introducing new datasets must include the following in the supplementary materials:
\begin{enumerate}
  \item Dataset documentation and intended uses. Recommended documentation frameworks include datasheets for datasets, dataset nutrition labels, data statements for NLP, and accountability frameworks.\\
  \answerYes{We provide the complete `datasheet for datasets' in Appendix~\ref{sec:datasheet}.}
  \item URL to website/platform where the dataset/benchmark can be viewed and downloaded by the reviewers.\\
  \answerYes{\url{https://segmentmeifyoucan.com/}}
  \item Author statement that they bear all responsibility in case of violation of rights, etc., and confirmation of the data license. \\ \answerYes{All authors bear responsibility in case of violation of rights, etc. Confirmation of the data license is given the repository items of \url{https://zenodo.org/communities/segmentmeifyoucan}.}
  \item Hosting, licensing, and maintenance plan. The choice of hosting platform is yours, as long as you ensure access to the data (possibly through a curated interface) and will provide the necessary maintenance.\\
  \answerYes{To ensure good availability, we chose professionally maintained platforms. Data is hosted at the public data repository \url{zenodo.com} and the benchmark website is hosted through \url{github.com}. Necessary maintenance such as updating the benchmark record etc. is shared between 3 different research groups such that there is always at least one person reachable.}
\end{enumerate}

\item To ensure accessibility, the supplementary materials for datasets must include the following:
\begin{enumerate}
  \item Links to access the dataset and its metadata. This can be hidden upon submission if the dataset is not yet publicly available but must be added in the camera-ready version. In select cases, e.g when the data can only be released at a later date, this can be added afterward. Simulation environments should link to (open source) code repositories.\\
  \answerYes{In general, all data is listed on \url{https://segmentmeifyoucan.com/}, and metadata more specifically in the zenodo mirrors: \url{https://zenodo.org/communities/segmentmeifyoucan}}
  \item The dataset itself should ideally use an open and widely used data format. Provide a detailed explanation on how the dataset can be read. For simulation environments, use existing frameworks or explain how they can be used.\\
  \answerYes{The data is stored in standard formats: png, webp, json. We provide ready-to-use code that reads the data at \url{https://github.com/SegmentMeIfYouCan/road-anomaly-benchmark}.}
  \item Long-term preservation: It must be clear that the dataset will be available for a long time, either by uploading to a data repository or by explaining how the authors themselves will ensure this.\\
  \answerYes{The data is uploaded to multiple mirrors, one of them is the public data repository \url{zenodo.org}.}
  \item Explicit license: Authors must choose a license, ideally a CC license for datasets, or an open source license for code (e.g. RL environments).\\
  \answerYes{All images in the obstacle track were recorded by the authors of this work and are published under CC-BY 4.0 license. The images of the anomaly track are all publicly available and licensed as one of \{public domain, CC-BY, CC-BY-SA\}. A complete list of images, licenses and creators is published as part of the data record: \url{https://zenodo.org/record/5185336}.}
  \item Add structured metadata to a dataset's meta-data page using Web standards (like schema.org and DCAT): This allows it to be discovered and organized by anyone. If you use an existing data repository, this is often done automatically.\\
  \answerYes{Metadata is part of the records on zenodo and accessible via different APIs, e.g. \url{https://zenodo.org/oai2d?verb=ListRecords&set=user-segmentmeifyoucan&metadataPrefix=oai_dc}.}
  \item Highly recommended: a persistent dereferenceable identifier (e.g. a DOI minted by a data repository or a prefix on identifiers.org) for datasets, or a code repository (e.g. GitHub, GitLab,...) for code. If this is not possible or useful, please explain why.\\
  \begin{itemize}
      \item anomaly track data \url{https://doi.org/10.5281/zenodo.5185335}
      \item obstacle track data \url{https://doi.org/10.5281/zenodo.5186546}
      \item code repository is on GitHub \url{https://github.com/SegmentMeIfYouCan/road-anomaly-benchmark}
  \end{itemize}
\end{enumerate}

\item For benchmarks, the supplementary materials must ensure that all results are easily reproducible. Where possible, use a reproducibility framework such as the ML reproducibility checklist, or otherwise guarantee that all results can be easily reproduced, i.e. all necessary datasets, code, and evaluation procedures must be accessible and documented.\\
\answerYes{While, as a public benchmark, we do not give access to the test labels and therefore nobody else should be able to produce the same measurements, we document all code that is used to create the benchmark results (directly yielding the \textit{results.json} that is used for updating the public leaderboard on the website). Further, we created a small validation datasets that allows researchers to check that their method runs as intended. For these validation datasets, we report results in \cref{tab:table-val-anomaly} and \cref{tab:table-val-obstacle} which can be reproduced with the set of methods included in our benchmark suite.}

\item For papers introducing best practices in creating or curating datasets and benchmarks, the above supplementary materials are not required. \answerNA{}
\end{enumerate}

\section{Datasheet for Datasets}
\label{sec:datasheet}
The following section is a complete answer to the datasheet questions from~\cite{Gebru2018-bh}.

\subsection{Motivation}

\begin{itemize}
    \item \textbf{For what purpose was the dataset created?} To evaluate and compare anomaly segmentation methods in driving  scenes. Such evaluation enables conclusions on how good methods, usually tested on simpler datasets, are, but also facilitates specific method development for autonomous driving.
    \item \textbf{Who created the dataset (e.g., which team, research group) and on
behalf of which entity (e.g., company, institution, organization)?} The authors of this work created this dataset to find answers to their research questions. In particular, there was no external party ordering or suggesting the creation of such a benchmark.
    \item \textbf{Who funded the creation of the dataset?} See section Acknowledgements.
    \item \textbf{Any other comments?} No.
\end{itemize}

\subsection{Composition}

\begin{itemize}
    \item \textbf{What do the instances that comprise the dataset represent (e.g.,
documents, photos, people, countries)?} The dataset comprise high resolution images of street scenes with unusual objects, which are all annotated on pixel-level. The objects appearing in the anomaly track data were \textbf{not} placed artificially and therefore represent naturally occurring anomalies in a global context. For the obstacle track, the objects were selected and placed by the authors, choosing from available objects that can reasonably appear on a street.
    \item \textbf{How many instances are there in total (of each type, if appropriate)?} 100 images for the anomaly track containing 262 ground truth components (+ 10 images for validation), 327 for the obstacle track containing 388 ground truth components (+ 85 images with hard weather or lightning conditions, + 30 images for validation). 
    \item \textbf{Does the dataset contain all possible instances or is it a sample
(not necessarily random) of instances from a larger set?} The set of possibly occurring anomalies in driving scenes is boundless. The instances in this dataset are therefore a subset. For the anomaly track, they are a random sample of openly licensed, available images on the web. Therefore, they have a good geographic coverage. For the obstacle track, all images were taken in Switzerland and Germany. They have a good coverage over weather and seasons, but are highly biased to European context for both the street background and the selected objects.
    \item \textbf{What data does each instance consist of?} Each data point is an RGB image and a corresponding segmentation map.
    \item \textbf{Is there a label or target associated with each instance?} Yes, our labelling policy is described in Section~\ref{sec:datasets}.
    \item \textbf{Is any information missing from individual instances?} No.
    \item  \textbf{Are relationships between individual instances made explicit
(e.g., users’ movie ratings, social network links)?} No.
    \item \textbf{Are there recommended data splits (e.g., training, development/validation,
testing)?} Yes, our data is supposed to be used for testing only and should not be used for training. We supply a small validation split that enables local testing before submission to the benchmark.
    \item \textbf{Are there any errors, sources of noise, or redundancies in the
dataset?} The annotations were created by humans and can therefore contain errors.
\item \textbf{Is the dataset self-contained, or does it link to or otherwise rely on
external resources (e.g., websites, tweets, other datasets)?} The dataset is self-contained.
\item \textbf{Does the dataset contain data that might be considered confidential (e.g., data that is protected by legal privilege or by doctorpatient confidentiality, data that includes the content of individuals’ non-public communications)?} No. All used images are licensed to be shared  publicly.
\item \textbf{Does the dataset contain data that, if viewed directly, might be offensive, insulting, threatening, or might otherwise cause anxiety?} No.
\item \textbf{Does the dataset relate to people?} People appear in some images of the Anomaly track.
\item \textbf{Does the dataset identify any subpopulations (e.g., by age, gender)?} No.
\item \textbf{Is it possible to identify individuals (i.e., one or more natural persons), either directly or indirectly (i.e., in combination with other
data) from the dataset?} It it possible to match faces in the dataset to any other database. However, this applies only to the images of the Anomaly track where all images used were already public, so our dataset did not change that. Regarding the images of Obstacle track identifying individuals is \textbf{not} possible.
\item \textbf{Does the dataset contain data that might be considered sensitive
in any way (e.g., data that reveals racial or ethnic origins, sexual
orientations, religious beliefs, political opinions or union memberships, or locations; financial or health data; biometric or genetic data; forms of government identification, such as social security numbers; criminal history)?} No.
\item \textbf{Any other comments?} No.
\end{itemize}

\subsection{Collection Process}

\begin{itemize}
    \item \textbf{How was the data associated with each instance acquired?} In the obstacle track, the images were taken by the authors. For the anomaly track, openly licensed images from the web were collected. All images were annotated by humans. 
    \item \textbf{What mechanisms or procedures were used to collect the data
(e.g., hardware apparatus or sensor, manual human curation, software program, software API)?} Manual human curation.
\item \textbf{If the dataset is a sample from a larger set, what was the sampling
strategy (e.g., deterministic, probabilistic with specific sampling
probabilities)?} Images in the anomaly track are a random sample of openly licensed, available images from the web, that show street scenes including at least one anomaly and are of sufficiently high quality. In the obstacle track, some images were extracted from sequences and only images at certain distances (at a rough guess) were included.
\item \textbf{Who was involved in the data collection process (e.g., students, crowdworkers, contractors) and how were they compensated (e.g.,
how much were crowdworkers paid)?} The authors and student/research assistants. Everyone involved in the data generation process was employed at a university at the time of collecting and therefore drew a regular salary. 
\item \textbf{Over what timeframe was the data collected?} The images of the Anomaly Track were collected between August 2019 and August 2021. The images of the Obstacle track were collected between August 2020 and August 2021.
\item \textbf{Were any ethical review processes conducted (e.g., by an institutional review board)?} No.
\item \textbf{Does the dataset relate to people?} People appear in some images of the Anomaly track.
\item \textbf{Did you collect the data from the individuals in question directly,
or obtain it via third parties or other sources (e.g., websites)?} We obtained this data via third parties or other sources.
\item \textbf{Were the individuals in question notified about the data collection?} The collected images were already licensed as public domain or creative commons, i.e., licensed to be shared and used.
\item \textbf{Did the individuals in question consent to the collection and use
of their data?} As we used images from the public domain or licensed a creative commons, we did not ask for consent ourselves.
\item \textbf{Has an analysis of the potential impact of the dataset and its use on data subjects (e.g., a data protection impact analysis)been conducted?} Not beyond the Broader Impact section.
\item \textbf{Any other comments?} No.
\end{itemize}

\subsection{Preprocessing/cleaning/labeling}

\begin{itemize}
    \item \textbf{Was any preprocessing/cleaning/labeling of the data done (e.g.,
discretization or bucketing, tokenization, part-of-speech tagging,
SIFT feature extraction, removal of instances, processing of missing values)?} The images for the anomaly track were resized and cropped to two different resolutions (1280$\times$720 and 2048$\times$1024).
\item \textbf{Was the “raw” data saved in addition to the preprocessed/cleaned/labeled
data (e.g., to support unanticipated future uses)?} No.
\item \textbf{Is the software used to preprocess/clean/label the instances available?} The open-source software ImageMagick was used for resizing the images. As labeling tool, LabelMe was used, which is publicly available (\url{https://github.com/wkentaro/labelme}).
\item \textbf{Any other comments?} No.
\end{itemize}

\subsection{Uses}

\begin{itemize}
    \item \textbf{Has the dataset been used for any tasks already?} Yes, for this paper.
    \item \textbf{Is there a repository that links to any or all papers or systems that
use the dataset?} Yes, the public leaderboard on \url{https://segmentmeifyoucan.com/leaderboard}.
\item \textbf{What (other) tasks could the dataset be used for?} No other task, since the labels are hidden.
\item \textbf{Is there anything about the composition of the dataset or the way
it was collected and preprocessed/cleaned/labeled that might impact future uses?} No.
\item \textbf{Are there tasks for which the dataset should not be used?} Certification of fittness for deployment would require at least a larger dataset.
\item \textbf{Any other comments?} No.
\end{itemize}

\subsection{Distribution}
\begin{itemize}
    \item \textbf{Will the dataset be distributed to third parties outside of the entity (e.g., company, institution, organization) on behalf of which
the dataset was created? } Yes, the images including the labels for the validation set are public. The labels of the test set however will not be distributed.
\item  \textbf{How will the dataset will be distributed (e.g., tarball on website,
API, GitHub)?} Via multiple mirrors as zip archives, all listed on the website \url{https://segmentmeifyoucan.com/datasets}.
\item \textbf{When will the dataset be distributed?} Now.
\item \textbf{Will the dataset be distributed under a copyright or other intellectual property (IP) license, and/or under applicable terms of use
(ToU)?} All parts that are distributed are under public domain or creative commons licenses.
\item \textbf{Have any third parties imposed IP-based or other restrictions on
the data associated with the instances?} No.
\item \textbf{Do any export controls or other regulatory restrictions apply to
the dataset or to individual instances?} No.
\item \textbf{Any other comments?} No.
\end{itemize}

\subsection{Maintenance}
\begin{itemize}
    \item \textbf{Who is supporting/hosting/maintaining the dataset?} The authors of this paper.
    \item  \textbf{How can the owner/curator/manager of the dataset be contacted
(e.g., email address)?} \texttt{blumh@ethz.ch},
\item \textbf{Is there an erratum?} No.
\item \textbf{Will the dataset be updated (e.g., to correct labeling errors, add
new instances, delete instances)? } In case that corrections are necessary, all versions are tracked in the \url{zenodo.com} data items.
\item \textbf{If the dataset relates to people, are there applicable limits on the
retention of the data associated with the instances (e.g., were individuals in question told that their data would be retained for a
fixed period of time and then deleted)?} No. All images are licensed to be shared.
\item \textbf{Will older versions of the dataset continue to be supported/hosted/maintained?} In case that there would be multiple versions, only the newest will be maintained.
\item \textbf{If others want to extend/augment/build on/contribute to the
dataset, is there a mechanism for them to do so?} Yes, we already have plans to incorporate another body of data into the benchmark. Similarly to the two existing datasets, each set of data is treated as a separate instance, but made comparable by using the same metrics.
This can also be observed in this paper and the further comparisons that are listed  in the Appendix.
\item \textbf{Any other comments?} No.
\end{itemize}

\section{More Details on Evaluation Metrics}

\subsection{Pixel level}
Let $\mathcal{Z}$ denote the set of image pixel locations. A model with a binary classifier providing scores $s(x)\in\mathbb{R}^{|\mathcal{Z}|}$ for an image $x\in\mathcal{X}$ (from a dataset $\mathcal{X} \subseteq [0,1]^{N\times |\mathcal{Z}| \times 3}$ of $N$ images) discriminates between the two classes anomaly and non-anomaly. We evaluate the separability of the pixel-wise anomaly scores via the area under the precision-recall curve (AuPRC).

Let $\mathcal{Y}\subseteq \{\textrm{``anomaly''}, \textrm{``not anomaly''}\}^{N\times |\mathcal{Z}|}$ be the set of ground truth labels per pixel for $\mathcal{X}$. Analogously, we denote the predicted labels with $\hat{\mathcal{Y}}(\delta)$, obtained by pixel-wise thresholding on $s(x)~ \forall ~ x\in\mathcal{X}$ \wrt some threshold value $\delta\in\mathbb{R}$.
Then, for the anomaly class ($c_1=\textrm{``anomaly''}$) we compute
\begin{equation}
    \textrm{precision}(\delta) = \frac{| \mathcal{Y}_{c_1} \! \cap \mathcal{\hat{Y}}_{c_1}(\delta) |}{ |\mathcal{\hat{Y}}_{c_1}(\delta) |}, ~~
    \textrm{recall}(\delta) = \frac{| \mathcal{Y}_{c_1} \!\cap \mathcal{\hat{Y}}_{c_1}(\delta) |}{ |\mathcal{Y}_{c_1} |}
\end{equation}
with $\mathcal{Y}_{c_1}$ and $\hat{\mathcal{Y}}_{c_1}$ representing the ground truth labels and predicted labels, respectively. For the AuPRC, precision and recall are considered as functions of $\delta$. The AuPRC approximates $\int \textrm{precision}(\delta) \, d\textrm{recall}(\delta)$ and is threshold independent \cite{Boyd2013}. It also puts emphasis on detecting the minority class, making it particularly well suited as our main evaluation metric since the pixel-wise class distributions of RoadAnomaly21 and RoadObstacle21 are considerably unbalanced, \cf \cref{sec:datasets}.

To consider the safety point of view, we also include the false positive rate at 95\% true positive rate (FPR$_{95}$) in our evaluation, where the true positive rate (TPR) is equal to the recall of the anomaly class.
The false positive rate (FPR) is the number of pixels falsely predicted as anomaly over the number of all non-anomaly pixels. Hence, for the anomaly class we compute
\begin{equation}
    \textrm{FPR}_{95} = \frac{|\mathcal{\hat{Y}}_{c_1}(\delta^\prime) \cap \mathcal{Y}_{c_2}|}{|\mathcal{Y}_{c_2}|}  ~~~~\mathrm{s.t.}~~ \textrm{TPR}(\delta^\prime) = 0.95 \;,
\end{equation}
where $\textrm{c}_2=\textrm{``not anomaly''}$. The metric FPR$_{95}$ indicates how many false positive predictions are necessary to guarantee a desired true positive rate. Note that, any prediction which is contained in a ground truth labeled region of class void is not counted as false positive, \cf \cref{sec:datasets}. In particular for the RoadObstacle21 dataset the evaluation is therefore restricted to the road area.

\subsection{Component level - Qualitative examples revealing the difference of IoU and sIoU}  \label{sec:siou-example}

\begin{figure}[ht]
    \captionsetup[subfigure]{labelformat=empty}
    \centering
    \subfloat{\includegraphics[width=0.25\linewidth]{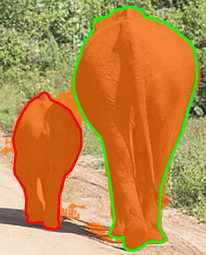}}~
    \subfloat{\includegraphics[width=0.25\linewidth]{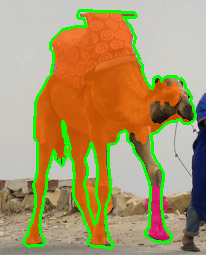}}
    \caption{Two examples underlining the difference between IoU and adjusted IoU (sIoU). The ground-truth components are indicated by green/red contours, predicted components are highlighted by other colors.
    \emph{Left}: Two ground-truth components (green \& red) intersect with one predicted component (orange). Green: \textbf{IoU 68.18\% vs. sIoU 87.01\%}; red: \textbf{IoU 21.68\% vs. sIoU 68.44\%}.
    \emph{Right}: Two predicted components (orange \& pink) intersect with one ground-truth component (green). Orange: \textbf{IoU 78.97\% vs. sIoU 81.69\%}; pink: \textbf{IoU 03.44\% vs. sIoU 18.91\%}.
    }
    \label{fig:siou}
\end{figure}

If we consider component-level metrics over ground-truth components, it may happen that several components are close together and therefore covered by one predicted component. Although the real error can be small, the IoU punishes both ground-truth components. The same holds the other way around when considering metrics over predicted components, \ie when one ground-truth component is covered by several predicted components.
A qualitative example is given in \cref{fig:siou}. 
A small number of incorrectly predicted pixels may cause a strong decrease in the IoU. The adjusted IoU (sIoU) is less sensitive in such cases.
sIoU focuses on correctly covering the regions of obstacles/anomalies in the image rather than finding such regions separately for each instance, as done by IoU. In self-driving it is more important to know the regions of anomaly rather than how many of them exist.

\section{Evaluated Methods} \label{sec:methods-supp}

In this section, we first briefly introduce the methods which are evaluated on our benchmark and constitute our initial leader board. Afterwards we additionally provide technical details to those introduced methods. 

\subsection{Brief Description of Methods}
All methods subject to evaluation are stated in boldface in the following. We evaluate at least one method per type discussed in \cref{sec:anomaly-seg}. All methods have an underlying semantic segmentation DNN trained on Cityscapes and they all provide pixel-wise anomaly scores. 

Given an input image, the \textbf{maximum softmax probability} (MSP) of a DNN's corresponding output is a commonly-used baseline for OoD detection at image level \cite{Hendrycks2017msp}. Adding small perturbations to every pixel of the input image and applying temperature scaling enhances the anomaly detection ability of MSP. The latter approach is known as \textbf{ODIN} \cite{liang18odin}. Another well-known method detects anomalies based on the \textbf{Mahalanobis distance}. It is computed by estimating Gaussian distributions of latent features of a DNN's penultimate layer, therefore yielding an estimate of the likelihood of a test sample \wrt the distribution in the training data. All these methods are originally designed for image classification but can be adapted straightforwardly to segmentation and represent good baselines in our benchmark.

As Bayesian approach to uncertainty estimation we employ \textbf{Monte Carlo (MC) dropout} in our evaluation. MC dropout has already been investigated for semantic segmentation. We follow \cite{mukhoti2019bdl} and use the mutual information as pixel-wise anomaly scores, which captures the epistemic uncertainty of a DNN. Furthermore, we additionally evaluate an \textbf{ensemble} of semantic segmentation networks.

In \cite{Blum19fishyscapes} several approaches to learning the confidence with respect to the presence of anomalies have been proposed. The \textbf{learned embedding density} aims to approximate the distribution of feature embeddings within a DNN via normalizing flows. At test time, the negative log-likelihood for each embedded representation of an image measures the discrepancy of a test embedding with respect to training embeddings, where high discrepancies indicate anomalies. These scores are then upsampled via bilinear interpolation to obtain the pixel-wise anomaly scores.
Alternatively, the segmentation DNN can be modified to learn the confidence for the presence of anomalies, requiring an OoD dataset. As in \cite{Blum19fishyscapes}, a Cityscapes DNN is trained with an additional model output for the Cityscapes void class. The anomaly scores are then the softmax scores for the that class, therefore this method is called \textbf{void classifier}. Additionally, one can also retrain a DNN with a different OoD proxy, such as the COCO dataset \cite{Lin14COCO}, and enforce \textbf{maximized softmax entropy} \cite{chan2020entropy} on samples of the OoD proxy. All theses methods tune previously-trained DNNs to the task of anomaly segmentation and are included in our evaluation.

As autoencoders in our evaluation, we employ \textbf{image resynthesis} together with a discrepancy network that extracts meaningful differences based on the information provided by the DNN's segmentation mask, the resynthesized input image and the original image itself \cite{Lis19}. This approach can be extended by including uncertainty estimates in the discrepancy module, aiming to boost the anomaly segmentation performance, known as \textbf{SynBoost} \cite{dibiase2021pixelwise}. One method specifically designed for obstacle segmentation is called \textbf{road inpainting} \cite{lis2020detecting}. This method inpaints road patches in a sliding window manner. The resulting synthesized image is then again presented to a discrepancy network, similarly as in \cite{Lis19}, for pixel-wise obstacle scores. 

\subsection{Method Description in Detail} \label{sec:methods-detail}
All methods provide pixel-wise anomaly scores $s(x) \in \mathbb{R}^{|\mathcal{Z}|}, x\in\mathcal{X}$ where anomalies correspond to higher values. As a reminder, $\mathcal{Z}$ denotes the set of image coordinates and $\mathcal{X} \subseteq[0,1]^{N\times|\mathcal{Z}|\times 3}$ a dataset with $N$ images. Below, we describe how $s$ is obtained for each approach.

\textbf{Maximum softmax probability.} Let $f: \mathcal{X} \to \mathbb{R}^{|\mathcal{Z}| \times |\mathcal{C}|}$ denote the output of a semantic segmentation DNN. The maximum softmax probability (MSP) is a commonly-used baseline for OoD detection at image level~\cite{Hendrycks2017msp}. It computes an anomaly score for each pixel $z\in\mathcal{Z}$ as
\begin{equation}
     s_z(x) = 1 - \max_{c\in\mathcal{C}} \sigma(f^c_z(x)),~ x\in\mathcal{X}\,, \label{eq:msp}
\end{equation} 
where $\sigma(\cdot): \mathbb{R}^{|\mathcal{C}|} \to (0,1)^{|\mathcal{C}|}$ denotes the softmax function over the non-anomalous class set $\mathcal{C}$.

\textbf{ODIN.}
Let $t\in\mathbb{R} \setminus \{0\}$ be a temperature scaling parameter and $\varepsilon\in\mathbb{R}$ a perturbation magnitude. Following \cite{liang18odin} small perturbations are added to every pixel $z\in\mathcal{Z}$ of image $x$ by
\begin{equation}
\tilde{x}_z = x_z - \varepsilon \text{sign}\left(- \frac{\partial}{\partial x_z} \log \max_{c\in\mathcal{C}}\sigma(f^c_z(x)/t)\right)\,. 
\end{equation}
Then, an anomaly score is obtained analogously to \cref{eq:msp} via the MSP as
\begin{equation}
    s_z(x) = 1 - \max_{c\in\mathcal{C}} \sigma(f^c_z(\tilde{x})/t) ~.
\end{equation}

\textbf{Mahalanobis distance.}
Let $h^{L-1}(\cdot)$ denote the output of the penultimate layer of a DNN with $L\in\mathbb{N}$ layers, \ie $f(x)=h^{L}(x), x\in\mathcal{X}$. Under the assumption that 
\begin{equation} \label{eq:gauss}
    P(h_z^{L-1}(x)~|~y_z(x)=c) = \mathcal{N}(h^{L-1}_z(x)~|~\mu^c, \Sigma^c)~,
\end{equation}
an anomaly score for each pixel $z$ can be computed as the Mahalanobis distance~\cite{Lee2018mahala} 
\begin{equation}
    s_z(x) = \min\limits_{c\in\mathcal{C}} ~ (h_z^{L-1}(x) - \hat{\mu}^c)^T {\hat{\Sigma}}^{c^{-1}} (h_z^{L-1}(x) - \hat{\mu}^c)\;,
\end{equation}
where $\hat{\mu}^c$ and $\hat{\Sigma}^c$ are estimates of the class mean $\mu^c$ and class covariance $\Sigma^c$, respectively, of the latent features in the penultimate layer. This Mahalanobis distance yields an estimate of the likelihood of a test sample with respect to the closest class distribution in the training data, which are assumed to be class-conditional Gaussians.

\textbf{Monte Carlo dropout.} Let $M\in\mathbb{N}$ denote the number of Monte Carlo sampling rounds and $\hat{q}^c_m := \sigma(f^c_z(x))$ the softmax probability of class $c\in\mathcal{C}$ for sample $m\in \{1, \ldots,  M \}$. The predictive entropy is computed as
\begin{equation}
    \hat{E}(f(x)) = -\sum_{c\in\mathcal{C}} \left(\frac{1}{M} \sum_{m=1}^M \hat{q}^c_m \right)\! \log \! \left(\frac{1}{M} \sum_{m=1}^M \hat{q}^c_m \right)\;.
\end{equation}
As suggested in~\cite{mukhoti2019bdl}, the mutual information can then be used to define an anomaly score
\begin{equation}
    s_z(x) = \hat{E}(f(x)) - \frac{1}{M} \sum_{c\in\mathcal{C}} \sum_{m=1}^M \hat{q}^c_m \log \left(\hat{q}^c_m\right)~.
\end{equation}

\textbf{Ensemble.} Similar to Monte Carlo dropout, multiple samples of softmax probabilities $\hat{q}^c_m := \sigma(f^c_z(x)), c\in\mathcal{C},m\in \{1, \ldots,  M\} $ are drawn from multiple semantic segmentation models. Those models have the same network architecture but are trained with different weights initialization \cite{Lakshminarayanan17}. Again, the mutual information is used as anomaly score
\begin{equation}
    s_z(x) = \hat{E}(f(x)) - \frac{1}{M} \sum_{c\in\mathcal{C}} \sum_{m=1}^M \hat{q}^c_m \log \left(\hat{q}^c_m\right)~.
\end{equation}

\textbf{Void classifier.} In~\cite{DeVries2018}, an approach to learning the confidence with respect to the presence of anomalies was proposed. Here, we adapt this by using the Cityscapes void class to approximate the anomaly distribution. We then trained a Cityscapes DNN $f:\mathcal{X}\mapsto\mathbb{R}^{|Z| \times (|\mathcal{C}|+1)}$ with an additional class, \ie, a dustbin~\cite{zhang17universum}, and compute the anomaly score for each pixel $z\in\mathcal{Z}$ as the softmax score for the void class, which yields
\begin{equation}
    s_z(x) = \sigma(f_z^{\mathrm{void}}(x)), ~ x\in\mathcal{X} ~.
\end{equation}

\textbf{Learned embedding density.}
Let ${h^l(x)\in \mathbb{R}^{ |\mathcal{Z}^\prime| \times n_l}}$, $n_l\in\mathbb{N}$, $\mathcal{Z}^\prime \subset \mathcal{Z}$, be the embedding vector of a segmentation DNN at layer $l\in\{1,\ldots,L\}$ for image $x\in\mathcal{X}$. The true distribution $p^*(h^l(x)), x \in \mathcal{X}_\textrm{train}\subset\mathcal{X}$ can be approximated with a normalizing flow $\hat{p}(h^l(x)) \approx p^*(h^l(x))$. At test time, the negative log-likelihood $-\log \hat{p}_{z^\prime}(h^l(x)) \in (0,\infty)$ for each embedding location $z^\prime \in \mathcal{Z}^\prime$ then measures the discrepancy of a test embedding with respect to training embeddings, where higher discrepancies indicate anomalies~\cite{Blum19fishyscapes}. 
The resulting anomaly score map are of size $ |\mathcal{Z}^\prime| = \frac{1}{n}|\mathcal{Z}|$, with $n\in\mathbb{N}$ the rescaling factor for $\mathcal{Z}^\prime$ to match the size of $\mathcal{Z}$, and hence bring back latent features to the full image resolution $|\mathcal{Z}|$ via bilinear interpolation $u: \mathbb{R}^{|\mathcal{Z}^\prime|} \to \mathbb{R}^{|\mathcal{Z}|}$. This yields an anomaly score for each $z\in\mathcal{Z}$ as
\begin{equation}
    s_z(x) = u_z \left(~ (-\log \hat{p}(h_{z^\prime}^l(x)))_{z^\prime\in\mathcal{Z}^\prime} ~ \right), x \in \mathcal{X} ~.
\end{equation}

\textbf{Image resynthesis.}
The semantic segmentation map $ \hat{y}(x):=(\arg \max_{c\in\mathcal{C}} f^c_z(x))_{z\in\mathcal{Z}}$ predicted by a DNN for image $x\in\mathcal{X}$ is passed to a generative network $g:\mathcal{C}^{|\mathcal{Z}|} \to \mathcal{X}^\prime$ whose goal is to resynthesize $x$, \ie $x \approx g(\hat{y}(x)) \in \mathcal{X}^\prime$, with $\mathcal{X}^\prime$ the resynthesized input space. Assuming that mislabeled pixels in the segmentation map, \ie anomaly pixels, will be poorly reconstructed, a \emph{discrepancy network}~\cite{Lis19} $d:\mathcal{C}^{|\mathcal{Z}|} \times \mathcal{X}^\prime \times \mathcal{X} \to \mathbb{R}^{\mathcal{|Z|}}$ is trained to extract the meaningful differences based on the information provided by $\hat{y}(x), g(\hat{y}(x))$ and $x$ itself. The output of $d(\cdot)$ serves as anomaly score for each $z\in\mathcal{Z}$, that is, 
\begin{equation}
    s_z(x) = d_z\left(~ \hat{y}(x), g(\hat{y}(x)), x ~\right), ~ x\in\mathcal{X} ~.
\end{equation}

\textbf{Road inpainting.} Another approach motivated by image resynthesis is road inpainting, which is specifically designed for obstacle segmentation. This method inpaints patches on the road (that is assumed to be known a-priori) in a sliding window manner and passes the resulting resynthesized image $g^\prime(x)$ to the discrepancy network together with the original input image. Thus, the anomaly score is 
\begin{equation}
    s_z(x) = d_z\left(~ g^\prime(x), x ~\right), ~ x\in\mathcal{X} ~.
\end{equation}

\textbf{SynBoost.}
This approach follows a similar idea as image resynthesis but includes further inputs in the discrepancy module. In particular, for all $ z\in\mathcal{Z}$ the pixel-wise softmax entropy
\begin{equation}
    H_z(x) = -\sum_{c\in\mathcal{C}} \sigma(f_z^c(x)) \log \left(\sigma(f_z^c(x)) \right)\end{equation}
and the pixel-wise softmax distance
\begin{equation}
    D_z(x) = 1-\max_{c\in\mathcal{C}} \sigma(f_z^c(x)) + \!\! \max_{c^\prime \in \mathcal{C} \setminus \{\argmax_{c\in\mathcal{C}}\}} \!\!\!\!\!\! \sigma(f_z^{c^\prime}(x))
\end{equation}
are included. The anomaly score for $x\in\mathcal{X}$ is then obtained via
\begin{equation}
    s_z(x) = d_z\left(~ \hat{y}(x), g(\hat{y}(x)), x, H(x), D(x) ~\right)~ .
\end{equation}

\textbf{Maximized entropy.} Starting from a pretrained DNN, a second training objective is introduced to maximize the softmax entropy on OoD pixels~\cite{chan2020entropy, hendrycks19oe, jourdan20}. This yields the multi-criteria loss function
\begin{equation} \label{eq:obj}
	(1-\lambda) \mathbb{E}_{(x,y) \sim \mathcal{D}_{in}} \left[\ell_{in} (\sigma(f_z(x)),y_z(x)) \right] + \lambda \mathbb{E}_{x^\prime \sim \mathcal{D}_{out}} \left[\ell_{out} (\sigma(f_z(x^\prime))) \right] ,~\lambda \in[0,1]\;,
\end{equation}
where $\ell_{in}$ is the empirical cross entropy and $\ell_{out}$ the averaged negative log-likelihood over all classes for the in-distribution data $\mathcal{D}_{in}$ and the out-distribution data $\mathcal{D}_{out}$, respectively.
To approximate $\mathcal{D}_{out}$, a subset of the COCO dataset~\cite{Lin14COCO} is used whose images do not depict any object classes also available in $\mathcal{D}_{in}$, which is the Cityscapes dataset \cite{Cordts2016Cityscapes}. The COCO subset together with the Cityscapes training data are then included into a tender retraining of the pretrained Cityscapes model. The anomaly score is then computed via the softmax entropy as
\begin{equation}
    s_z(x) =  -\sum_{c\in\mathcal{C}} \sigma(f_z^c(x)) \log \left(\sigma(f_z^c(x)) \right)~.
\end{equation}

\subsection{Underlying Segmentation DNNs} Most of our evaluated methods build upon variants of DeepLab \cite{Deeplab18} network architectures for semantic segmentation. In particular, for MC dropout, void classifier and learned embedding density we use a DeepLabv3+ model with an Xception backbone \cite{Chen2018ECCV}, as presented first in \cite{Blum19fishyscapes}. For maximum softmax, ODIN, Mahalanobis distance and maximized entropy, we employ a more modern DeepLabv3+ model with a WideResNet38 backbone \cite{Zhu19}. For image resynthesis we use the more lightweight PSPNet as underlying model for semantic segmentation just like originally proposed by \cite{Lis19}. All these networks are initialized with publicly available weights which are pretrained on the Cityscapes dataset. To show the capacity of the network, we report the mean Intersection over Union (mIoU) on the Cityscapes validation dataset in \Cref{tab:run-time}.

\subsection{Inference Time Comparison}

In practice, anomaly segmentation is desired to be obtained in real time. Therefore, we report the run-time of the evaluated anomaly segmentation methods as further performance metric that expresses a method's suitability as online application. We measure the total inference time for RoadAnomaly21, \ie the time from feeding all images through a model to obtaining pixel-wise anomaly scores. Afterwards we average the time per image and report them in \cref{tab:run-time}. All methods are compared with the same hardware (NVIDIA Quadro P6000), however they might differ in the underlying network architecture.

\begin{table}[ht]
    \begin{center}
    \scalebox{0.8}{
    \begin{tabular}{l|l|l|r}
    \toprule
     & Semantic segmentation & mIoU $\uparrow$ on & time in s $\downarrow$ \\
    Method & Network architecture & Cityscapes Val. & per image \\ 
    \midrule
    \midrule
    Maximum softmax & DeepLabv3+ WideResNet38 backbone \cite{Zhu19} & 90.3\% & 1.17 \\
    ODIN & DeepLabv3+ WideResNet38 backbone \cite{Zhu19} & 90.3\% & 16.74 \\
    Mahalanobis Distance & DeepLabv3+ WideResNet38 backbone \cite{Zhu19} & 90.3\% & 63.60 \\
    MC dropout & DeepLabv3+ Xcpection backbone \cite{Chen2018ECCV} & 80.3\% & 19.68 \\
    Void Classifier & DeepLabv3+ Xcpection backbone \cite{Chen2018ECCV} & 80.3\% & 2.02\\
    Embedding density & DeepLabv3+ Xcpection backbone \cite{Chen2018ECCV} & 80.3\% & 10.66 \\
    Image resynthesis & PSPNet \cite{Zhao2017PSP} & 79.9\% & 1.43 \\
    SynBoost & DeepLabv3+ WideResNet38 backbone \cite{Zhu19} & 90.3\% & 2.09 \\
    Maximized entropy & DeepLabv3+ WideResNet38 backbone \cite{Zhu19} & 89.3\% & 1.07 \\
    \midrule
    \end{tabular}
    }
    \end{center}
    \caption{Run time comparison on a NVIDIA Quadro P6000 for different anomaly segmentation methods. The averaged inference time for one image of RoadAnomaly21 is reported in seconds. Moreover, the mean Intersection over Union (mIoU) on the Cityscapes validation dataset is reported to check whether anomaly segmentation decreases the original semantic segmentation performance.}
    \label{tab:run-time}
\end{table}

\section{Parameter Study} \label{sec:param-study-supp}

\begin{figure}
    \centering
    ~~~~~\includegraphics[width=0.9\linewidth]{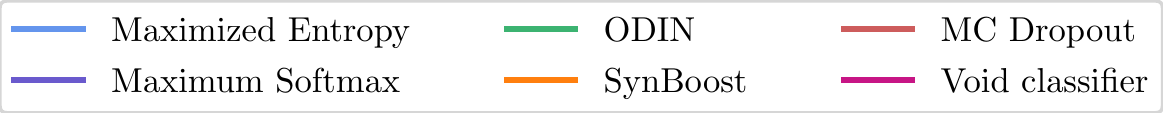}
    \subfloat[RoadAnomaly21]{\includegraphics[height=0.46\linewidth]{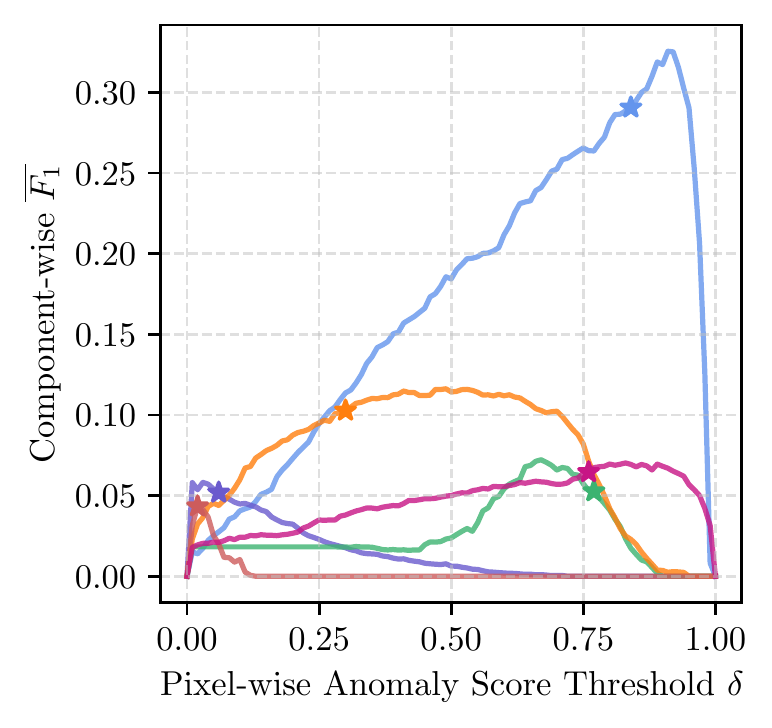}} ~
    \subfloat[RoadObstacle21]{\includegraphics[height=0.46\linewidth]{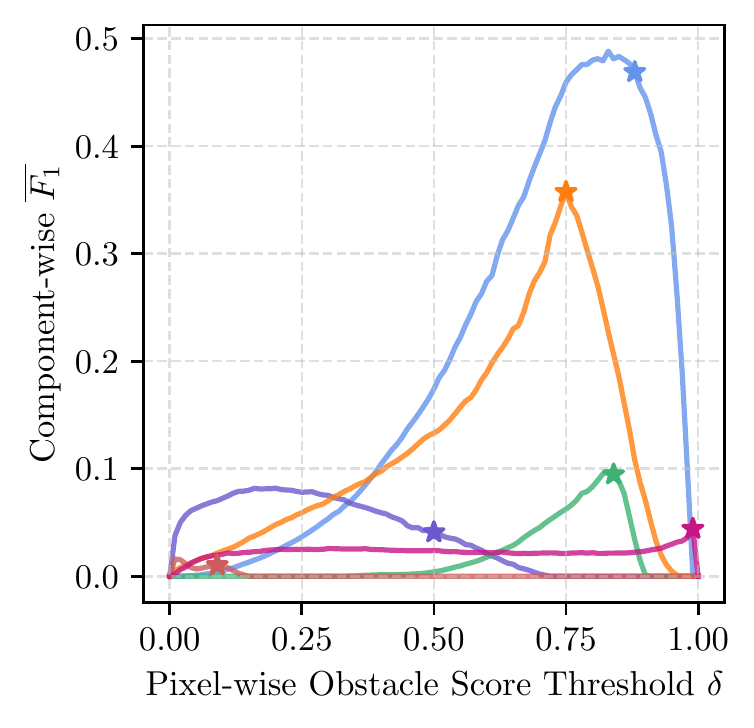}} ~
    \caption{The averaged component-wise metric $\overline{F_1}$ as function of the pixel-wise anomaly / obstacle threshold $\delta$ for RoadAnomaly21 and RoadObstacle21, respectively, \cf \cref{tab:table1} and \ref{tab:table2}. The ``star'' marker indicates a method's $\overline{F_1}$-score at the chosen threshold $\delta^*$ according to \cref{eq:optim-f1-pixel}, which is used in our default procedure for generating segmentation masks from pixel-wise anomaly / obstacle scores. We observe that for most methods $\delta^*$ yields near optimal component-wise $\overline{F_1}$-scores, however not for every single tested method. Therefore, we encourage competitors to submit their own anomaly segmentation masks based on more sophisticated methods.}
    \label{fig:param-study-delta}
\end{figure}

In our evaluation, the component-wise $F_1$ score (\cref{eq:comp-f1}) does not only depend on the parameter $\tau$ but also $\delta$. Recall that $\tau$ is the threshold for sIoU at which one component is considered to be false negative and true positive, respectively, see also \cref{sec:metrics}.
As we generate anomaly segmentation masks from pixel-wise anomaly scores, we introduced another threshold $\delta$ at which a given pixel is considered as anomaly. For generating segmentation masks with our default method, we chose that threshold as $\delta^*$ (\cref{eq:optim-f1-pixel}) which is the parameter for which a method achieves its best pixel-wise $F_1$ score, \ie the optimal threshold according to the precision recall curve.

In this section, we perform a parameter study to show what impact the choice of $\delta$ has on the component-wise performance. By considering $\overline{F_1}$ as component-wise performance metric we already cover varying values for $\tau$, since $\overline{F_1}$ is the average of component-wise $F_1$-scores over different values of $\tau$. The dependence of $\overline{F_1}$ on the parameter $\delta$ is illustrated in \cref{fig:param-study-delta} for RoadAnomaly21 and RoadObstacle21, respectively. For the sake of clarity, we only include six methods in total in this study, with at least one per type as discussed in \cref{sec:anomaly-seg}.

We observe that for most of the evaluated methods the choice of $\delta^*$ leads to an $\overline{F_1}$-score close its optimum, with some methods even reaching their optimal scores at $\delta^*$, \eg MC dropout on RoadAnomaly21 and SynBoost as well as the void classifier on RoadObstacle21. For the other methods the gap to the optimal $\overline{F_1}$-score reaches up to 2.8 percent points for maximized entropy on RoadAnomaly21 and even 4.1 percent points for maximum softmax on RoadObstacle21. However, except for the latter case where the distance between $\delta^*$ and the actual optimal location for $\overline{F_1}$ is 0.30, for all other methods the distance (in terms of $\overline{F_1}$) of $\delta^*$ to the optimal $\delta$
is at most 0.05.

This parameter study shows that our default method for generating segmentation masks from pixel-wise anomaly scores via the threshold $\delta^*$ is a legitimate choice, reaching a near optimal component-wise performance. Nonetheless, the parameter study also demonstrates that for some methods the $\overline{F_1}$-score can still be improved. Consequently, we allow (and encourage) competitors in the benchmark to submit their own anomaly segmentation masks with more sophisticated image operations and other post-processing techniques.

Another parameter included in the computation of the evaluation metrics is the size of predicted components in segmentation masks when generated from pixel-wise score maps. In our default post-processing method, we remove all components smaller than 500 pixels and 50 pixels in the anomaly and obstacle track, respectively, to reduce the amount of false positive components. To use this kind of filtering is completely optional. However, as can be seen in \Cref{tab:table-ra-filtering}, \Cref{tab:table-ro-filtering}, \Cref{tab:table-flaf-filtering} and \Cref{tab:table-lafnk-filtering}, we recommend using the post-processing option when competitors do not include a more sophisticated method. This is also why, we make our post-processing step transparent in this work since the size parameters are based on knowledge of ground truth components.

\begin{table*}[t]
    \setlength{\tabcolsep}{2pt}
    \begin{center}
    \scalebox{0.66}{
    \begin{tabular}{l||c||cc|ccc|g||cc|ccc|g}
    \toprule
    & & \multicolumn{6}{c||}{Component-level metrics with filtering} & \multicolumn{6}{c}{Component-level metrics without filtering}  \\
    \midrule
    \midrule
    & OoD & $k\in\mathcal{K}$ & $\hat{k}\in\mathcal{\hat{K}}$ & \multicolumn{3}{c|}{$\tau=0.50$} & \multicolumn{1}{c||}{} & $k\in\mathcal{K}$ & $\hat{k}\in\mathcal{\hat{K}}$ & \multicolumn{3}{c|}{$\tau=0.50$} & \multicolumn{1}{c}{} \\
    
    Method & data & $\overline{\mathrm{sIoU}}$ $\uparrow$ & $\overline{\mathrm{PPV}}$ $\uparrow$ & FN $\downarrow$ & FP $\downarrow$ & $F_1\uparrow$ & \multicolumn{1}{c||}{$\overline{F_1}\uparrow$} &
    $\overline{\mathrm{sIoU}}$ $\uparrow$ & $\overline{\mathrm{PPV}}$ $\uparrow$ & FN $\downarrow$ & FP $\downarrow$ & $F_1\uparrow$ & \multicolumn{1}{c}{$\overline{F_1}\uparrow$}\\
    \midrule
    
    Maximum softmax \cite{Hendrycks2017msp} & \xmark & 15.5 & 15.3 & 233 & 714 & 5.8 & 5.9 & 15.4 & 15.7 & 232 & 713 & 6.0 & 5.8 \\
    ODIN \cite{liang18odin} & \xmark & 19.6 & 17.9 & 226 & 985 & 5.6 & 6.0 & 19.7 & 17.5 & 227 & 983 & 5.5 & 6.0 \\
    Mahalanobis \cite{Lee2018mahala} & \xmark & 14.8 & 10.2 & 241 & 1478 & 2.4 & 2.9 & 14.8 & 10.5 & 241 & 1464 & 2.4 & 2.9 \\
    MC dropout \cite{mukhoti2019bdl} & \xmark  & 20.5 & 17.3 & 225 & 1391 & 4.4 & 4.9 & 20.5 & 17.3 & 225 & 1391 & 4.4 & 4.9 \\
    Ensemble \cite{Lakshminarayanan17} & \xmark & 16.4 & 20.8 & 233 & 1511 & 3.2 & 3.4 & 19.8 & 12.6 & 225 & 1528861 & 0.0 & 0.0 \\
    Void classifier \cite{Blum19fishyscapes} & \cmark & 21.1 & 22.1 & 219 & 845 & 7.5 & 7.6 & 21.1 & 22.1 & 219 & 845 & 7.5 & 7.6 \\ 
    Embedding density \cite{Blum19fishyscapes} & \xmark & 33.8 & 20.5 & 176 & 1485 & 9.4 & 9.2  & 34.0 & 20.8 & 176 & 1491 & 9.4 & 9.2 \\
    Image resynthesis \cite{Lis19} & \xmark & 39.5 & 11.0 & 153 & 1225 & 13.7 & 12.9 & 39.6 & 11.1 & 152 & 1225 & 13.8 & 13.0 \\
    SynBoost \cite{dibiase2021pixelwise} & \cmark & 35.0 & 18.3 & 178 & 1114 & 11.5 & 11.5 & 34.7 & 17.8 & 179 & 1129 & 11.3 & 11.2 \\
    Maximized entropy \cite{chan2020entropy} & \cmark & \textbf{49.2} &\textbf{39.5} & \textbf{115} & \textbf{421} & \textbf{35.4} & \textbf{34.5} & \textbf{49.2} & \textbf{39.4} &\textbf{115} & \textbf{421} & \textbf{35.4} & \textbf{34.4} \\
    \bottomrule
    
    \end{tabular}
    }
    \end{center}
    \vspace{-2mm}
    \caption{Comparison of benchmark results for our RoadAnomaly21 dataset when not using the filtering included in our default segmentation post-processing step. This dataset contains 262 ground-truth components in total. The main performance metrics are highlighted with gray columns.}
    \label{tab:table-ra-filtering}
\end{table*}

\begin{table*}[t]
    \setlength{\tabcolsep}{2pt}
    \begin{center}
    \scalebox{0.66}{
    \begin{tabular}{l||c||cc|ccc|g||cc|ccc|g}
    \toprule
    & & \multicolumn{6}{c||}{Component-level metrics with filtering} & \multicolumn{6}{c}{Component-level metrics without filtering}  \\
    \midrule
    \midrule
    & OoD & $k\in\mathcal{K}$ & $\hat{k}\in\mathcal{\hat{K}}$ & \multicolumn{3}{c|}{$\tau=0.50$} & \multicolumn{1}{c||}{} & $k\in\mathcal{K}$ & $\hat{k}\in\mathcal{\hat{K}}$ & \multicolumn{3}{c|}{$\tau=0.50$} & \multicolumn{1}{c}{} \\
    
    Method & data & $\overline{\mathrm{sIoU}}$ $\uparrow$ & $\overline{\mathrm{PPV}}$ $\uparrow$ & FN $\downarrow$ & FP $\downarrow$ & $F_1\uparrow$ & \multicolumn{1}{c||}{$\overline{F_1}\uparrow$} &
    $\overline{\mathrm{sIoU}}$ $\uparrow$ & $\overline{\mathrm{PPV}}$ $\uparrow$ & FN $\downarrow$ & FP $\downarrow$ & $F_1\uparrow$ & \multicolumn{1}{c}{$\overline{F_1}\uparrow$}\\
    \midrule
    
    Maximum softmax \cite{Hendrycks2017msp} & \xmark &                   19.7 &                     15.9 &                     326 &                    1503 &                      6.3 &                        6.9& 21.5 &                                                8.1 &                                              325 &                                            9624 &                                             1.3 &                                              1.4 \\
    ODIN \cite{liang18odin} & \xmark &                   20.7 &                     18.5 &                     312 &                    1079 &                      9.9 &                       10.0& 22.3 &                                                9.6 &                                              308 &                                            7260 &                                             2.1 &                                              2.1 \\ 
    Mahalanobis \cite{Lee2018mahala} & \xmark  &                   14.0 &                     21.8 &                     352 &                    1104 &                      4.7 &                        5.5& 17.0 &                                                7.5 &                                              348 &                                           13630 &                                             0.6 &                                              0.7 \\ 
    MC dropout \cite{mukhoti2019bdl} & \xmark &                    6.3 &                      5.8 &                     375 &                    2784 &                      0.8 &                        1.0 & 7.0 &                                                2.9 &                                                                                          375 &                                           20727 &                                             0.1 &                                              0.1\\
    Ensemble \cite{Lakshminarayanan17} & \xmark & 8.6 & 4.7 & 365 & 3768 & 1.1 & 1.3 & 11.0 & 1.8 & 364 & 369439 & 0.0 & 0.0\\
    Void classifier \cite{Blum19fishyscapes} & \cmark &                    6.3 &                     20.3 &                     365 &                     350 &                      6.0 &                        5.9 & 2.8 &                                               \textbf{42.8} &                                             
 384 &                                             \textbf{123} &                                             1.6 &                                              2.6\\ 
    Embedding density \cite{Blum19fishyscapes} & \xmark &                   35.6 &                      2.9 &                     244 &                   11037 &                      2.5 &                        2.4  &                                         
                                           36.1 &                                                1.6 &                                              246 &                                           33598 &                                             0.8 &                                              0.8 \\
    Image resynthesis \cite{Lis19} & \xmark &                   16.6 &                     20.5 &                     334 &                     773 &                      8.9 &                        9.5& 17.4 &                                               15.8 &                                              332 &                                            7003 &                                             1.5 &                                              1.6 \\ 
    Road inpainting \cite{lis2020detecting} & \xmark &                   \textbf{57.6} &                     39.5 &                     131 &                     586 &                     41.8 &                       40.2& \textbf{59.7} &                                               17.2 &                                               \textbf{127} &                                            4789 &                                             9.6 &                                              9.3 \\ 
    SynBoost \cite{dibiase2021pixelwise} & \cmark &                   44.3 &                     41.8 &                     185 &                     363 &                     42.6 &                       40.4&  45.2 &                                               22.6 &                                              185 &                                            1432 &                                            20.1 &                                              19.2 \\ 
    Maximized entropy \cite{chan2020entropy} & \cmark &                   47.9 &                     \textbf{62.6} &                     \textbf{177} &                     \textbf{158} &                     \textbf{55.7} &                       \textbf{54.2}& 48.7 & 35.1 & 177 & 758 & \textbf{31.1} & \textbf{30.3} \\
    \bottomrule
    
    \end{tabular}
    }
    \end{center}
    \vspace{-2mm}
    \caption{Comparison of benchmark results for our RoadObstacle21 dataset when not using the filtering included in our default segmentation post-processing step. This dataset contains 388 ground-truth components in total. The main performance metrics are highlighted with gray columns.}
    \label{tab:table-ro-filtering}
\end{table*}

\begin{table*}[t]
    \setlength{\tabcolsep}{2pt}
    \begin{center}
    \scalebox{0.66}{
    \begin{tabular}{l||c||cc|ccc|g||cc|ccc|g}
    \toprule
    & & \multicolumn{6}{c||}{Component-level metrics with filtering} & \multicolumn{6}{c}{Component-level metrics without filtering}  \\
    \midrule
    \midrule
    & OoD & $k\in\mathcal{K}$ & $\hat{k}\in\mathcal{\hat{K}}$ & \multicolumn{3}{c|}{$\tau=0.50$} & \multicolumn{1}{c||}{} & $k\in\mathcal{K}$ & $\hat{k}\in\mathcal{\hat{K}}$ & \multicolumn{3}{c|}{$\tau=0.50$} & \multicolumn{1}{c}{} \\
    
    Method & data & $\overline{\mathrm{sIoU}}$ $\uparrow$ & $\overline{\mathrm{PPV}}$ $\uparrow$ & FN $\downarrow$ & FP $\downarrow$ & $F_1\uparrow$ & \multicolumn{1}{c||}{$\overline{F_1}\uparrow$} &
    $\overline{\mathrm{sIoU}}$ $\uparrow$ & $\overline{\mathrm{PPV}}$ $\uparrow$ & FN $\downarrow$ & FP $\downarrow$ & $F_1\uparrow$ & \multicolumn{1}{c}{$\overline{F_1}\uparrow$}\\
    \midrule
    Maximum softmax \cite{Hendrycks2017msp} & \xmark  & 3.5 & 9.5 & 164 & 199 & 0.5 & 1.8  &                                              11.7 &                                                3.1 &          159 &                                           23134 &                                             0.1 &          0.1                                                                    \\
    ODIN \cite{liang18odin} & \xmark & 9.9 & 21.9  & 146 & 142 & 11.7 & 9.7    &                                              19.5 &                                                5.5 &                              136 &                                            6113 &                                             0.9 &      0.9                                                    \\
    Mahalanobis \cite{Lee2018mahala} & \xmark & 19.6 & 29.4 & 132 & 147 & 19.1 & 19.2 &                                              28.9 &                                                8.8 &                        124 &                                            4009 &                                             1.9 &                   2.1                                               \\
    MC dropout \cite{mukhoti2019bdl} & \xmark  & 4.8 & 18.1 & 160 & 120 & 3.4 & 4.3&                                               8.7 &                                               14.8 &                            158 &                                            1835 &                                             0.7 &              0.9                                               \\
    Ensemble \cite{Lakshminarayanan17} & \xmark & 3.1 & 1.1 & 162 & 1643 & 0.3 & 0.4 & 6.6 & 0.5 & 156 & 226622 & 0.0 & 0.0 \\
    Void classifier \cite{Blum19fishyscapes} & \cmark  & 9.2 & 39.1 & 149 & \textbf{38} & 14.6 & 14.9 &                                               9.6 &                                               16.6 &            149 &                                             \textbf{304} &                                             6.6 &                 6.6                                                           \\
    Embedding density \cite{Blum19fishyscapes} & \xmark  & 5.9 & 10.8 & 155 & 202 & 5.3 & 4.9 &                                              12.1 &                                                5.7 &                     150 &                                            3990 &                                             0.7 &                               0.7                                      \\
    Image resynthesis \cite{Lis19} & \xmark & 5.1 & 12.6 & 157 & 191 & 4.4 & 4.1 &                                               6.3 &                                                6.0 &                                157 &                                            5875 &                                             0.3 &           0.3                                              \\
    SynBoost \cite{dibiase2021pixelwise} & \cmark & \textbf{27.9} & \textbf{48.6} & \textbf{107} & 62 & \textbf{40.7} & \textbf{38.0} &                                              \textbf{35.3} &                                               \textbf{16.6} &       \textbf{97} &                                             723 &                                            \textbf{14.2} &                                           \textbf{13.3}                                       \\
    Maximized entropy \cite{chan2020entropy} & \cmark  & 21.1 & \textbf{48.6}  & 121 & 56 & 33.2  & 30.0 &                                              27.1 &                                               12.1 &           113 &                                            1160 &                                             7.6 &                         6.9                                                    \\
   
    \bottomrule
    
    \end{tabular}
    }
    \end{center}
    \vspace{-2mm}
    \caption{Comparison of benchmark results for the Fishyscapes LostAndFound validation dataset when not using the filtering included in our default segmentation post-processing step. This dataset contains 165 ground-truth components in total. The main performance metrics are highlighted with gray columns.}
    \label{tab:table-flaf-filtering}
\end{table*}

\begin{table*}[t]
    \setlength{\tabcolsep}{2pt}
    \begin{center}
    \scalebox{0.66}{
    \begin{tabular}{l||c||cc|ccc|g||cc|ccc|g}
    \toprule
    & & \multicolumn{6}{c||}{Component-level metrics with filtering} & \multicolumn{6}{c}{Component-level metrics without filtering}  \\
    \midrule
    \midrule
    & OoD & $k\in\mathcal{K}$ & $\hat{k}\in\mathcal{\hat{K}}$ & \multicolumn{3}{c|}{$\tau=0.50$} & \multicolumn{1}{c||}{} & $k\in\mathcal{K}$ & $\hat{k}\in\mathcal{\hat{K}}$ & \multicolumn{3}{c|}{$\tau=0.50$} & \multicolumn{1}{c}{} \\
    
    Method & data & $\overline{\mathrm{sIoU}}$ $\uparrow$ & $\overline{\mathrm{PPV}}$ $\uparrow$ & FN $\downarrow$ & FP $\downarrow$ & $F_1\uparrow$ & \multicolumn{1}{c||}{$\overline{F_1}\uparrow$} &
    $\overline{\mathrm{sIoU}}$ $\uparrow$ & $\overline{\mathrm{PPV}}$ $\uparrow$ & FN $\downarrow$ & FP $\downarrow$ & $F_1\uparrow$ & \multicolumn{1}{c}{$\overline{F_1}\uparrow$}\\
    \midrule
    Maximum softmax \cite{Hendrycks2017msp} & \xmark & 14.2 & 62.2 & 1575 & 602 & 11.0 & 13.4 & 16.3 & 17.5 & 1572 & 31481 & 0.8 & 1.1 \\
    ODIN \cite{liang18odin} & \xmark & 38.9 & 48.0 & 971 & 1303 & 39.4 & 38.1  & 40.2 & 29.9 & 967 & 5962 & 17.6 & 17.2 \\
    Mahalanobis \cite{Lee2018mahala} & \xmark & 33.8 & 31.7 & 1126 & 2314 & 25.3 & 24.6 & 34.7 & 22.8 & 1124 & 7677 & 11.7 & 11.6 \\
    MC dropout \cite{mukhoti2019bdl} & \xmark & 17.0 & 34.7 & 1453 & 1641 & 14.2 & 14.7 & 17.7 & 20.0 & 1451 & 9560 & 4.5 & 4.7 \\
    Ensemble \cite{Lakshminarayanan17} & \xmark & 6.7 & 7.6 & 1604 & 5649 & 2.8 & 2.7 & 7.5 & 3.8 & 1600 & 299431 & 0.1 & 0.1 \\
    Void classifier \cite{Blum19fishyscapes} & \cmark & 0.7 & 35.1 & 1698 & \textbf{108} & 1.2 & 1.1 & 0.7 & 25.1 & 1698 & \textbf{351} & 1.1 & 1.0 \\
    Embedding density \cite{Blum19fishyscapes} & \xmark & 37.8 & 35.2 & 963 & 1973 & 33.7 & 30.8 & 38.6 & 18.9 & 961 & 6862 & 16.1 & 14.8 \\
    Image resynthesis \cite{Lis19} & \xmark & 27.2 & 30.7 & 1232 & 2093 & 22.3 & 21.5 & 28.0 & 19.7 & 1228 & 15418 & 5.5 & 5.3 \\
    Road inpainting \cite{lis2020detecting} & \xmark & \textbf{49.2} & 60.7 & \textbf{749} & 646 & \textbf{57.9} & \textbf{56.9} &   \textbf{50.4} & 33.0 & \textbf{743} & 4852 & 25.7 & 25.2 \\
    SynBoost \cite{dibiase2021pixelwise} & \cmark & 37.2 & \textbf{72.3} & 930 & 230 & 57.3 & 53.0 & 37.6 & \textbf{63.3} & 931 & 535 & \textbf{51.5} & \textbf{47.7} \\
    Maximized entropy \cite{chan2020entropy} & \cmark & 45.9 & 63.1 & 781 & 598 & 57.4 & 55.0 & 46.7 & 35.8 & 778 & 2813 & 34.1 & 32.7 \\
    
    \bottomrule
    
    \end{tabular}
    }
    \end{center}
    \vspace{-2mm}
    \caption{Comparison of benchmark results for the LostAndFound test-NoKnown dataset when not using the filtering included in our default segmentation post-processing step. This dataset contains 1709 ground-truth components in total. The main performance metrics are highlighted with gray columns.}
    \label{tab:table-lafnk-filtering}
\end{table*}

\section{Evaluated Datasets}

Besides RoadAnomaly21 and RoadObstacle21 we also performed analogous benchmark evaluations for three additional publicly available datasets: Fishyscapes LostAndFound \cite{Blum19fishyscapes}, LostAndFound test set \cite{Pinggera16LostAndFound}, and the LiDAR guided Small obstacle Segmentation dataset \cite{Singh2020-fw}. For the sake of comparison, we chose the Fishyscapes LostAndFound validation set for the anomaly track and the LostAndFound test set as well as the Small Obstacle dataset for the obstacle track.

\subsection{RoadAnomaly21 \& RoadObstacle21 Validation Dataset}
In order to ensure that methods run as intended with our benchmark code, we provide small validation sets (including ground truth annotations) for the anomaly track, called \emph{RoadAnomaly21 validation}, and for the obstacle track, called \emph{RoadObstacle21 validation}.

These datasets show similar scenes and objects as in RoadAnomaly21 test and RoadObstacle21 test, respectively. The splits contain 10 images with 16 ground truth components and 30 images with 45 ground truth objects in total, respectively. Note that although both datasets share the same setup as in the corresponding test splits, they are still \textbf{not} representative for the test data since they contains only a very limited number of different road surfaces and diverse obstacle types. Therefore we do not recommend to fine-tune methods on these two validation datasets.

Moreover, we applied our set of anomaly segmentation methods to RoadObstacle21 validation, see \cref{tab:table-val-obstacle}. Some of those methods are also made publicly available in our benchmark code to compare to and reproduce the reported results.

\subsection{Fishyscapes LostAndFound}
The Fishyscapes LostAndFound validation dataset \cite{Blum19fishyscapes} consists of 100 images from the original LostAndFound data \cite{Pinggera16LostAndFound} with refined labels. With this labeling, anomalous objects are not restricted to only appear on the road but everywhere in the image, therefore Fishyscapes LostAndFound fits our benchmark's anomaly track. 

Comparing the RoadAnomaly21 and Fishyscapes LostAndFound datasets in terms of anomaly class frequency per pixel location, as observed in \cref{fig:pixel-distributions}, one notices a clear difference in the variation of object locations and sizes. While in Fishyscapes LostAndFound the objects appear mostly in the center of the image and are also rather small, the objects in RoadAnomaly21 may appear everywhere in the image and have sizes ranging from 122 up to 883,319 pixels (thus covering up to more than one third of the image). The low variety in object sizes is also noticeable in the pixel-wise class distribution, as in RoadAnomaly21 $13.8\%$ of the pixels belong to the anomaly class and $82.2\%$ to non-anomaly whereas in Fishyscapes LostAndFound only $0.23\%$ belong to anomaly and $81.13\%$ to non-anomaly.

\begin{figure*}[ht]
    \centering
    \captionsetup[subfigure]{labelformat=empty, position=top}
    \subfloat[Image \& annotation]{\includegraphics[width=0.199\textwidth]{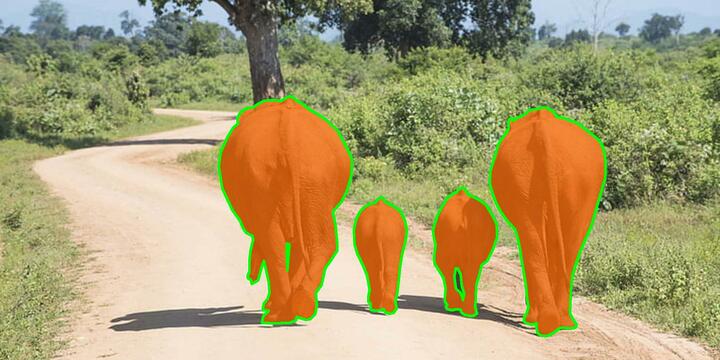}}
    \subfloat[Maximum softmax]{\includegraphics[width=0.199\textwidth]{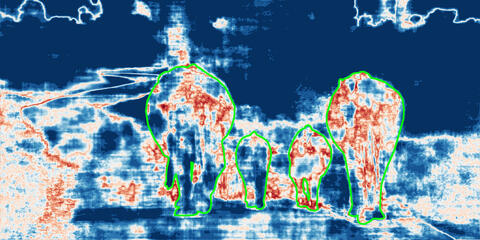}}
    \subfloat[ODIN]{\includegraphics[width=0.199\textwidth]{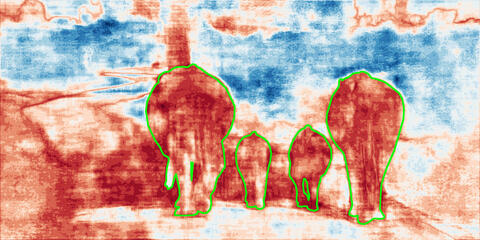}}
    \subfloat[Mahalanobis]{\includegraphics[width=0.199\textwidth]{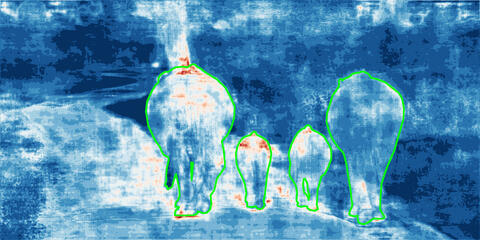}}
    \subfloat[MC dropout]{\includegraphics[width=0.199\textwidth]{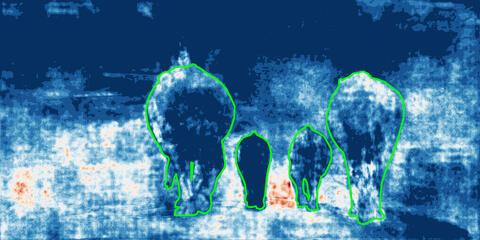}}\\
    \vspace{-1.07\baselineskip}
    \captionsetup[subfigure]{labelformat=empty, position=bottom}
    \subfloat[Void classifier]{\includegraphics[width=0.199\textwidth]{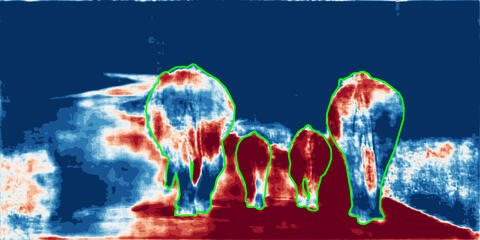}}
    \subfloat[Embedding density]{\includegraphics[width=0.199\textwidth]{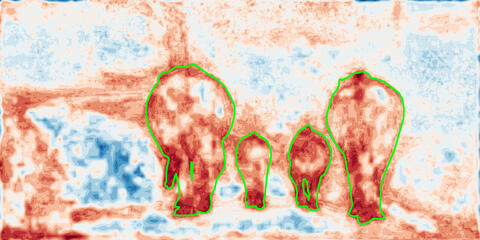}}
    \subfloat[Image resynthesis]{\includegraphics[width=0.199\textwidth]{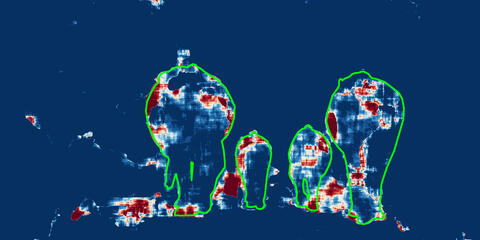}}
    \subfloat[SynBoost]{\includegraphics[width=0.199\textwidth]{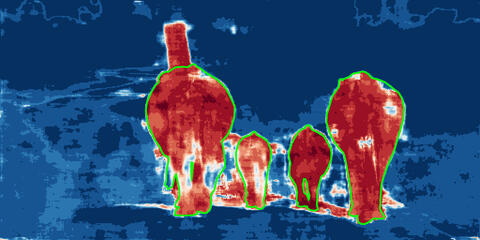}}
    \subfloat[Maximized entropy]{\includegraphics[width=0.199\textwidth]{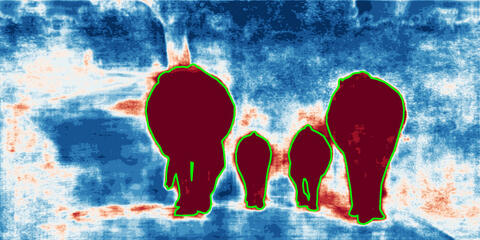}}
    \caption{Qualitative comparison of the methods introduced in \cref{sec:methods} and \cref{sec:methods-detail} on a sample from RoadAnomaly21. In this example, the anomalous objects have a large size and the environment differs from scenes shown in Cityscapes. Green contours indicate the annotation of the anomaly.
    }
    \label{fig:example1}
\end{figure*}

As already discussed in \cref{sec:discussion}, we observe a less pronounced gap between methods designed for image classification and those specifically designed for anomaly segmentation. A detailed overview of our benchmark results on Fishyscapes LostAndFound is given in \cref{tab:table5}. In this evaluation, we see that the number of false positive components (relative to the number of ground truth components) over multiple thresholds $\tau$ is significantly less than on RoadAnomaly21, shown in Table~\ref{tab:table1}. This holds for all evaluated methods, resulting in relatively strong component-wise performance (compared to SynBoost and maximized entropy). Even Mahalanobis and void classifier report strong results, which is due to similarity of this dataet to Cityscapes as all LostAndFound images share the same setup as in Cityscapes. These results further indicate the lack in diversity in Fishyscapes LostAndFound. More specifically, the environments of the scenes shown in LostAndFound do not considerably differ to those shown in Cityscapes whereas our RoadAnomaly21 dataset has a wide variety of scenes since all images are gathered from the web, see \cref{fig:example1}.

\subsection{LostAndFound test-NoKnown}
The LostAndFound dataset \cite{Pinggera16LostAndFound} shares the same setup as Cityscapes but includes small obstacles on the road. Therefore, this dataset fits our benchmark's obstacle track.
When a model is trained on Cityscapes, the LostAndFound dataset then contains images with objects that have been previously seen and therefore are not anomalies. As most of our methods are designed for anomaly detection, we filtered out all scenes in the LostAndFound test split where the obstacles belong to known classes, \eg children or bicycles, and call this subset LostAndFound test-NoKnown. In this way, the results obtained with our evaluated methods on LostAndFound test-NoKnown and on our RoadObstacle21 dataset are comparable.

\begin{figure*}[ht]
    \centering
    \captionsetup[subfigure]{labelformat=empty}
    \subfloat[RoadAnomaly21 ]{\includegraphics[height=.23\linewidth]{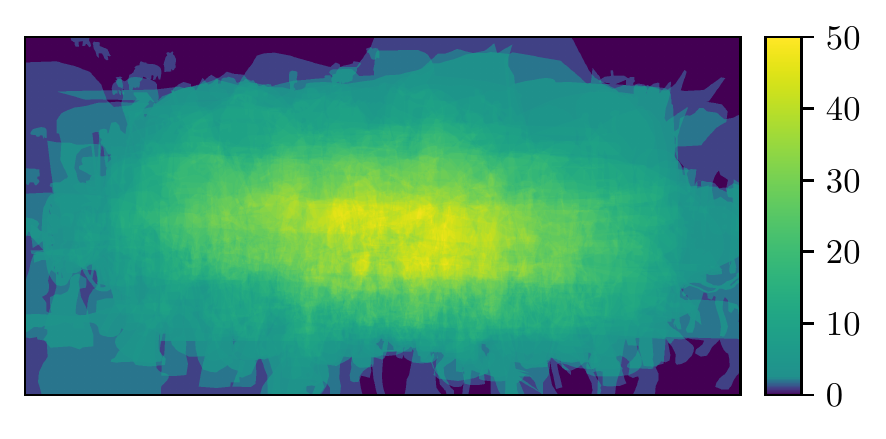}} ~
    \subfloat[RoadObstacle21]{\includegraphics[height=.23\linewidth]{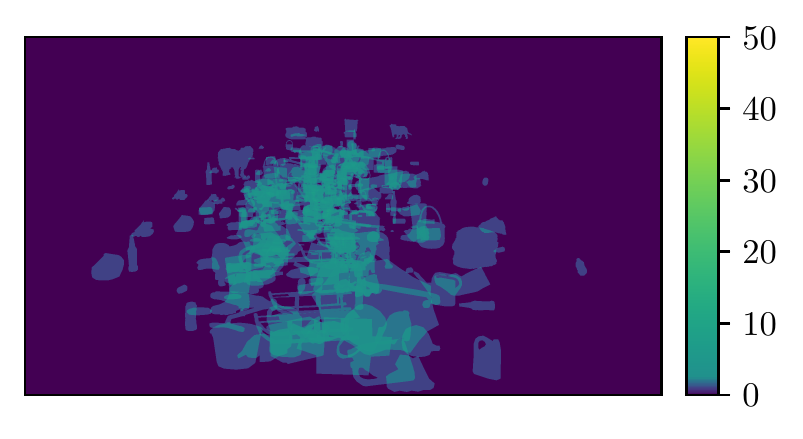}} \\
    ~~~~~~~\subfloat[Fishyscapes LostAndFound validation]{\includegraphics[height=.23\linewidth]{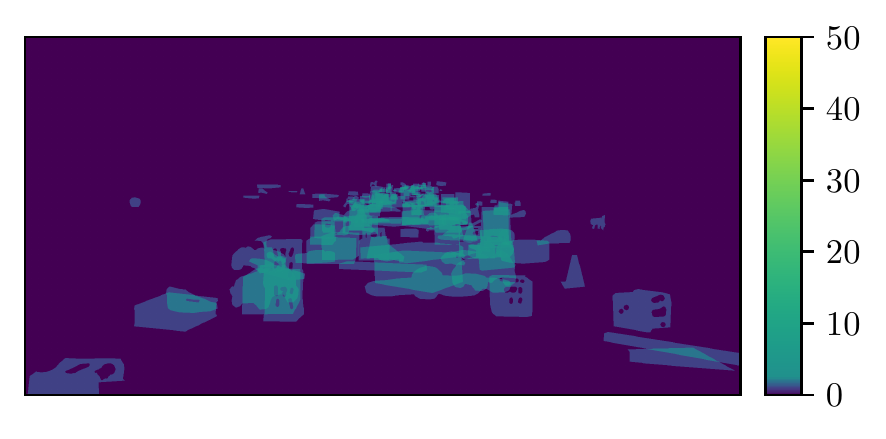}} ~
    \subfloat[LostAndFound test]{\includegraphics[height=.23\linewidth]{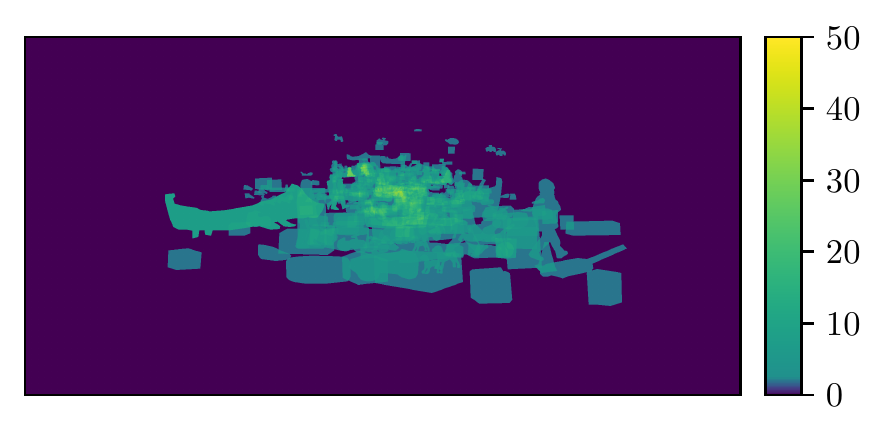}}
    \caption{Comparison of the (spatial) pixel distributions between RoadAnomaly21 and Fishyscapes LostAndFound (100 images each) as well as RoadObstacle21 and a subset of randomly sampled images from the LostAndFound test dataset (327 images each). The color indicates the frequency of observing an anomaly in each pixel location, averaged over the images in the dataset.}
    \label{fig:pixel-distributions}
\end{figure*}

\begin{figure*}[ht]
    \centering
    \captionsetup[subfigure]{labelformat=empty, position=top}
    \subfloat[Image \& annotation]{\includegraphics[width=0.199\textwidth]{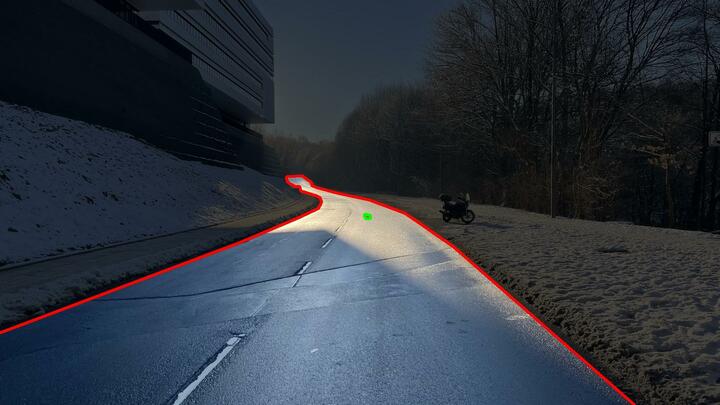}}
    \subfloat[Maximum softmax]{\includegraphics[width=0.199\textwidth]{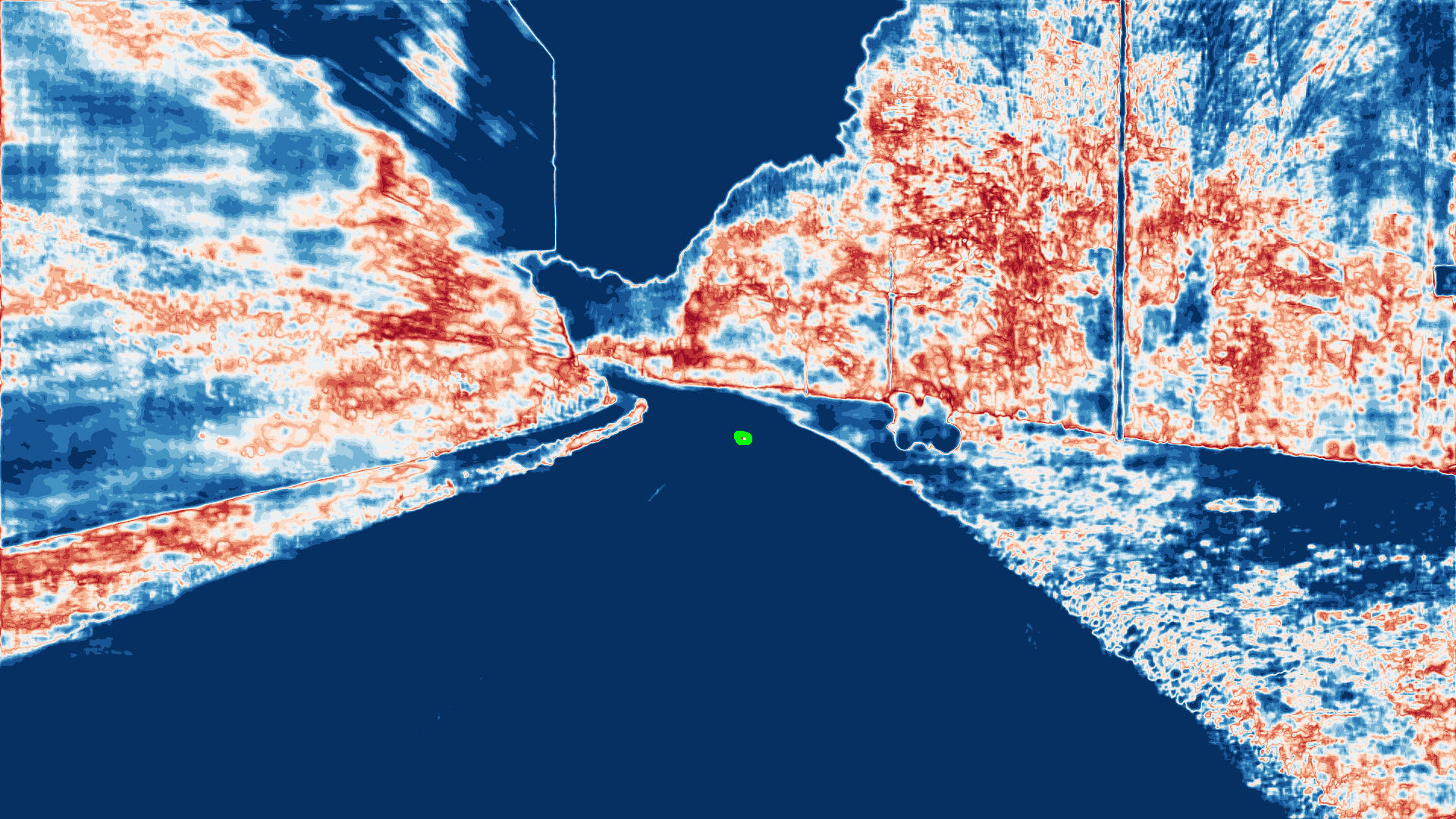}}
    \subfloat[ODIN]{\includegraphics[width=0.199\textwidth]{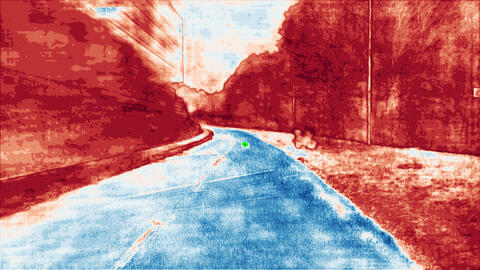}}
    \subfloat[Mahalanobis]{\includegraphics[width=0.199\textwidth]{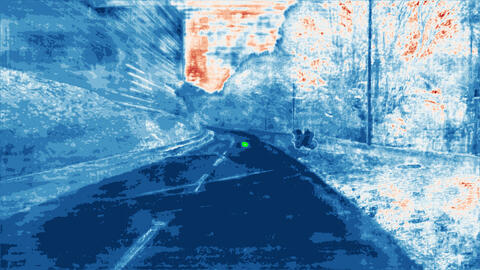}}
    \subfloat[MC dropout]{\includegraphics[width=0.199\textwidth]{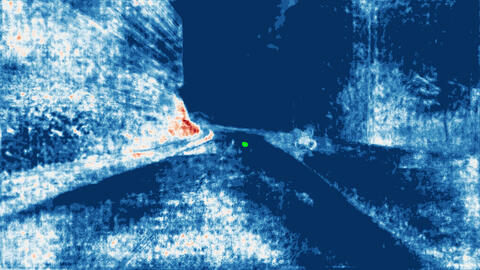}}\\
    \vspace{-1.07\baselineskip}
    \captionsetup[subfigure]{labelformat=empty, position=bottom}
    \subfloat[Void classifier]{\includegraphics[width=0.199\textwidth]{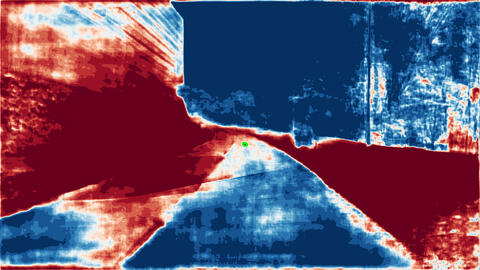}}
    \subfloat[Embedding density]{\includegraphics[width=0.199\textwidth]{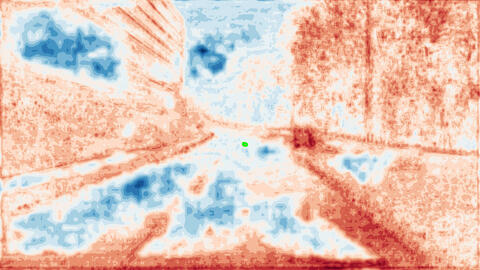}}
    \subfloat[Image resynthesis]{\includegraphics[width=0.199\textwidth]{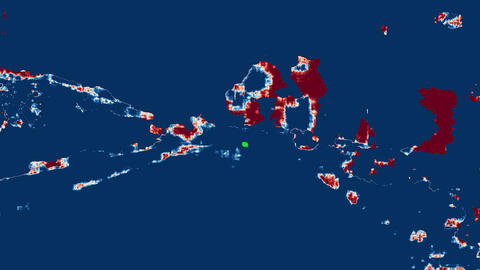}}
    \subfloat[SynBoost]{\includegraphics[width=0.199\textwidth]{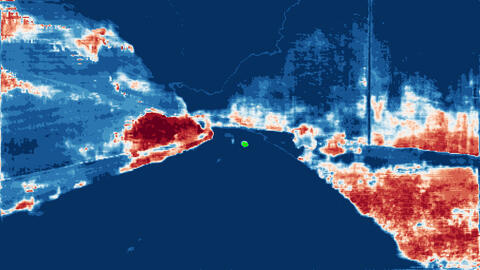}}
    \subfloat[Maximized entropy]{\includegraphics[width=0.199\textwidth]{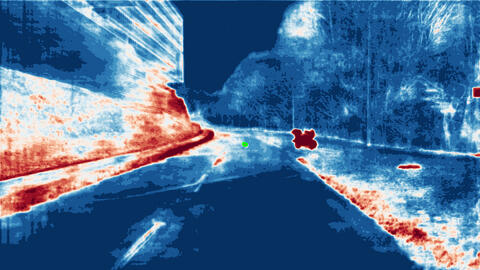}}
    \caption{Qualitative comparison of the methods introduced in \cref{sec:methods} and \cref{sec:methods-detail} for an example from RoadObstacle21, where the obstacle is small and far away. Green contours indicate the annotation of the obstacle, red contours the road.
    }
    \label{fig:example2}
\end{figure*}

\begin{figure*}[ht]
    \centering
    \captionsetup[subfigure]{labelformat=empty, position=top}
    \subfloat[Image \& annotation]{\includegraphics[width=0.199\textwidth]{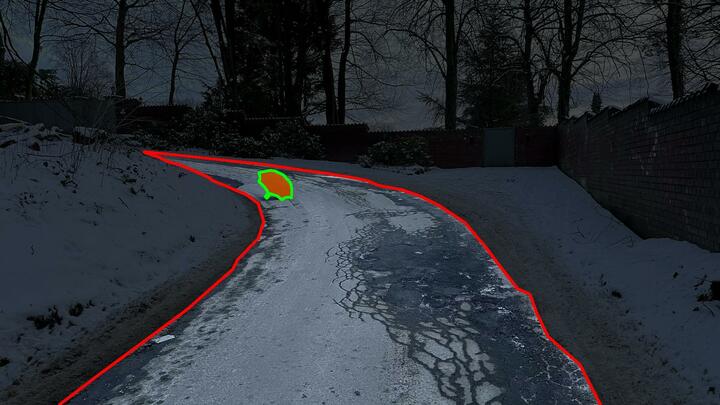}}
    \subfloat[Maximum softmax]{\includegraphics[width=0.199\textwidth]{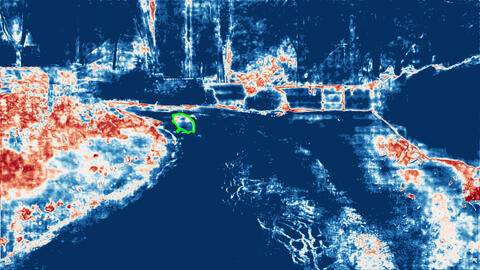}}
    \subfloat[ODIN]{\includegraphics[width=0.199\textwidth]{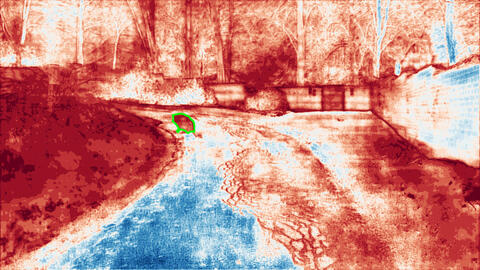}}
    \subfloat[Mahalanobis]{\includegraphics[width=0.199\textwidth]{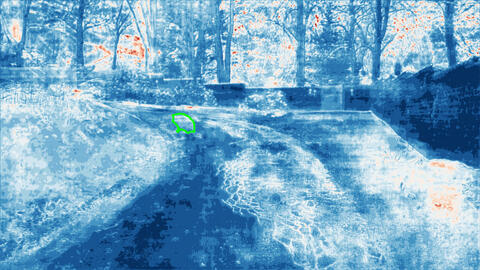}}
    \subfloat[MC dropout]{\includegraphics[width=0.199\textwidth]{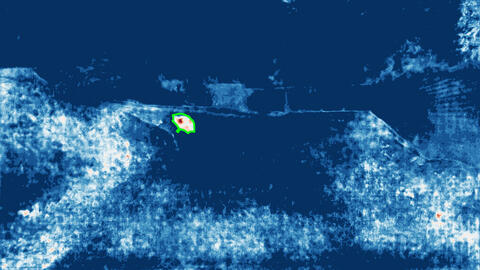}}\\
    \vspace{-1.07\baselineskip}
    \captionsetup[subfigure]{labelformat=empty, position=bottom}
    \subfloat[Void classifier]{\includegraphics[width=0.199\textwidth]{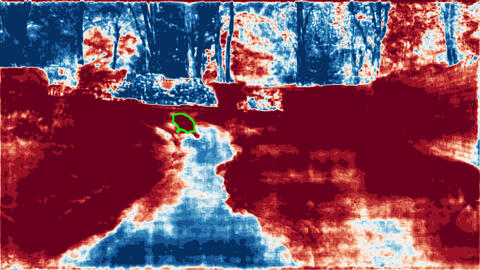}}
    \subfloat[Embedding density]{\includegraphics[width=0.199\textwidth]{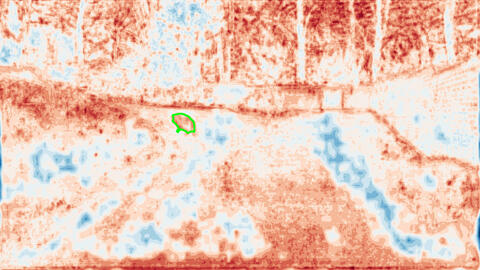}}
    \subfloat[Image resynthesis]{\includegraphics[width=0.199\textwidth]{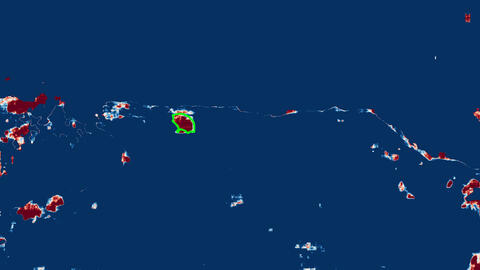}}
    \subfloat[SynBoost]{\includegraphics[width=0.199\textwidth]{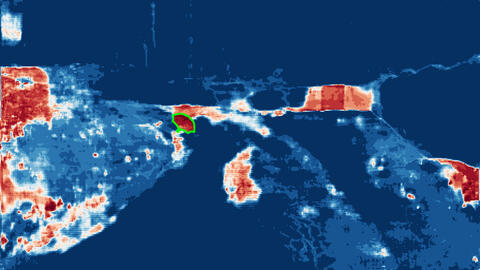}}
    \subfloat[Maximized entropy]{\includegraphics[width=0.199\textwidth]{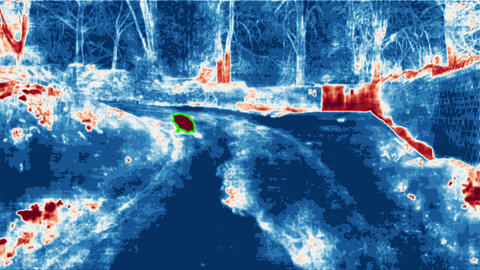}}
    \caption{Qualitative comparison of the methods introduced in \cref{sec:methods} and \cref{sec:methods-detail} for an example from RoadObstacle21, showing a road surface with cracks. Green contours indicate the annotation of the obstacle, red contours the road.
    }
    \label{fig:example3}
\end{figure*}

Both datasets have obstacles in the same size range. Both RoadObstacle21 and LostAndFound test-NoKnown have $0.12\%$ of the pixels labeled as obstacles, while $39.08\%$ and $15.31\%$ of the pixels belong to not obstacles, respectively. Regarding the object locations in images, the obstacles in RoadObstacle21 are distributed wider over the image than in LostAndFound, as observed in \cref{fig:pixel-distributions}. This also implies that in RoadObstacle21 the obstacles appear at stronger varying distances. For an illustration as well as of that variation, we refer to \cref{fig:example2,fig:example3}. Looking at the results in \cref{tab:table6}, we observe for LostAndFound test-NoKnown, just as in Fishyscapes LostAndFound (\cref{tab:table5}), that methods from image classification perform relatively well in comparison to methods designed for anomaly segmentation. This is again due the limited variety of environments, \ie the road surfaces in this dataset. In our RoadObstacle21 dataset, we therefore provide scenes with obstacles on different road surfaces, such as gravel or a road with cracks, see \cref{fig:example3}.

\begin{figure*}
    \centering
    \captionsetup[subfigure]{labelformat=empty}
    \subfloat[]{\includegraphics[width=0.25\textwidth]{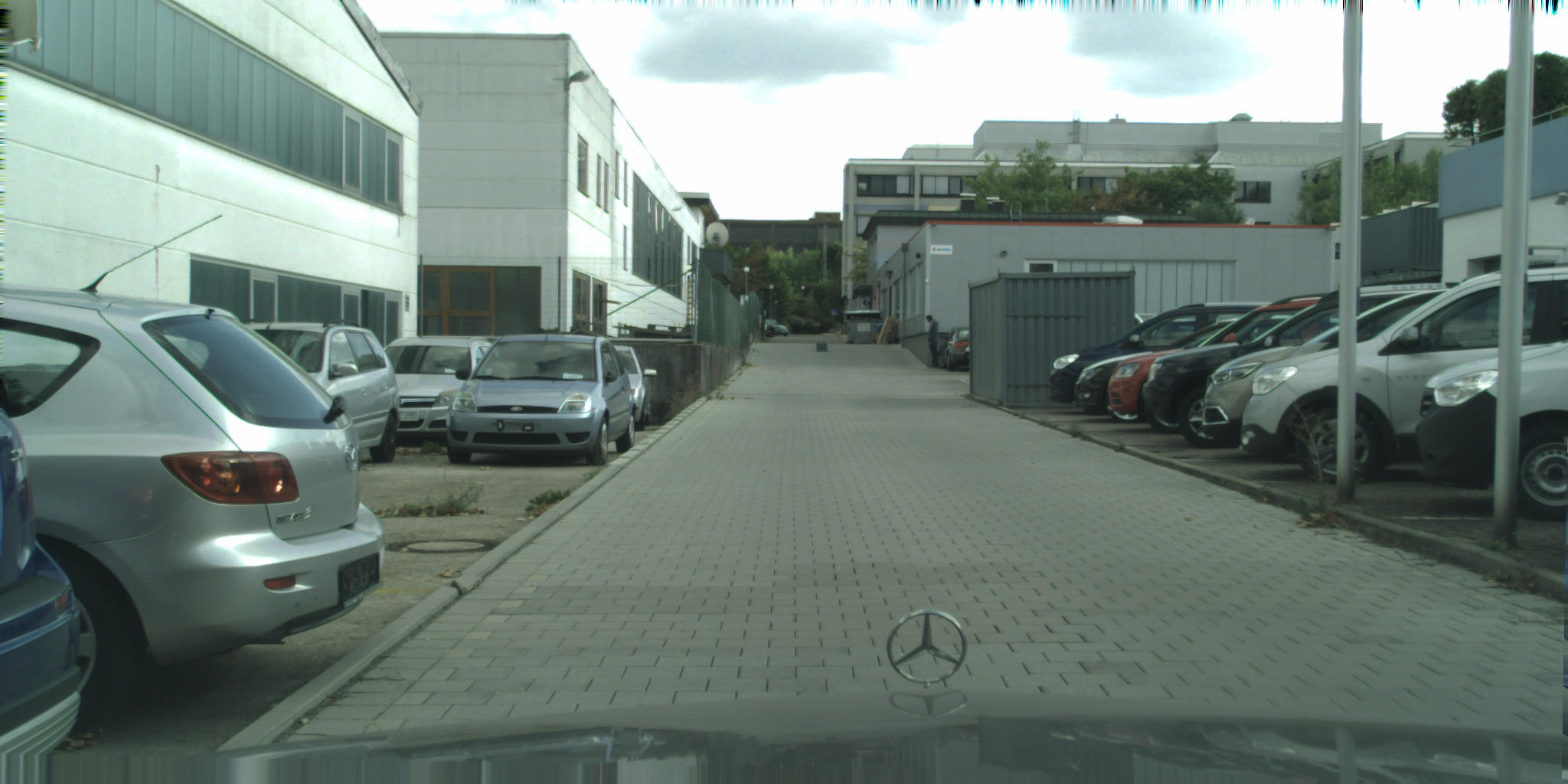}}%
    \subfloat[]{\includegraphics[width=0.25\textwidth]{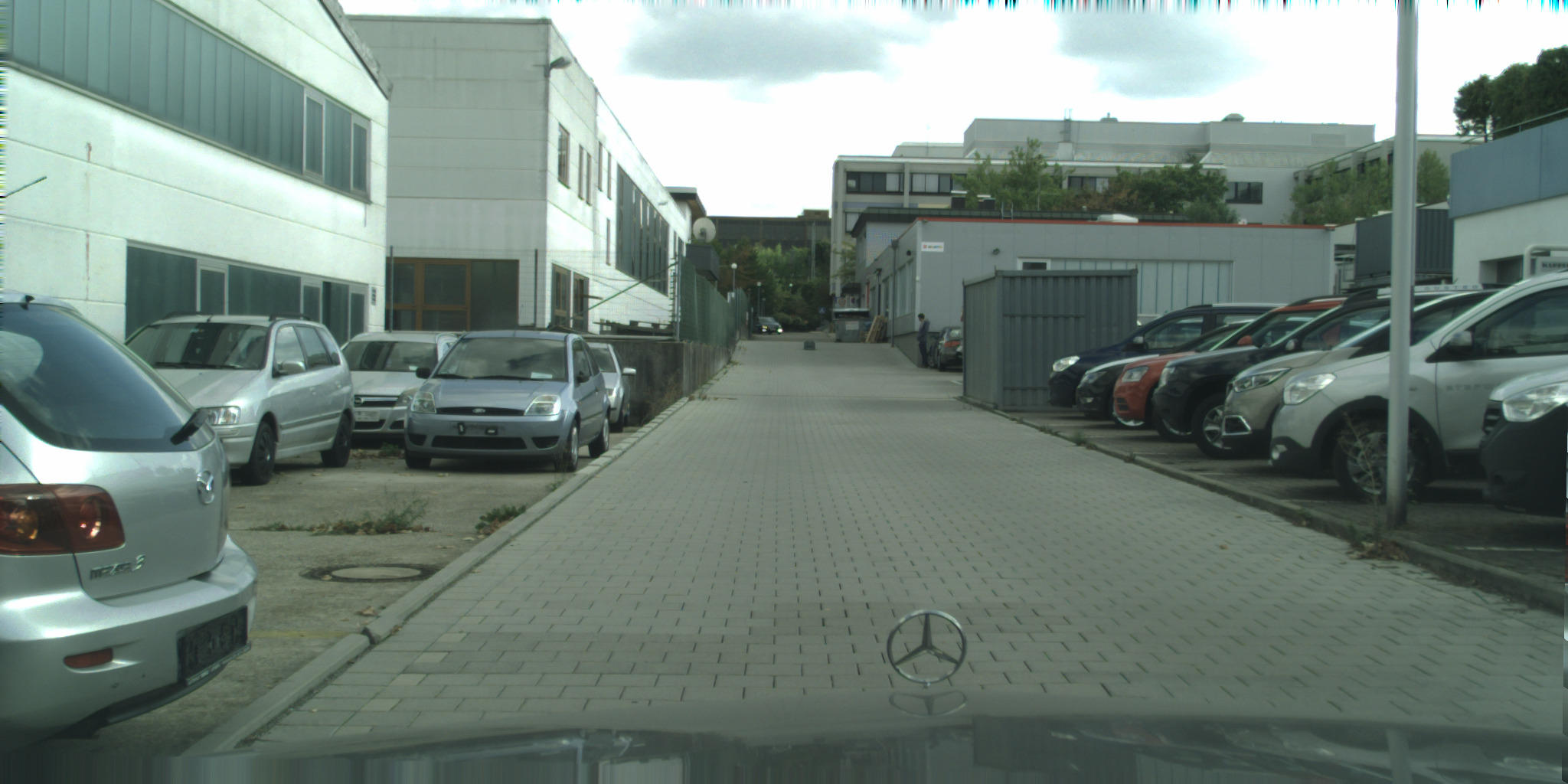}}%
    \subfloat[]{\includegraphics[width=0.25\textwidth]{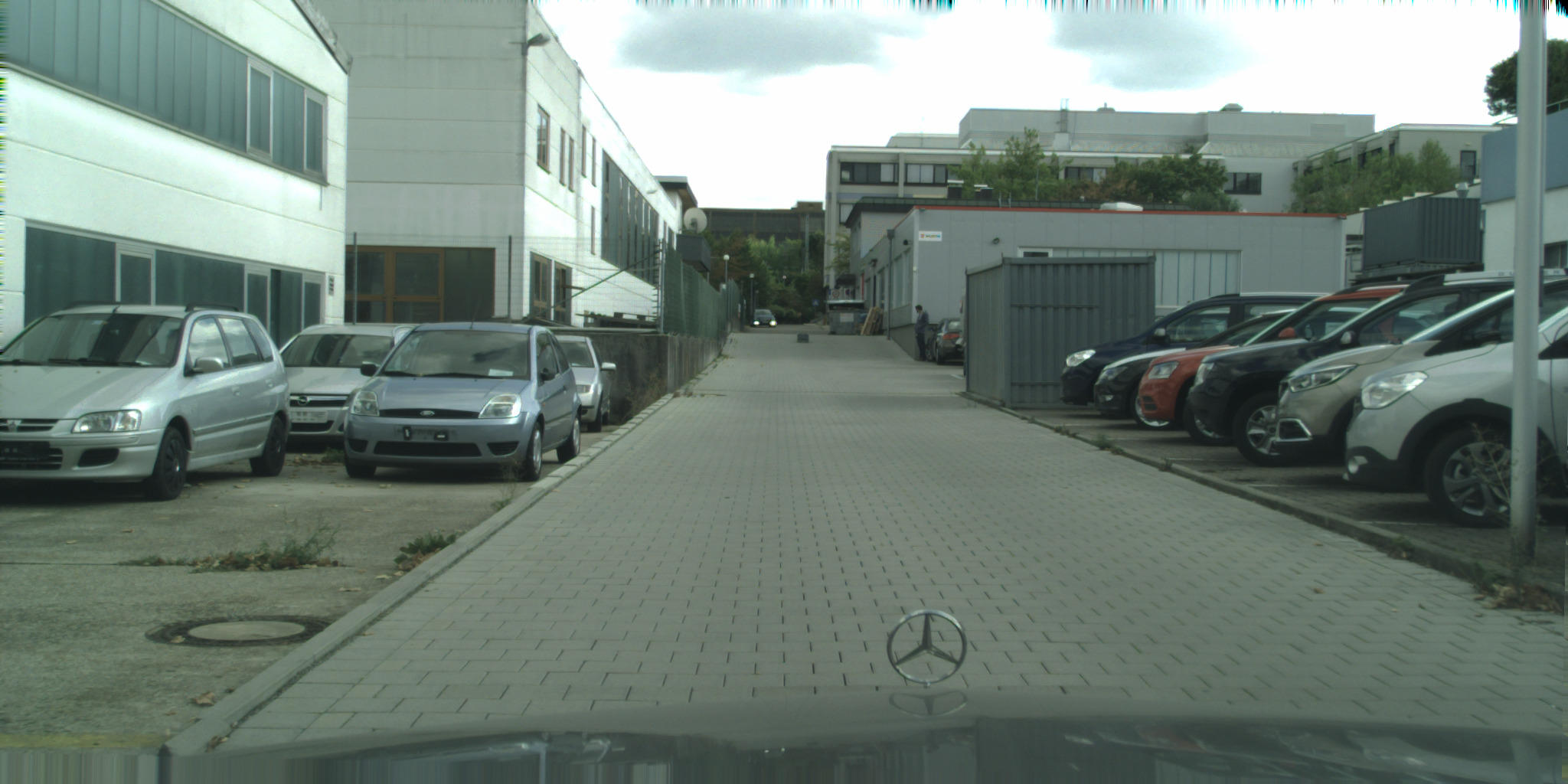}}%
    \subfloat[]{\includegraphics[width=0.25\textwidth]{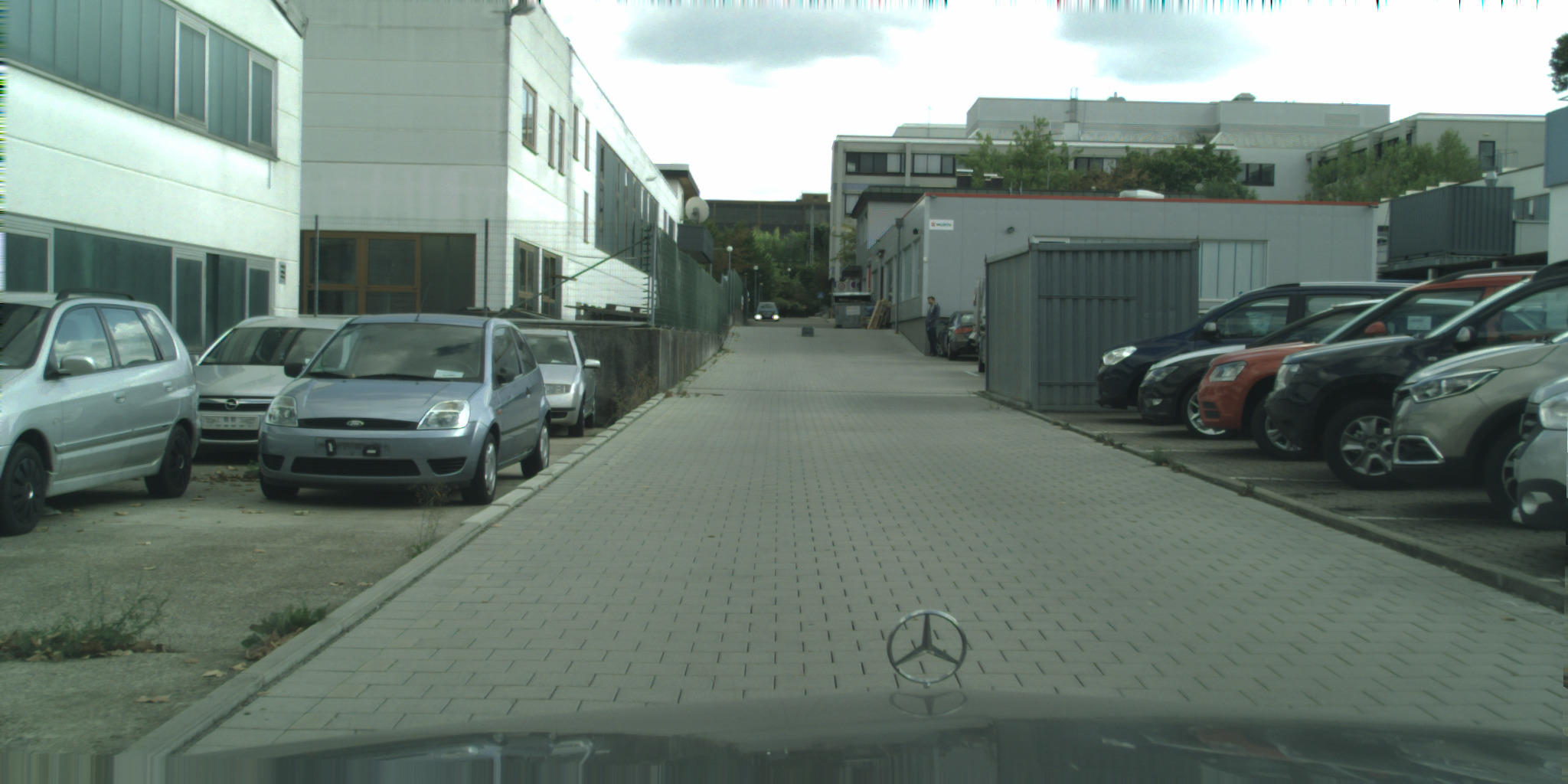}}\\
    \vspace{-2.3\baselineskip}
    \captionsetup[subfigure]{labelformat=empty, position=bottom}
    \subfloat[Frame 1]{\includegraphics[width=0.25\textwidth]{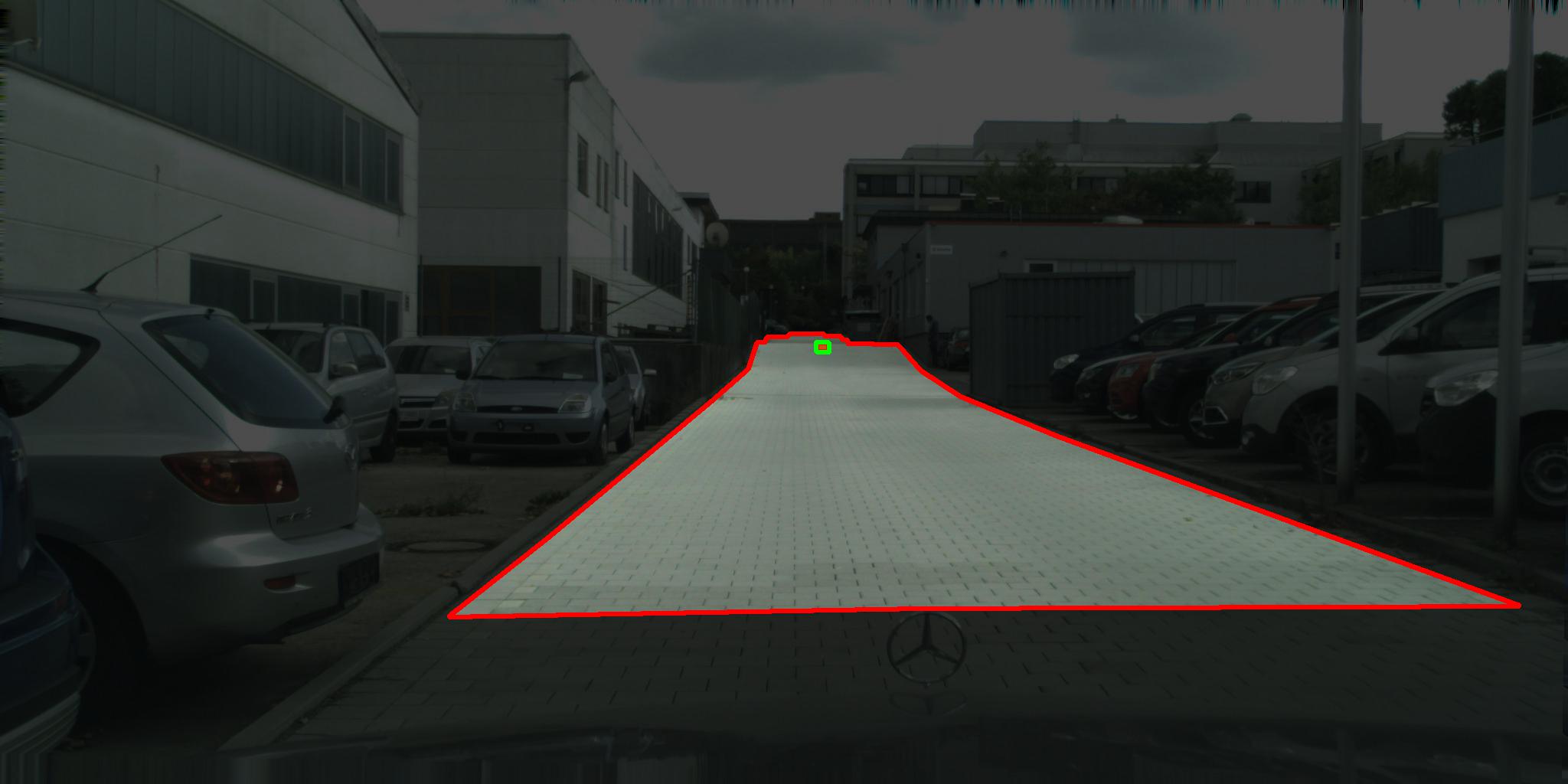}}%
    \subfloat[Frame 2]{\includegraphics[width=0.25\textwidth]{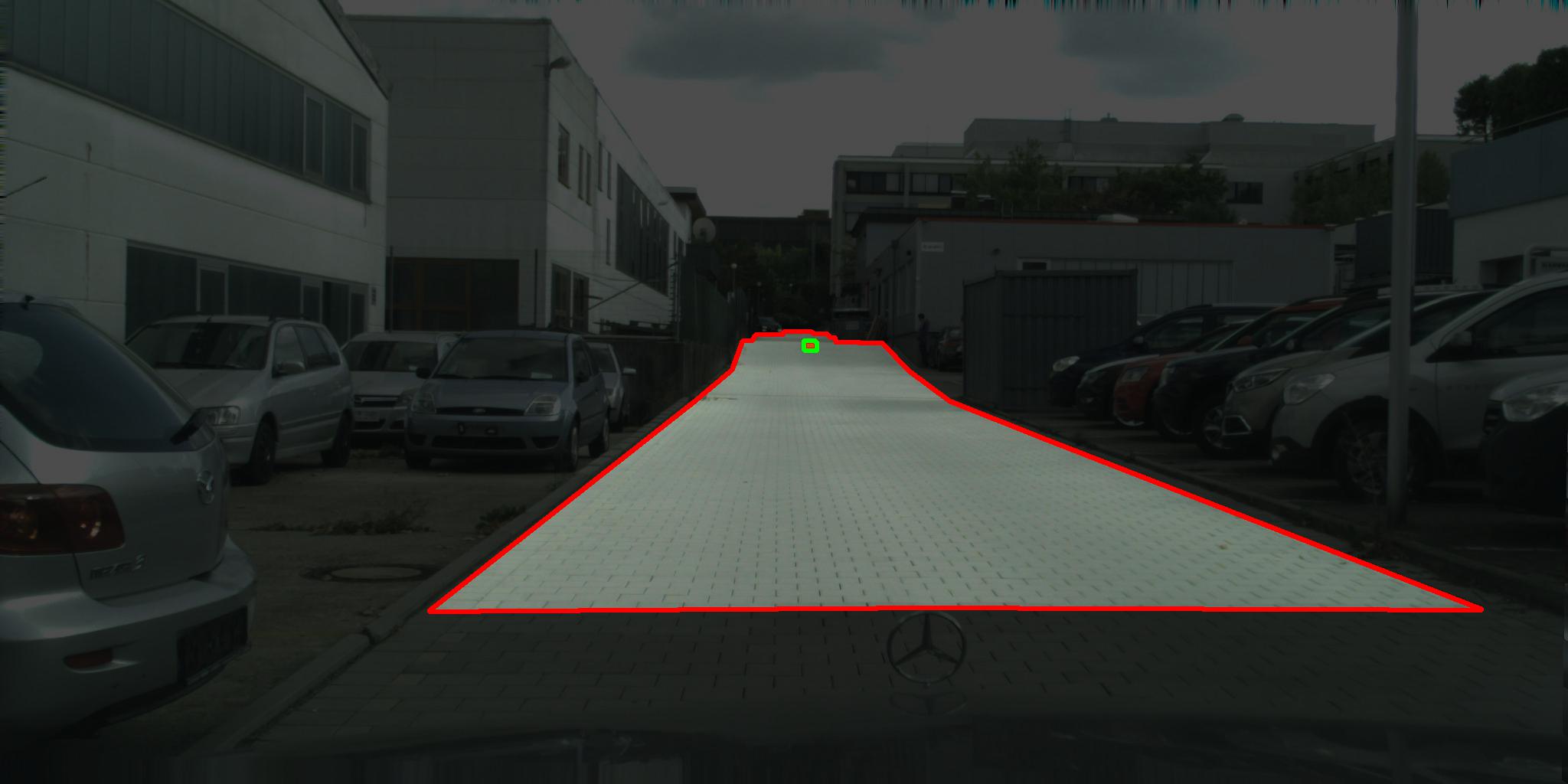}}%
    \subfloat[Frame 3]{\includegraphics[width=0.25\textwidth]{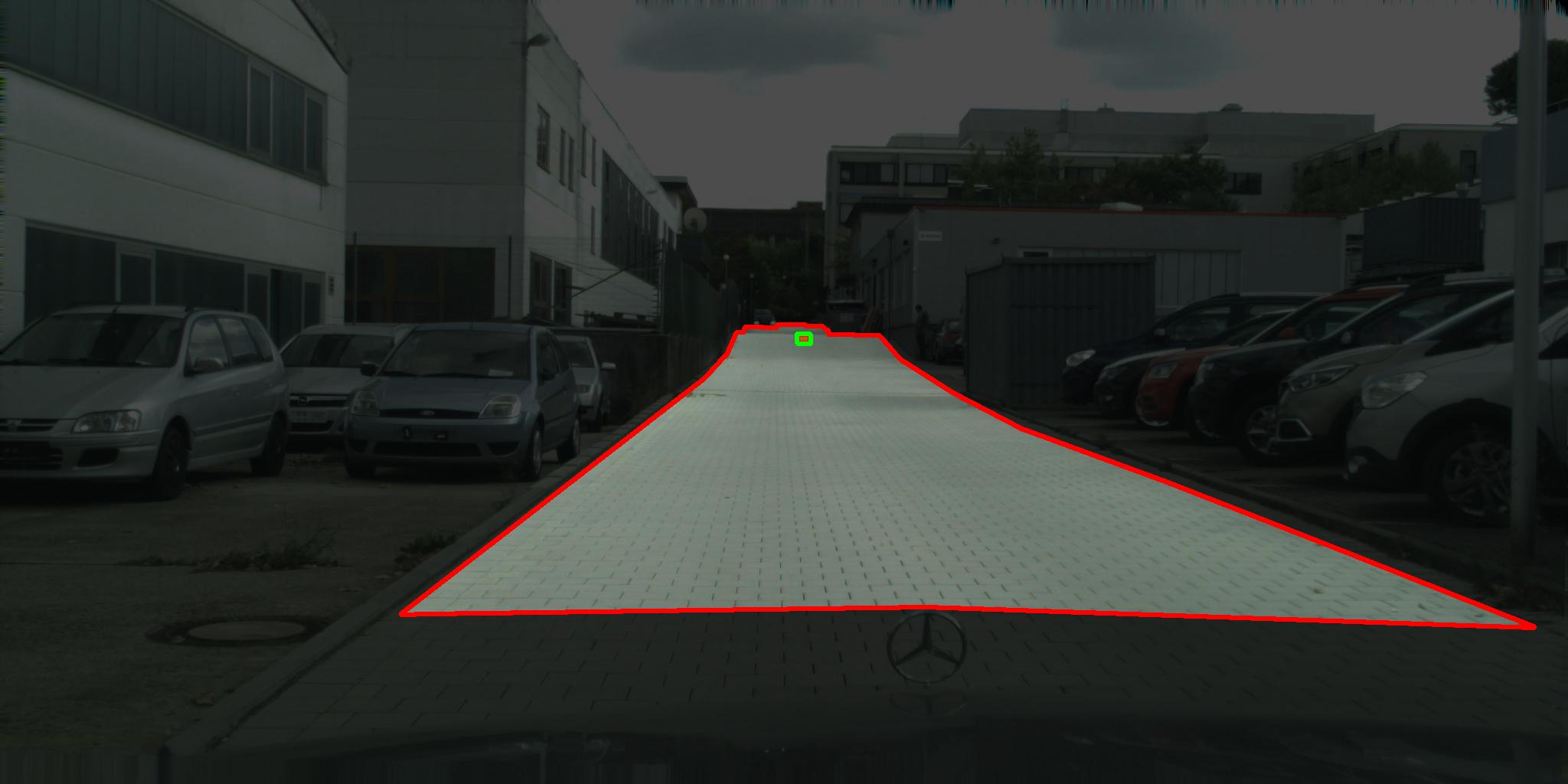}}%
    \subfloat[Frame 4]{\includegraphics[width=0.25\textwidth]{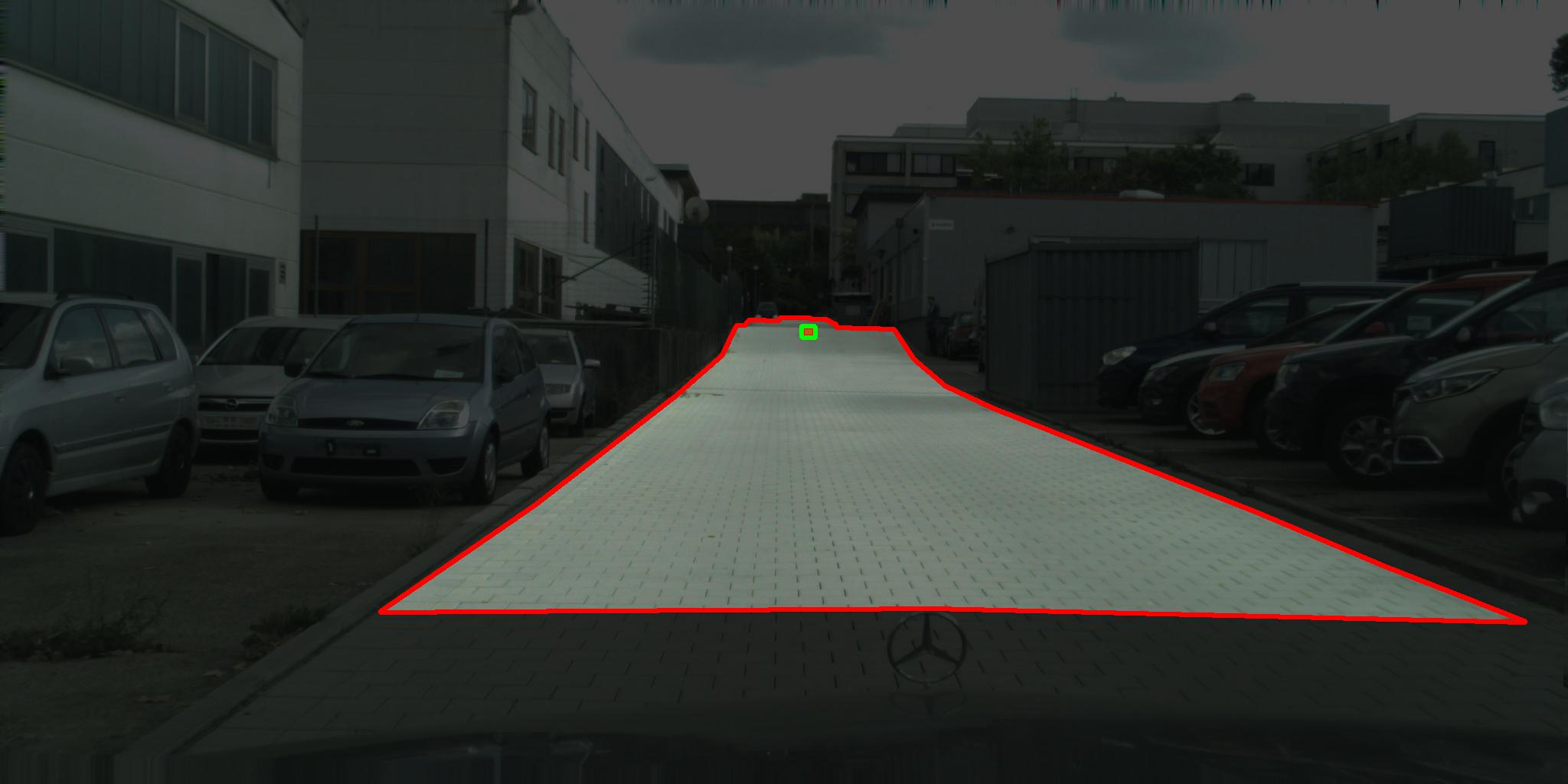}}%
    \caption{Four example images of densely sampled frames from a video sequence (of 18 frames in total) with ground truth annotation in the LostAndFound test set. Due to this sampling, LostAndFound achieve their high number of images but, as shown in this figure, several images are nearly identical.} 
    \label{fig:laf-frames}
\end{figure*}

\begin{figure}
    \centering
    \captionsetup[subfigure]{labelformat=empty, position=top}
    \subfloat[LostAndFound sequence frame 1]{\includegraphics[width=0.4\textwidth]{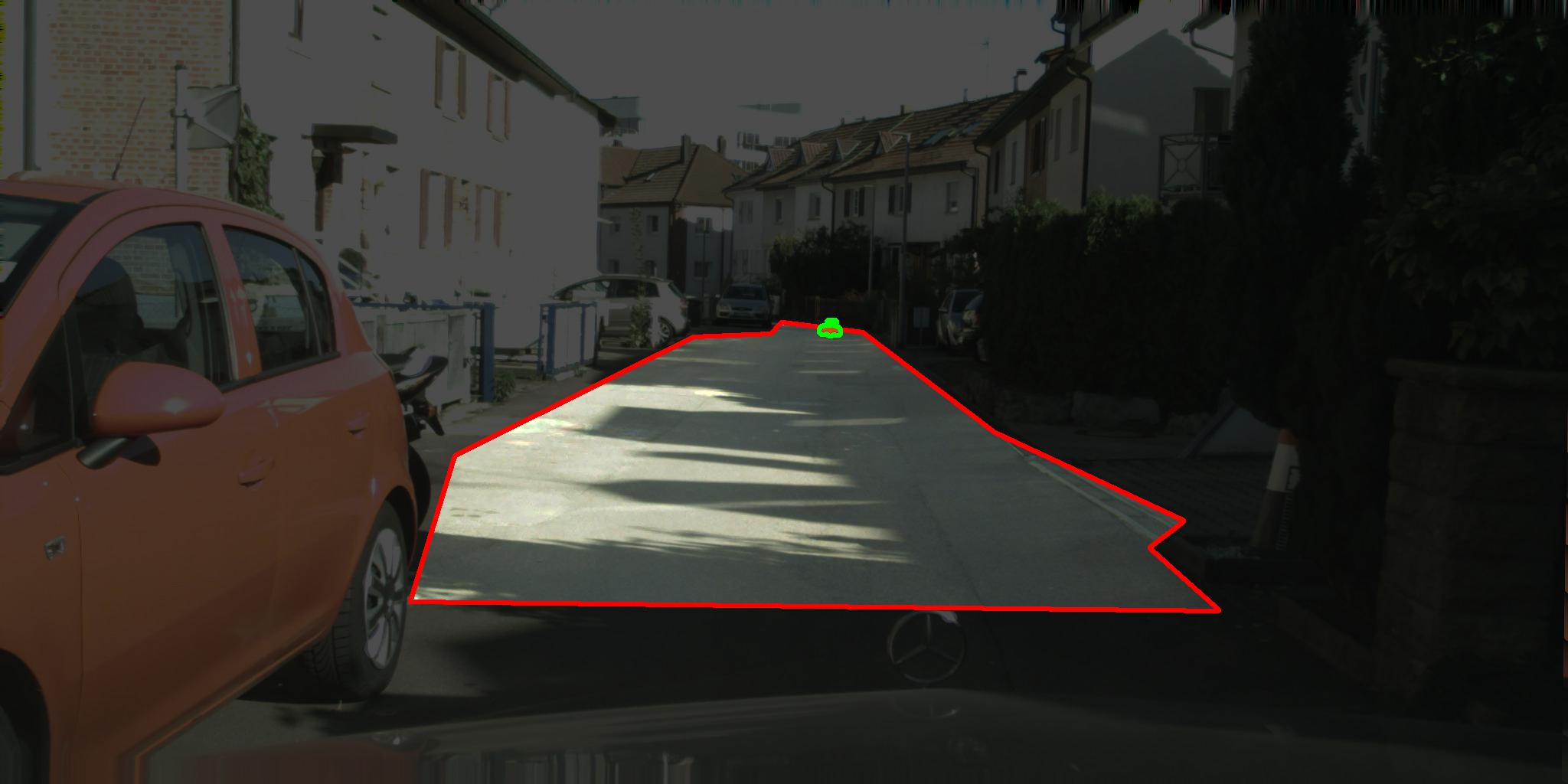}}~
    \subfloat[LostAndFound sequence frame 18]{\includegraphics[width=0.4\textwidth]{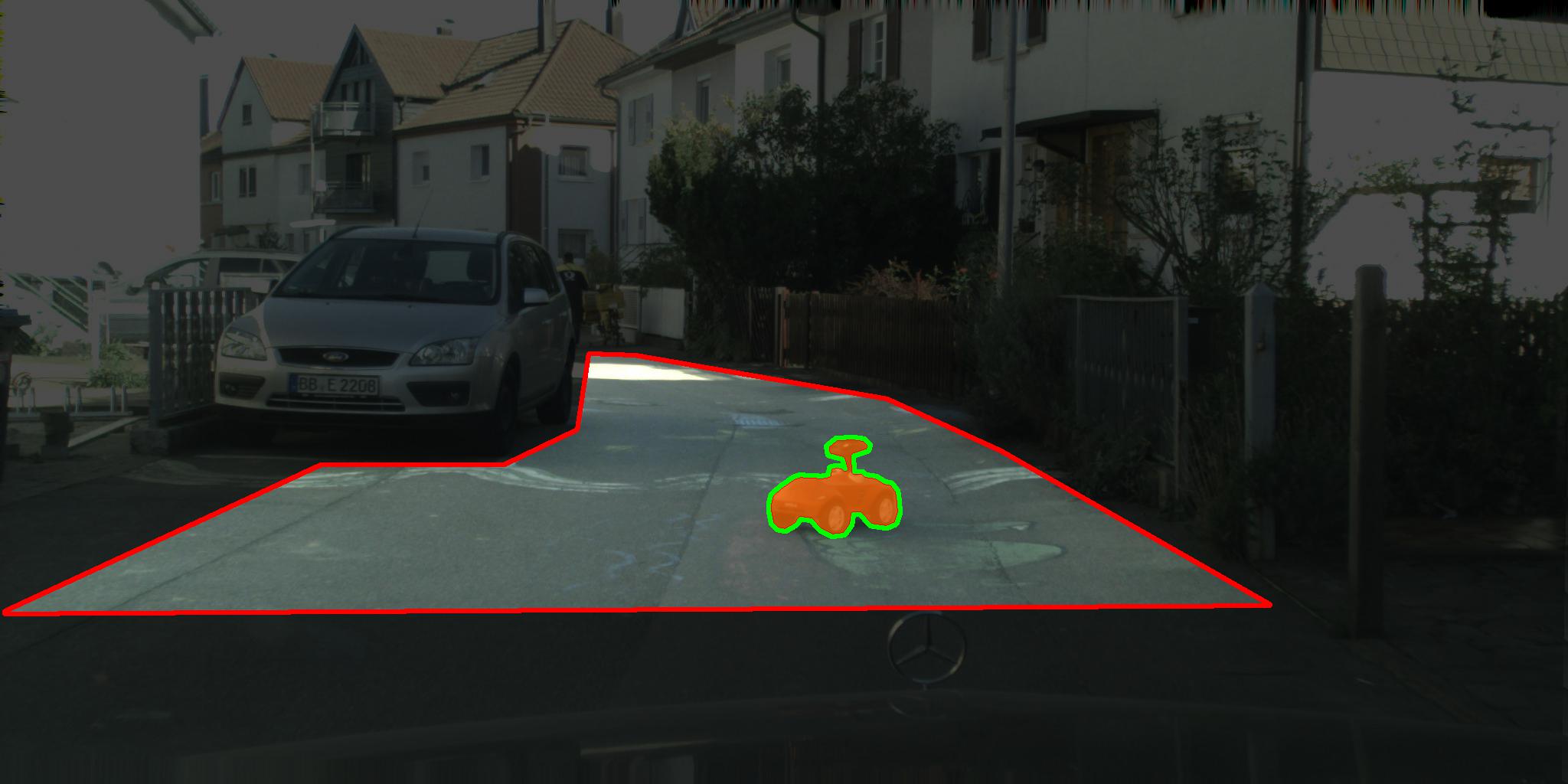}}\\
    \captionsetup[subfigure]{labelformat=empty, position=bottom}
    \subfloat[RoadObstacle21 sequence frame 1 ]{\includegraphics[width=0.4\textwidth]{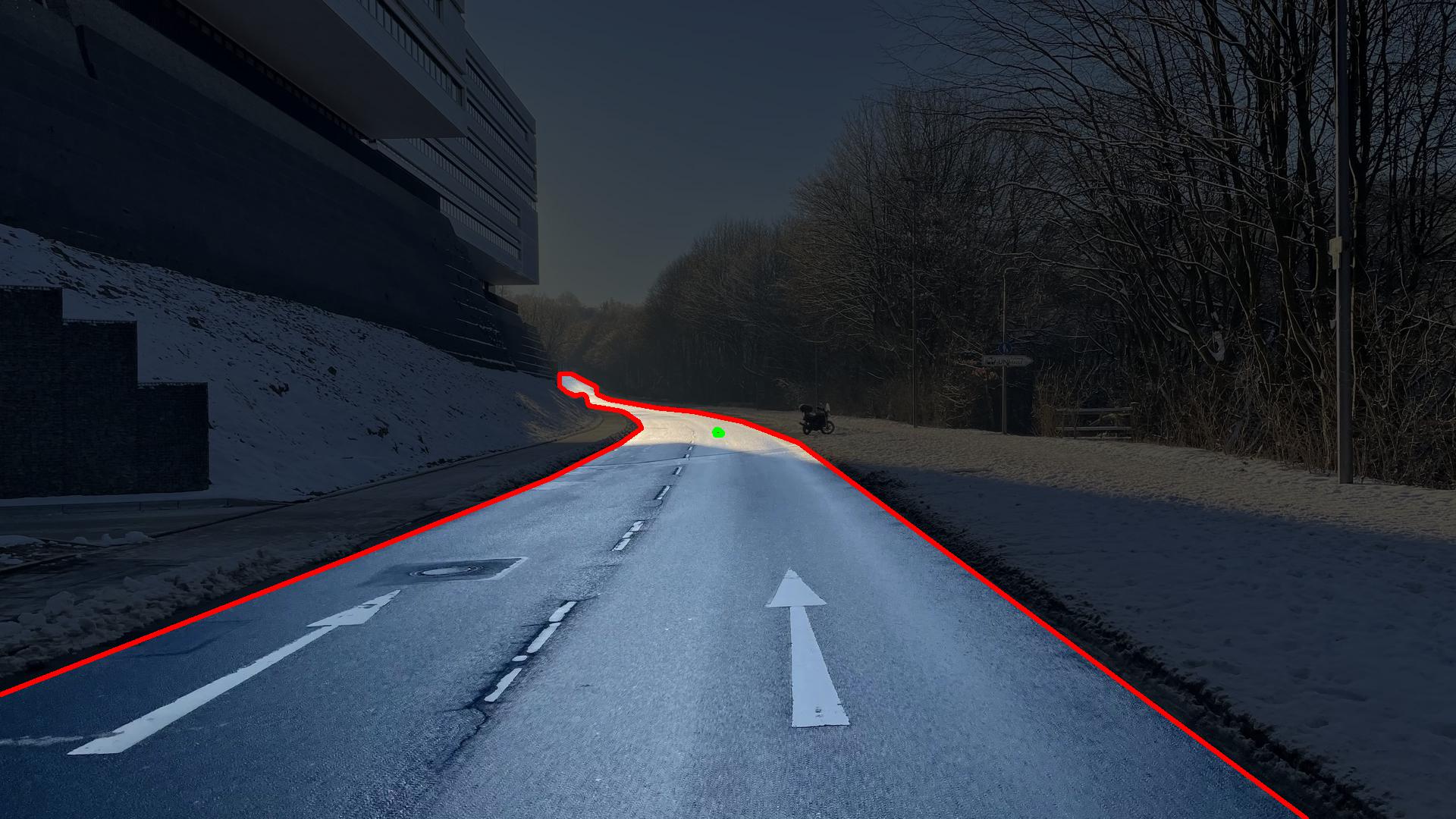}}~
    \subfloat[RoadObstacle21 sequence frame 6 ]{\includegraphics[width=0.4\textwidth]{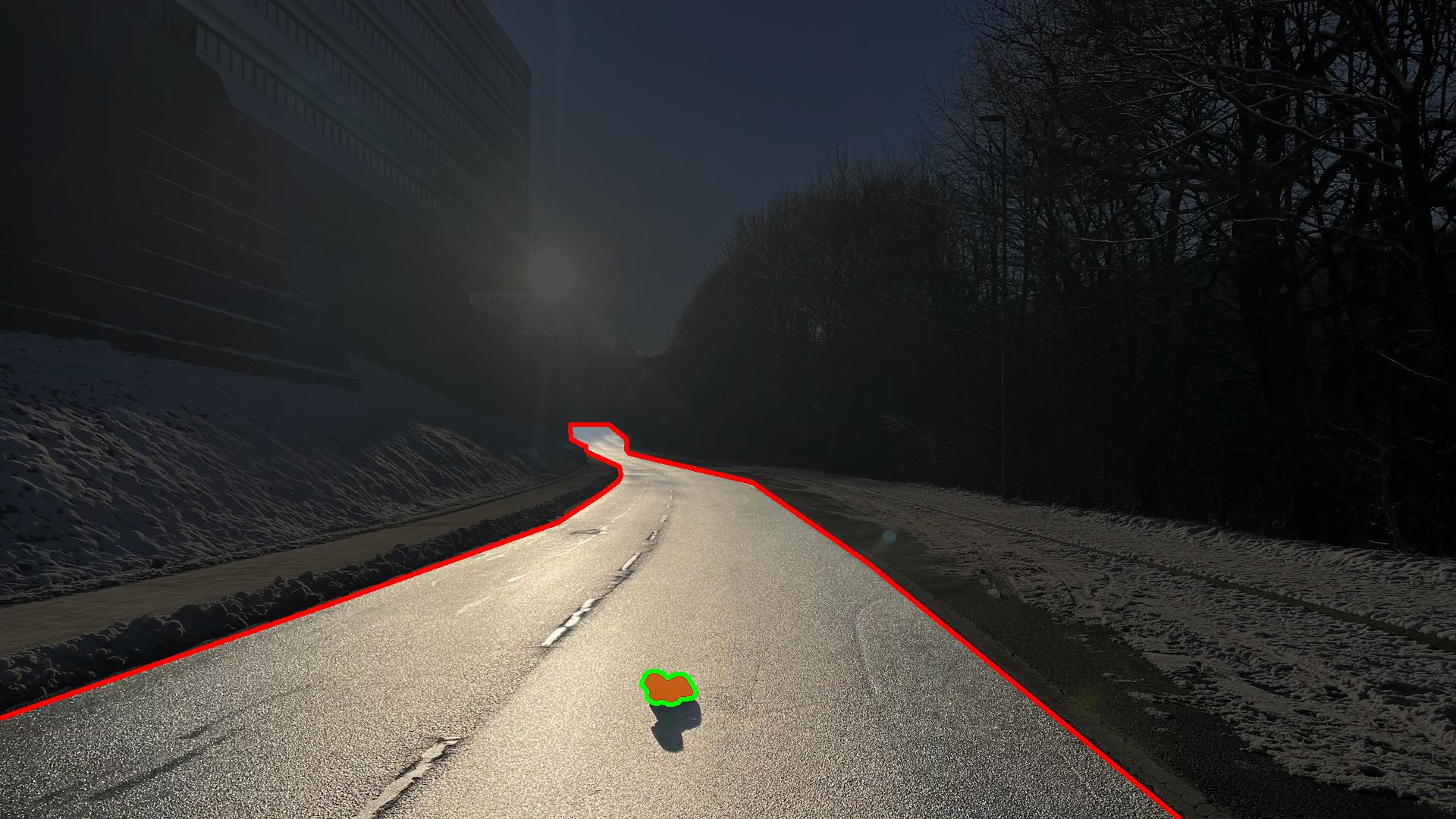}}
    \caption{Comparison of one sequence from the LostAndFound test set (top row) and one sequence from the RoadObstacle21 test set (bottom row). 
    In this figure, the first and last frame of a video sequence which are included in the respective test set, are shown.
    We observe that in this LostAndFound example 18 images of one sequence are included in the test while in RoadObstacle21 at most 6 frames are included (which differ significantly in lighting in this example).
    } 
    \label{fig:laf-obstacle-frames}
\end{figure}

Regarding the dataset size, LostAndFound achieves their high number of images by densely sampling from video sequences. Consequently, some images depict nearly identical scenes (same environment and obstacle combination with the obstacle approximately at the same distance), see \eg \Cref{fig:laf-frames}. In RoadObstacle21 the number of different environment and obstacle combinations is considerably higher due to the wide variety of 31 object types in the dataset. If multiple images depict the same scene, we made sure that the distance to the obstacle (and therefore the size of the obstacle in the image) varies noticeably from image to image, \cf \Cref{fig:laf-obstacle-frames}.

\subsection{LiDAR Guided Small Obstacle Dataset} \label{sec:eval-small-obstacle}

\begin{figure*}[ht!]
    \centering
    \subfloat[Input image]{\includegraphics[width=0.32\textwidth]{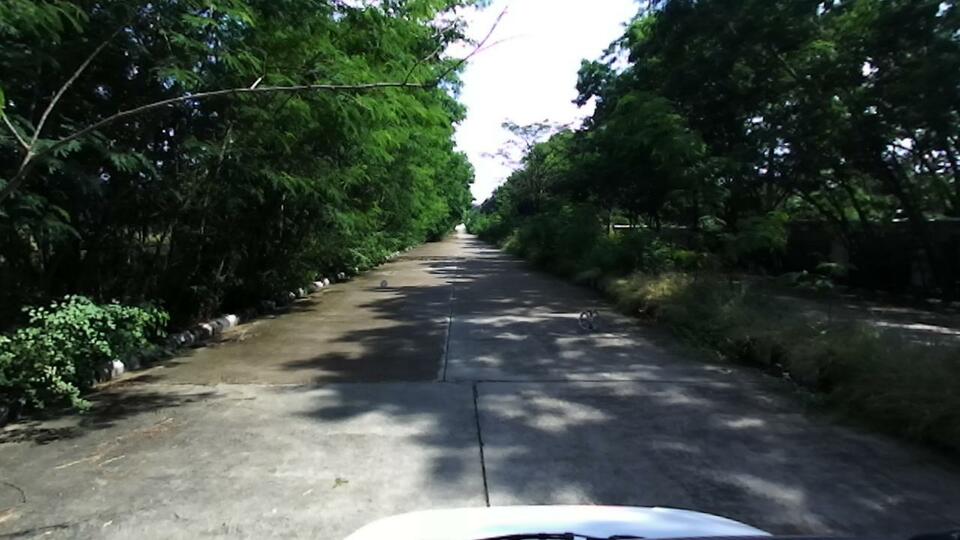}}
    \subfloat[Ground truth]{\includegraphics[width=0.32\textwidth]{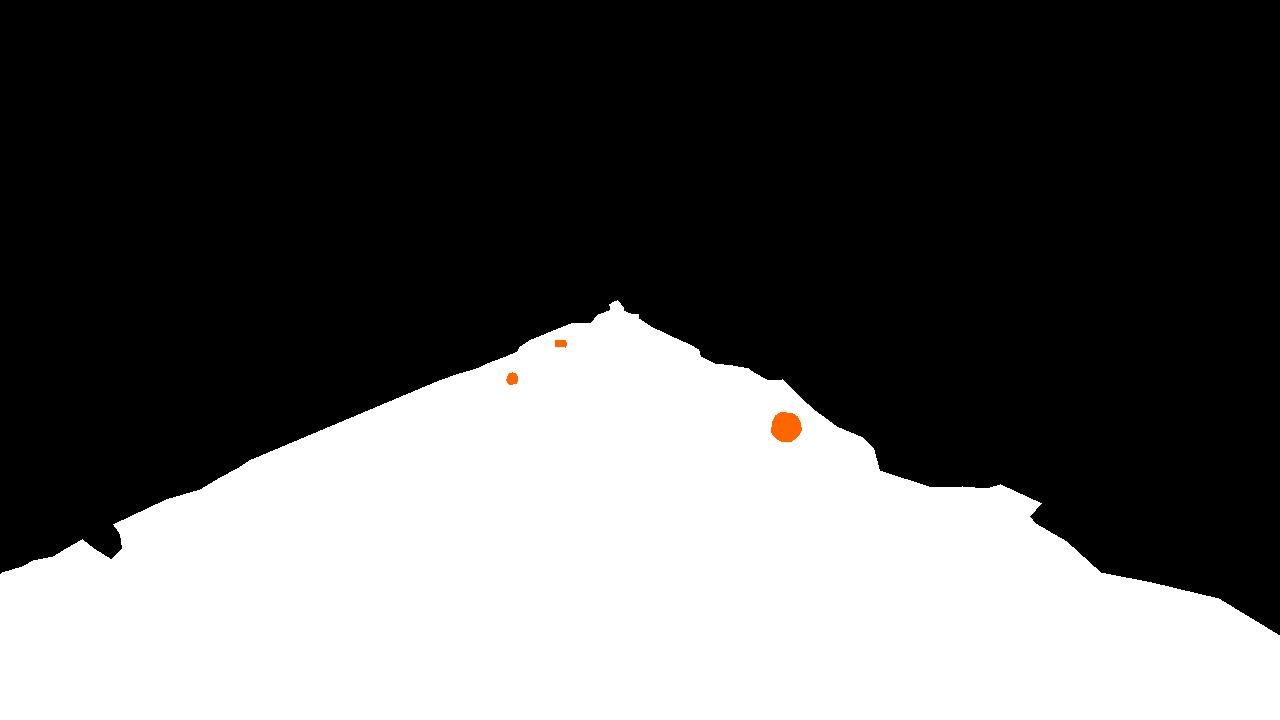}}
    \subfloat[Maximized entropy]{\includegraphics[width=0.32\textwidth]{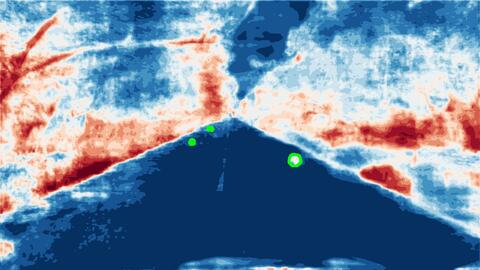}}
    \caption{An example image (a) from the Small Obstacle dataset with the corresponding ground truth annotation (b) and an obstacle score heatmap obtained with maximized entropy (c). Here, the obstacles are barely visible in the input image due to their size and the scene's illumination, that is why camera-only based segmentation techniques tend to fail for this dataset.}
    \label{fig:sod}
\end{figure*}

The third publicly available dataset to which we applied our benchmark suite is the LiDAR guided Small obstacle Segmentation dataset \cite{Singh2020-fw}, which can be viewed as a reference dataset for our obstacle track. The results corresponding to this dataset are given in \cref{tab:table7}. In general, the given set of methods exhibits poor performance on this dataset. More precisely, obstacles are mostly overlooked, \eg SynBoost as best-performing method still misses 1100 of 1203 components in total at the lowest sIoU threshold $\tau=0.25$. As the LiDAR guided Small obstacle Segmentation dataset rather focuses on the challenge of detecting obstacles via multiple sensors, including LiDAR, the camera images of this dataset are purposely challenging, \eg due to low illumination, blurry images and barely visible obstacles. \Cref{fig:sod} shows an example of this dataset, which highlights the difficulty of anomaly detection. This dataset can easily be included into our benchmark and it also fits the obstacle track, however, from our experiments we conclude that this dataset is less suitable to camera-only obstacle segmentation as obstacles are not well captured via cameras.

\subsection{CAOS BDD-Anomaly} \label{sec:bdd-anomaly-supp}

\begin{figure*}[ht!]
\centering
\subfloat[Vegetation as train]{\includegraphics[width=0.32\textwidth]{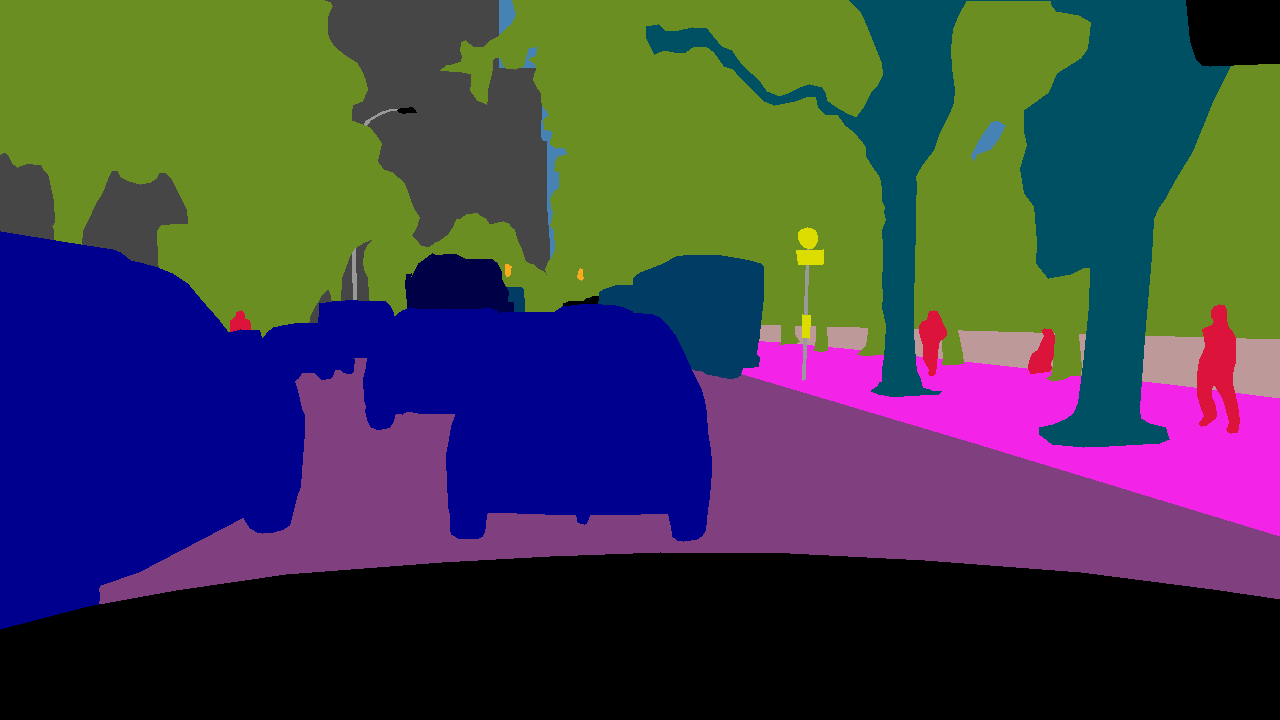}}
\subfloat[Cars as bicycle]{\includegraphics[width=0.32\textwidth]{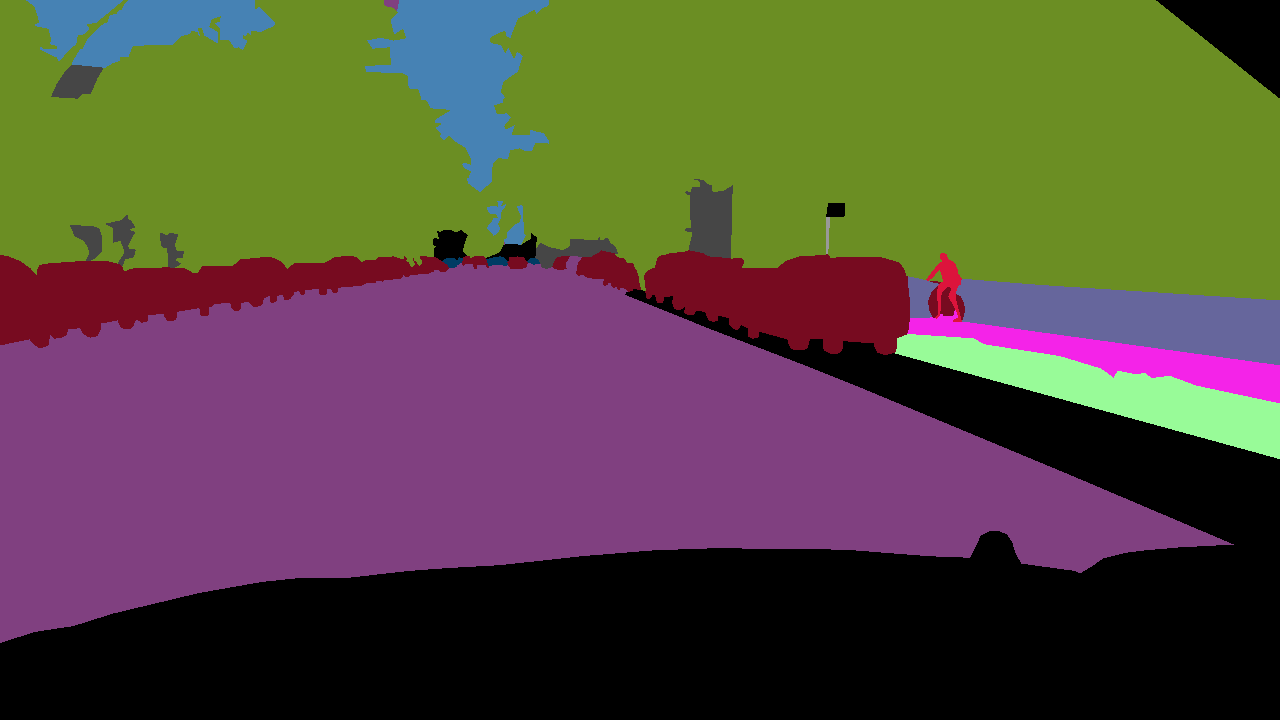}}
\subfloat[Bus as bicycle]{\includegraphics[width=0.32\textwidth]{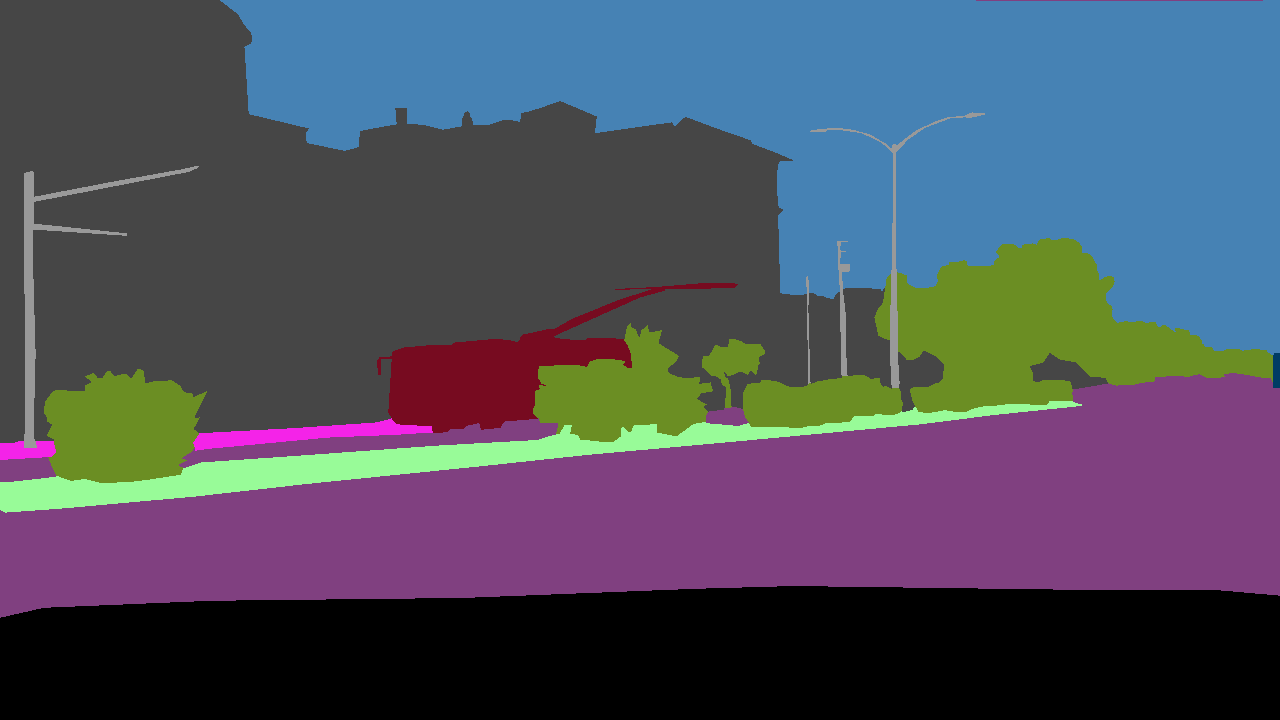}}
\caption{Some examples of labeling mistakes in the CAOS BDD-Anomaly dataset, where in-distribution objects are incorrectly annotated as anomaly, \ie train and bicycle.}
\label{fig:BDD-labeling-mistakes}
\end{figure*}

The CAOS BDD-Anomaly dataset \cite{hendrycks2020scaling} consists of images sourced from BDD100k \cite{yu2020bdd100k}. In order to create an anomaly segmentation dataset, the authors split the BDD100k data such that images with motorcycles, bicycles and trains are separated from the rest. These left out objects are then considered as anomalies. We do not perform any experiments on CAOS BDD-Anomaly since the considered anomalous objects are not strictly unknown. They also appear in Cityscapes \cite{Cordts2016Cityscapes} on which most semantic segmentation models are trained. Moreover, we find several labeling mistakes that hinder proper evaluation of anomaly segmentation performance, see \cref{fig:BDD-labeling-mistakes}.

\section{Evaluation per Environment Category}
\begin{figure*}[ht]
\centering
\captionsetup[subfigure]{labelformat=empty, position=bottom}

\subfloat[road cracked]{\includegraphics[width=0.33\textwidth]{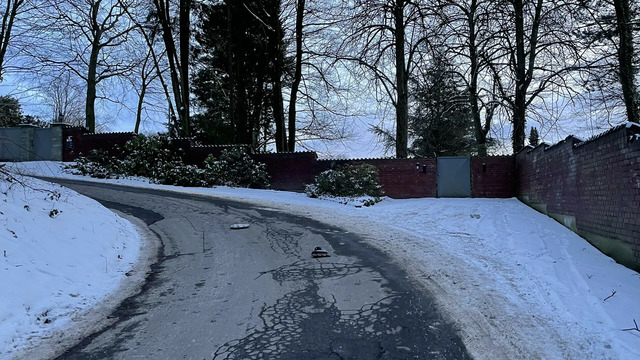}}
\subfloat[asphalt dark]{\includegraphics[width=0.33\textwidth]{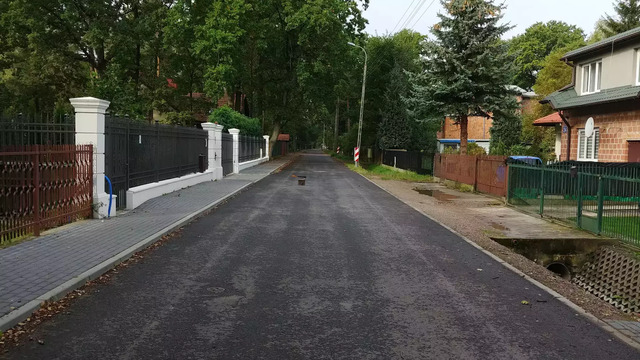}}
\subfloat[road gravel]{\includegraphics[width=0.33\textwidth]{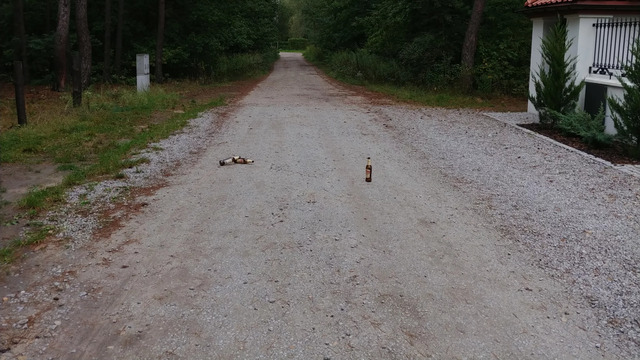}} \\
\subfloat[asphalt gray]{\includegraphics[width=0.33\textwidth]{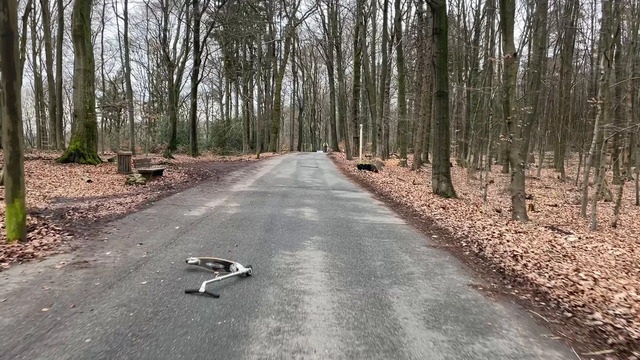}}
\subfloat[motorway]{\includegraphics[width=0.33\textwidth]{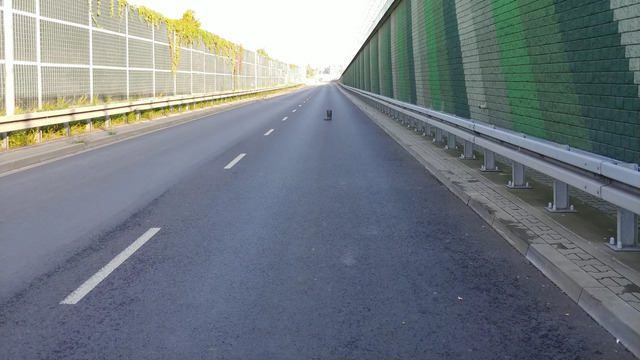}}
\subfloat[sun reflection]{\includegraphics[width=0.33\textwidth]{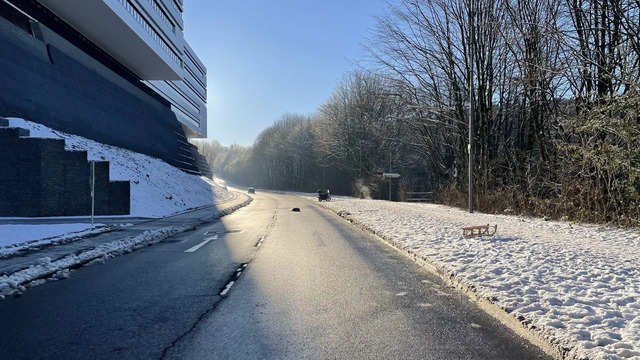}} \\
\subfloat[road bricks]{\includegraphics[width=0.33\textwidth]{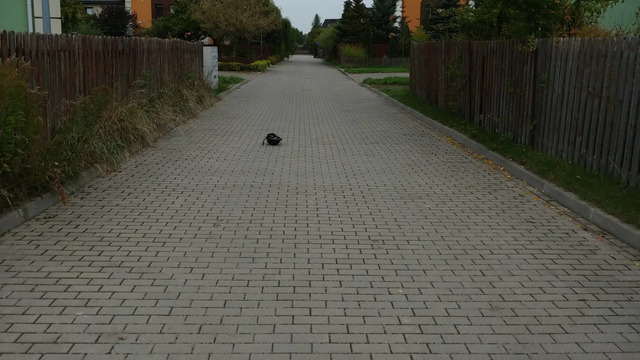}}
\subfloat[asphalt night]{\includegraphics[width=0.33\textwidth]{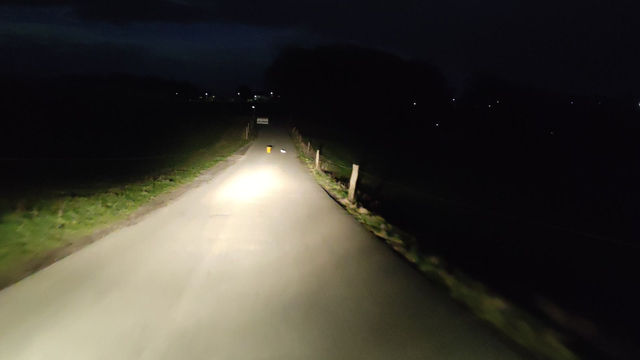}}
\subfloat[snowstorm]{\includegraphics[width=0.33\textwidth]{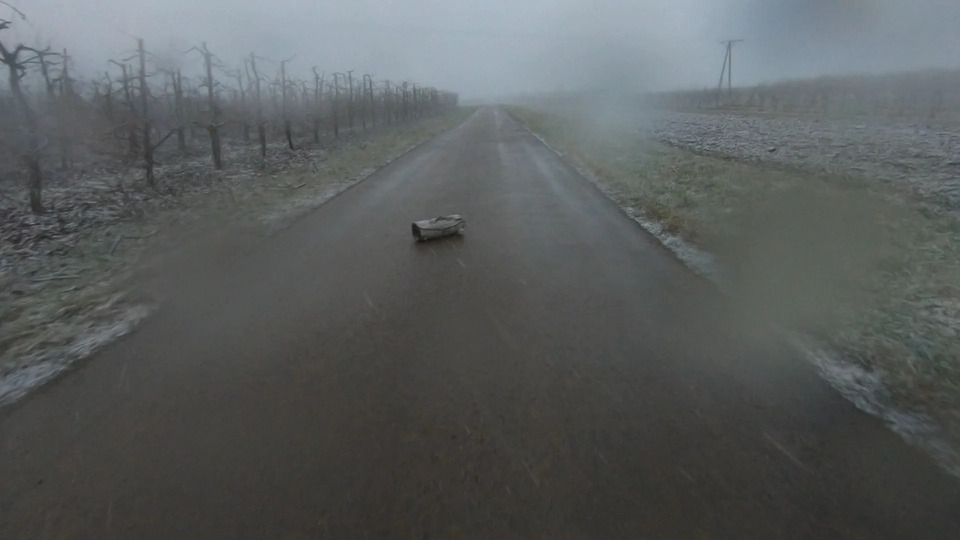}}

\caption{The scenes of our RoadObstacle21 dataset feature a variety of road surfaces.}
\label{fig:obstacle-track-scenes}
\end{figure*}

We already emphasized that in our RoadObstacle21 dataset a wide variety of road surfaces are available, representing different scenes which might pose unique challenges.
In this section, we provide more insights by evaluating our set of methods on each of these surfaces. In total, we split our datasets into 9 different scenes, shown in \cref{fig:obstacle-track-scenes}:
\begin{enumerate}
    \item cracked road, surrounded by snow (road cracked)
    \item dark asphalt after rain, with leaves (asphalt dark)
    \item gravel road, no snow (road gravel)
    \item gray asphalt in village and forest (asphalt gray)
    \item motorway with side railing (motorway)
    \item sun reflection off wet road (sun reflection)
    \item road made of bricks (road bricks)
    \item night images (asphalt night)
    \item and snowstorm images~.
\end{enumerate}

We evaluate each subset using our benchmark suite and report the results in \cref{tab:table_scenes}. This more detailed evaluation shows that the reported set of methods perform differently across the data splits, with no method having consistent performance on each of these subsets. Our dataset offers extra difficulty caused by the diversity of road texture, surrounding environments, weather and lighting variations. Cracks and leaves may trigger false positives, and a gravel or wet road surface may itself be sufficiently different from training images to be mistaken for an anomaly.

\section{Evaluation for Different Component Sizes} \label{sec:eval-comp-size-supp}

In this section we provide further insights of the segmentation quality of ground truth components in RoadAnomaly21 and RoadObstacle21. To this end, we conduct a more fine-grained analysis by grouping ground truth components into size intervals and perform the evaluation for each size interval separately. In total, RoadAnomaly21 contains 259 ground truth components, ranging in size from 122 to 883,319 pixels. RoadObstacle21 contains 388 obstacles ranging from 18 up to 77,435 pixels. For each dataset we divide these components into eight size intervals such that each interval contains same number of components. 

\begin{figure*}[ht]
    \centering
    \captionsetup[subfigure]{labelformat=empty}
    \includegraphics[width=0.9\linewidth]{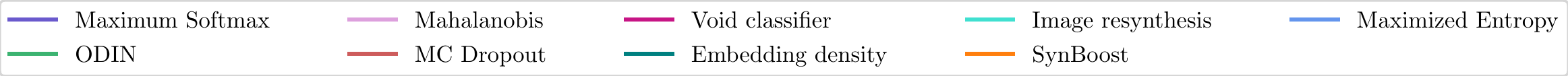}\\
    \vspace{-.7\baselineskip}
    \subfloat[RoadAnomaly21, each size interval contains 32 components except the first one (very left) that contains 35 components
    ]{\includegraphics[width=0.9\linewidth]{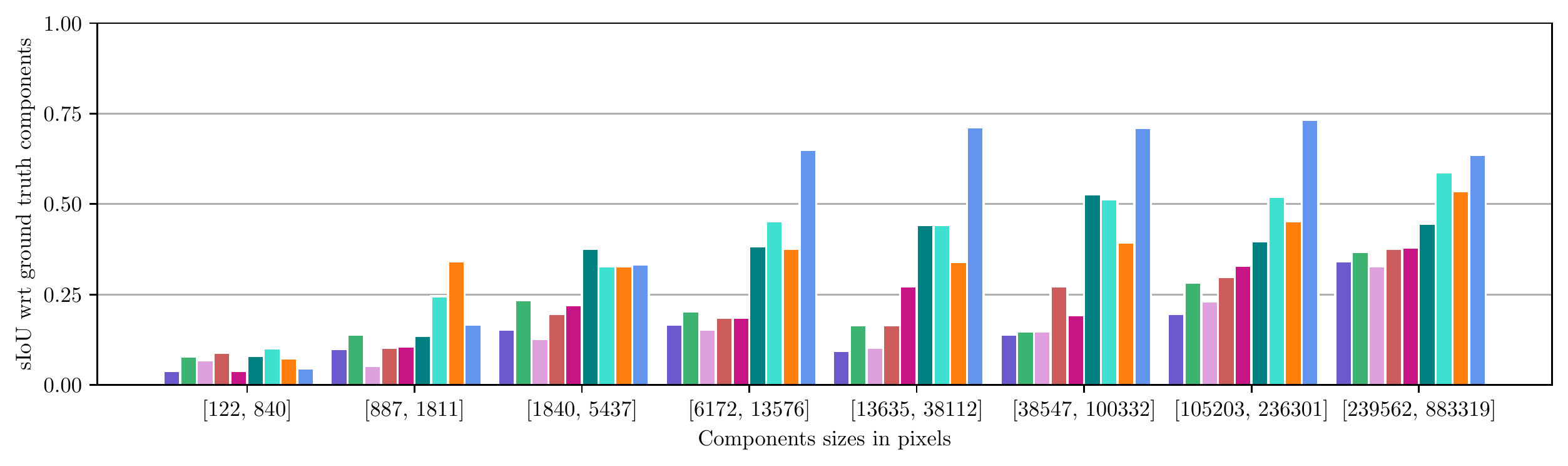}} \\
    \vspace{-1.07\baselineskip}
    \subfloat[RoadObstacle21, each size interval contains 46 components except the first one (very left) that contains 66 components]{\includegraphics[width=0.95\linewidth]{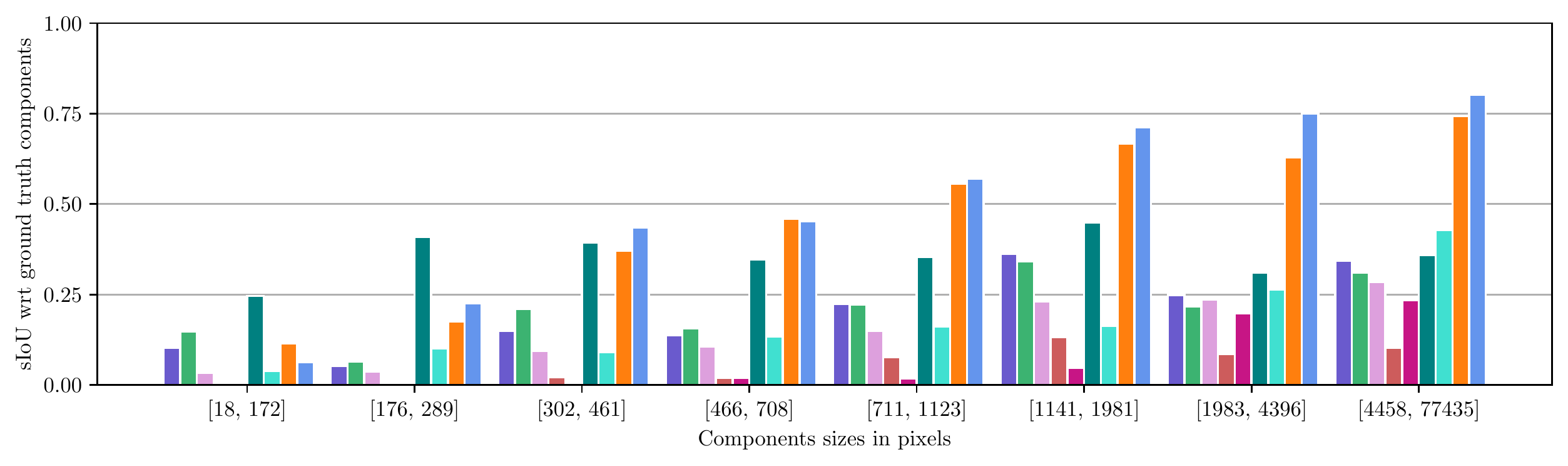}} ~
    \caption{Comparison of the averaged $\mathrm{sIoU}$ \wrt ground truth components within a certain range of the components size, produced by the methods discussed in \cref{sec:methods} and \cref{sec:methods-detail}.}
    \label{fig:size-evaluation}
\end{figure*}

In \cref{fig:size-evaluation}, we report the averaged $\mathrm{sIoU}$ (\cref{eq:adj_iou_gt}) \wrt the ground truth components within each size interval. As illustrated in this figure, we observe a positive correlation of $\mathrm{sIoU}$ with the component size. Especially in RoadObstacle21, methods designed for the task of anomaly segmentation like maximized entropy or SynBoost perform significantly better than the other approaches. 

\begin{figure*}[ht]
    \centering
    \captionsetup[subfigure]{labelformat=empty}
    \includegraphics[width=0.9\linewidth]{figures/legend_SizeEval.pdf}\\
    \vspace{-.7\baselineskip}
    \subfloat[RoadAnomaly21, each size interval contains 32 components except the first one (very left) that contains 35 components
    ]{\includegraphics[width=0.9\linewidth]{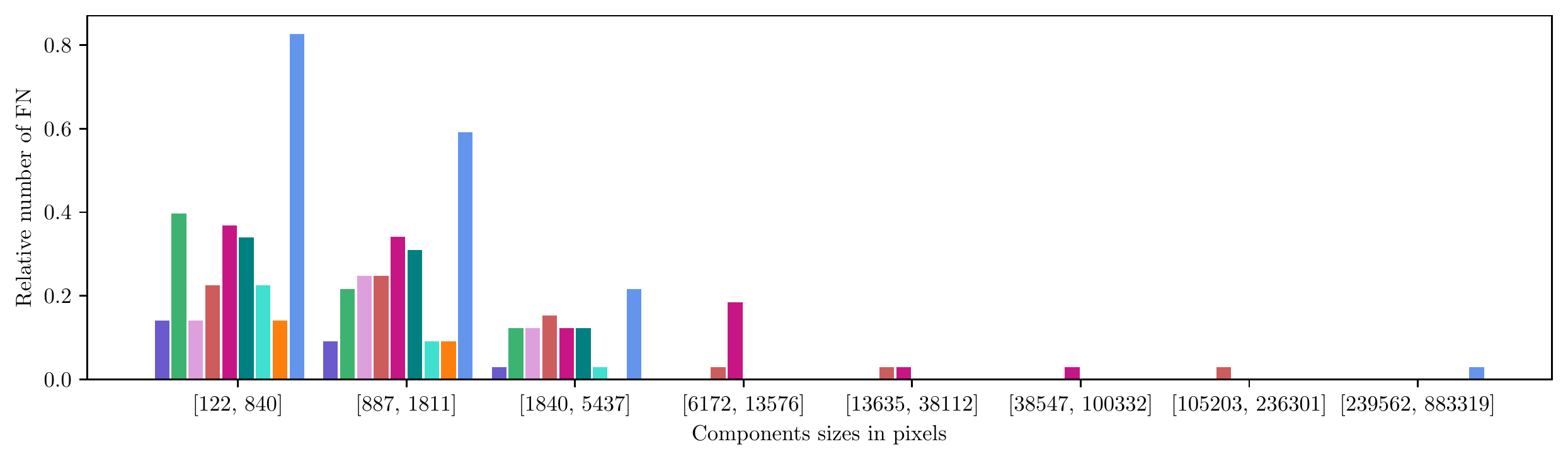}} \\
    \vspace{-1.07\baselineskip}
    \subfloat[RoadObstacle21, each size interval contains 46 components except the first one (very left) that contains 66 components]{\includegraphics[width=0.95\linewidth]{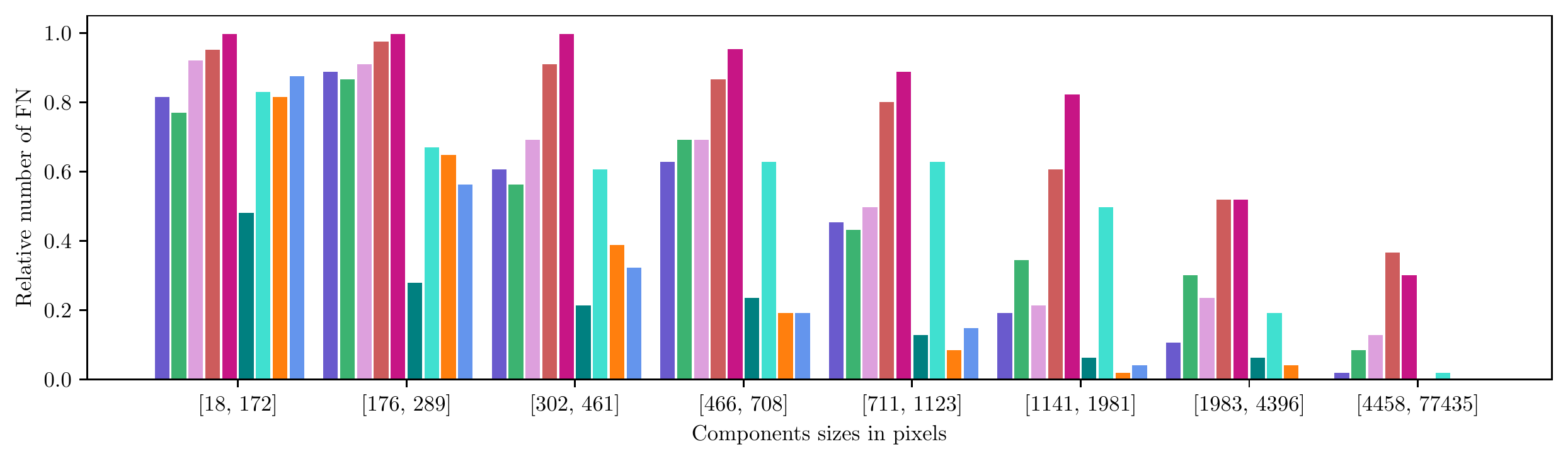}} ~
    \caption{Comparison of the relative number of $\mathrm{FN}$ to $\mathrm{TP}$ at threshold $\tau=0$, \ie the fraction of overlooked components to the total number of ground truth components within a certain range of the components size. The evaluated methods are discussed in \cref{sec:methods} and \cref{sec:methods-detail}.}
    \label{fig:size-evaluation-fn}
\end{figure*}

In addition, we consider the amount of entirely neglected components, meaning the objects for which not even one pixel is detected. To do so, we measure the relative ratio of FN to all ground truth components within different object size intervals, see \cref{fig:size-evaluation-fn}. As a threshold, therefore, for discriminating between FN and TP, we choose $\tau = 0$, \ie a ground truth component is considered as TP if at least one of its pixels is detected by the respective method. Indeed, we observe a negative correlation of the number of FN with the component size, but even more conspicuous is the amount of totally overlooked components of small size.
This analysis shows the challengingness of anomaly segmentation, particularly for small obstacles at component-level, and emphasizes the need for further research in this direction. 

\section{Evaluation per Object Category} \label{sec:eval-cate-supp}

As part of our benchmark, we also provide an evaluation with respect to different object categories. An exemplary evaluation with the given set of methods is provided in \cref{tab:table_categories}. In particular, the methods specifically designed for anomaly segmentation perform worse on the vehicle category than on the other ones. This general trends shows that our choice of vehicles, including classes such as jet ski, rickshaw and carriage, is rather challenging. This additional dimension of granularity offers further insight to users of our benchmark such that one can identify the drawbacks of an anomaly segmentation method under inspection.

\newpage
\section*{Tables and Figures}

\begin{table*}[ht]
    \setlength{\tabcolsep}{2pt}
    \begin{center}
    \scalebox{0.68}{
\begin{tabular}{l||c||acc||cc|ccc|ccc|ccc|g}
    \toprule
    & & \multicolumn{3}{c||}{Pixel-level} & \multicolumn{12}{c}{Component-level}  \\
    \midrule
    \midrule
    & requires & \multicolumn{3}{c||}{Anomaly scores} & $k\in\mathcal{K}$ & $\hat{k}\in\mathcal{\hat{K}}$ & \multicolumn{3}{c|}{$\tau=0.25$} & \multicolumn{3}{c|}{$\tau=0.50$} & \multicolumn{3}{c|}{$\tau=0.75$} &  \multicolumn{1}{c}{}\\
    Method & OoD data & \multicolumn{1}{c}{AuPRC $\uparrow$} & FPR$_{95}$ $\downarrow$ & $F_1^*\uparrow$ & $\overline{\mathrm{sIoU}}$ $\uparrow$ & $\overline{\mathrm{PPV}}$ $\uparrow$ & FN $\downarrow$ & FP $\downarrow$ & $F_1\uparrow$ & FN $\downarrow$ & FP $\downarrow$ & $F_1\uparrow$ & FN $\downarrow$ & FP $\downarrow$ & $F_1\uparrow$ & \multicolumn{1}{c}{$\overline{F_1}\uparrow$}\\
    \midrule
Maximum softmax \cite{Hendrycks2017msp} & \xmark &   40.4 &                                              60.2 &                                            42.6 &                                               34.2 &                                               18.6 &                                                  7 &                                                 37 &                                               29.0 &                                                 10 &                                                 39 &                                               19.7 &                                                 15 &                                                 44 &                                                3.3 &                                               16.7 \\
\bottomrule
    \end{tabular}
    }
    \end{center}
    \caption{Benchmark results for the RoadAnomaly21 validation set. This dataset contains 16 ground truth components.}
    \label{tab:table-val-anomaly}
\end{table*}

\begin{table*}[ht]
    \setlength{\tabcolsep}{2pt}
    \begin{center}
    \scalebox{0.68}{
    \begin{tabular}{l||c||acc||cc|ccc|ccc|ccc|g}
    \toprule
    & & \multicolumn{3}{c||}{Pixel-level} & \multicolumn{12}{c}{Component-level}  \\
    \midrule
    \midrule
    & requires & \multicolumn{3}{c||}{Anomaly scores} & $k\in\mathcal{K}$ & $\hat{k}\in\mathcal{\hat{K}}$ & \multicolumn{3}{c|}{$\tau=0.25$} & \multicolumn{3}{c|}{$\tau=0.50$} & \multicolumn{3}{c|}{$\tau=0.75$} &  \multicolumn{1}{c}{}\\
    Method & OoD data & \multicolumn{1}{c}{AuPRC $\uparrow$} & FPR$_{95}$ $\downarrow$ & $F_1^*\uparrow$ & $\overline{\mathrm{sIoU}}$ $\uparrow$ & $\overline{\mathrm{PPV}}$ $\uparrow$ & FN $\downarrow$ & FP $\downarrow$ & $F_1\uparrow$ & FN $\downarrow$ & FP $\downarrow$ & $F_1\uparrow$ & FN $\downarrow$ & FP $\downarrow$ & $F_1\uparrow$ & \multicolumn{1}{c}{$\overline{F_1}\uparrow$}\\
    \midrule
Maximum softmax \cite{Hendrycks2017msp} & \xmark &    43.4 &                                                3.8 &                                             53.7 &                                               40.6 &                                               20.9 &                                                 20 &                                                113 &                                               27.3 &                                                 22 &                                                113 &                                               25.4 &                                                 32 &                                                117 &                                               14.9 &                                               22.5 \\
\bottomrule
    \end{tabular}
    }
    \end{center}
    \caption{Benchmark results for the RoadObstacle21 validation set. This dataset contains 45 ground truth objects.}
    \label{tab:table-val-obstacle}
\end{table*}

\begin{table*}[ht]
    \setlength{\tabcolsep}{2pt}
    \begin{center}
    \scalebox{0.68}{
    \begin{tabular}{l||c||acc||cc|ccc|ccc|ccc|g}
    \toprule
    & & \multicolumn{3}{c||}{Pixel-level} & \multicolumn{12}{c}{Component-level}  \\
    \midrule
    \midrule
    & requires & \multicolumn{3}{c||}{Anomaly scores} & $k\in\mathcal{K}$ & $\hat{k}\in\mathcal{\hat{K}}$ & \multicolumn{3}{c|}{$\tau=0.25$} & \multicolumn{3}{c|}{$\tau=0.50$} & \multicolumn{3}{c|}{$\tau=0.75$} &  \multicolumn{1}{c}{}\\
    Method & OoD data & \multicolumn{1}{c}{AuPRC $\uparrow$} & FPR$_{95}$ $\downarrow$ & $F_1^*\uparrow$ & $\overline{\mathrm{sIoU}}$ $\uparrow$ & $\overline{\mathrm{PPV}}$ $\uparrow$ & FN $\downarrow$ & FP $\downarrow$ & $F_1\uparrow$ & FN $\downarrow$ & FP $\downarrow$ & $F_1\uparrow$ & FN $\downarrow$ & FP $\downarrow$ & $F_1\uparrow$ & \multicolumn{1}{c}{$\overline{F_1}\uparrow$}\\
    \midrule
    Maximum softmax \cite{Hendrycks2017msp} & \xmark &  5.6 & 40.5 & 12.1 & 3.5 & 9.5 & 152 & 199 & 6.9 & 164 & 199 & 0.5 & 165 & 199 & 0.0 & 1.8 \\
    ODIN \cite{liang18odin} & \xmark & 15.5 & 38.4 & 22.5 & 9.9 & 21.9 & 139 & 142 & 15.6 & 146 & 142 & 11.7 & 163 & 143 & 1.3 & 9.7 \\
    Mahalanobis \cite{Lee2018mahala} & \xmark & 32.9 & \textbf{8.7} & 37.3 & 19.6 & 29.4 & 111 & 145 & 29.7 & 132 & 147 & 19.1 & 155 & 157 & 6.0 & 19.2 \\
    MC dropout \cite{mukhoti2019bdl} & \xmark & 14.4 & 47.8 & 20.0 & 4.8 & 18.1 & 149 & 120 & 10.6 & 160 & 120 & 3.4 & 164 & 121 & 0.7 & 4.3 \\
\bottomrule
    \end{tabular}
    }
    \end{center}
    \caption{Benchmark results for the Fishyscapes LostAndFound validation set. This dataset contains 165 ground truth objects.}
    \label{tab:table5}
\end{table*}

\begin{table*}[ht]
    \setlength{\tabcolsep}{2pt}
    \begin{center}
    \scalebox{0.68}{
    \begin{tabular}{l||c||acc||cc|ccc|ccc|ccc|g}
    \toprule
    & & \multicolumn{3}{c||}{Pixel-level} & \multicolumn{12}{c}{Component-level}  \\
    \midrule
    \midrule
    & requires & \multicolumn{3}{c||}{Anomaly scores} & $k\in\mathcal{K}$ & $\hat{k}\in\mathcal{\hat{K}}$ & \multicolumn{3}{c|}{$\tau=0.25$} & \multicolumn{3}{c|}{$\tau=0.50$} & \multicolumn{3}{c|}{$\tau=0.75$} &  \multicolumn{1}{c}{}\\
    Method & OoD data & \multicolumn{1}{c}{AuPRC $\uparrow$} & FPR$_{95}$ $\downarrow$ & $F_1^*\uparrow$ & $\overline{\mathrm{sIoU}}$ $\uparrow$ & $\overline{\mathrm{PPV}}$ $\uparrow$ & FN $\downarrow$ & FP $\downarrow$ & $F_1\uparrow$ & FN $\downarrow$ & FP $\downarrow$ & $F_1\uparrow$ & FN $\downarrow$ & FP $\downarrow$ & $F_1\uparrow$ & \multicolumn{1}{c}{$\overline{F_1}\uparrow$}\\
    \midrule
    Maximum softmax \cite{Hendrycks2017msp} & \xmark & 30.1 &                                               33.2 &                                             32.5 &                                               14.2 &                                               62.2 &                                               1256 &                                                580 &                                               33.0 &                                               1575 &                                                602 &                                               11.0 &                                               1701 &                                                644 &                                                0.7 &                                               13.4 \\
\bottomrule
    \end{tabular}
    }
    \end{center}
    \caption{Benchmark results for the LostAndFound test-NoKnown dataset. This dataset contains 1709 ground truth objects.}
    \label{tab:table6}
\end{table*}

\begin{table*}[ht]
    \setlength{\tabcolsep}{2pt}
    \begin{center}
    \scalebox{0.68}{
    \begin{tabular}{l||c||acc||cc|ccc|ccc|ccc|g}
    \toprule
    & & \multicolumn{3}{c||}{Pixel-level} & \multicolumn{12}{c}{Component-level}  \\
    \midrule
    \midrule
    & requires & \multicolumn{3}{c||}{Anomaly scores} & $k\in\mathcal{K}$ & $\hat{k}\in\mathcal{\hat{K}}$ & \multicolumn{3}{c|}{$\tau=0.25$} & \multicolumn{3}{c|}{$\tau=0.50$} & \multicolumn{3}{c|}{$\tau=0.75$} &  \multicolumn{1}{c}{}\\
    Method & OoD data & \multicolumn{1}{c}{AuPRC $\uparrow$} & FPR$_{95}$ $\downarrow$ & $F_1^*\uparrow$ & $\overline{\mathrm{sIoU}}$ $\uparrow$ & $\overline{\mathrm{PPV}}$ $\uparrow$ & FN $\downarrow$ & FP $\downarrow$ & $F_1\uparrow$ & FN $\downarrow$ & FP $\downarrow$ & $F_1\uparrow$ & FN $\downarrow$ & FP $\downarrow$ & $F_1\uparrow$ & \multicolumn{1}{c}{$\overline{F_1}\uparrow$}\\
    \midrule
    Maximum softmax \cite{Hendrycks2017msp} & \xmark&                                                0.7 &                                               57.1 &                                               2.2 &                                                0.5 &                                                1.5 &                                               1196 &                                               1652 &                                                0.5 &                                               1202 &                                               1653 &                                                0.1 &                                               1203 &                                               1653 &                                                0.0 &                                                0.2 \\
\bottomrule
    \end{tabular}
    }
    \end{center}
    \caption{Benchmark results for the LiDAR guided Small obstacle Segmentation dataset. This dataset contains 1203 ground truth components in total.
    }
    \label{tab:table7}
\end{table*}

\begin{table*}[ht]
    \setlength{\tabcolsep}{2pt}
    \begin{center}
    \scalebox{0.68}{
    \begin{tabular}{l||c||acg|acg|acg|acg}
    \toprule
    & & \multicolumn{3}{c|}{all anomalies} & \multicolumn{3}{c|}{animals} & \multicolumn{3}{c|}{vehicles} & \multicolumn{3}{c}{other anomalies} \\
    \midrule
    & OoD & \multicolumn{3}{c|}{$N=100$} & \multicolumn{3}{c|}{$N=59$} &  \multicolumn{3}{c|}{$N=23$} & \multicolumn{3}{c}{$N=11$} \\
    Method & data & \multicolumn{1}{c}{AuPRC $\uparrow$} & FPR$_{95}$ $\downarrow$ & \multicolumn{1}{c|}{$\overline{F_1}\uparrow$} & \multicolumn{1}{c}{AuPRC $\uparrow$} & FPR$_{95}$ $\downarrow$ & \multicolumn{1}{c|}{$\overline{F_1}\uparrow$} & \multicolumn{1}{c}{AuPRC $\uparrow$} & FPR$_{95}$ $\downarrow$ & \multicolumn{1}{c|}{$\overline{F_1}\uparrow$} & \multicolumn{1}{c}{AuPRC $\uparrow$} & FPR$_{95}$ $\downarrow$ & \multicolumn{1}{c}{$\overline{F_1}\uparrow$}  \\
    \midrule
    Maximum softmax \cite{Hendrycks2017msp} & \xmark & 28.0 & 72.0 & 5.4 & 25.2 & 75.6 & 4.9 & 30.2 & 71.9 & 5.0 & 26.1 & 57.7 & 9.3 \\
    ODIN \cite{liang18odin} & \xmark & 31.1 & 71.7 & 5.2 & 32.1 & 72.9 & 4.9 & 30.6 & 74.0 & 4.9 & 35.5 & 61.7 & 9.7 \\
    Mahalanobis \cite{Lee2018mahala} & \xmark & 20.0 & 87.0 & 2.7 & 21.3 & 87.4 & 2.5 & 16.7 & 87.5 & 1.8 & 34.9 & 66.1 & 12.5 \\
    MC dropout \cite{mukhoti2019bdl} & \xmark & 28.9 & 69.5 & 4.3 & 24.8 & 74.0 & 3.2 & 35.2 & 72.2 & 4.5 & 20.1 & 62.3 & 15.2\\
    Ensemble \cite{Lakshminarayanan17} & \xmark & 17.7 & 91.1 & 3.4 & 16.7 & 91.3 & 2.9 & 18.8 & 89.7 & 1.1 & 10.4 & 85.6 & 4.8 \\
\bottomrule
    \end{tabular}
    }
    \end{center}
    \caption{Effect of different anomalies in the RoadAnomaly21 dataset. In total, RoadAnomaly21 contains 59 images with only animals, 23 images with only vehicles and 11 with other anomalies (\eg cones, tents, \ldots). The number of images according to a subset is denoted with $N$ in the table. Images containing objects from both the animal and the vehicle category (7 images in total) are excluded in this evaluation.
    }
    \label{tab:table_categories}
\end{table*}

\begin{table*}[t]
    \setlength{\tabcolsep}{2pt}
    \begin{center}
    \scalebox{0.67}{
    \begin{tabular}{l||c||ag|ag|ag|ag|ag|ag|ag}
    \toprule
    & & \multicolumn{2}{c|}{road cracked} & \multicolumn{2}{c|}{asphalt dark} & \multicolumn{2}{c|}{road gravel} & \multicolumn{2}{c|}{asphalt gray} & \multicolumn{2}{c|}{motorway} & \multicolumn{2}{c|}{sun reflection} & \multicolumn{2}{c}{road bricks} \\
    \midrule
    & OoD & \multicolumn{2}{c|}{$N=40$} & \multicolumn{2}{c|}{$N=47$} & \multicolumn{2}{c|}{$N=33$} & \multicolumn{2}{c|}{$N=66$} & \multicolumn{2}{c|}{$N=30$} & \multicolumn{2}{c|}{$N=72$} & \multicolumn{2}{c}{$N=39$}  \\
    Method & data & \multicolumn{1}{c}{AuPRC} & \multicolumn{1}{c|}{$\overline{F_1}$} & \multicolumn{1}{c}{AuPRC} & \multicolumn{1}{c|}{$\overline{F_1}$} & \multicolumn{1}{c}{AuPRC} & \multicolumn{1}{c|}{$\overline{F_1}$} & \multicolumn{1}{c}{AuPRC} & \multicolumn{1}{c|}{$\overline{F_1}$} & \multicolumn{1}{c}{AuPRC} & \multicolumn{1}{c|}{$\overline{F_1}$} & \multicolumn{1}{c}{AuPRC} & \multicolumn{1}{c|}{$\overline{F_1}$} & \multicolumn{1}{c}{AuPRC} & \multicolumn{1}{c}{$\overline{F_1}$}   \\
    \midrule
    Maximum softmax \cite{Hendrycks2017msp} & \xmark &                    11.7 &                              3.2 &                             69.3 &                                      25.7 &                     39.5 &                              21.0 &                          43.4 &                                   14.9 &                        4.8 &                                 0.8 &                     2.1 &                              4.4 &                     32.7 &                              26.8 \\
\bottomrule
    \end{tabular}
    }
    \end{center}
    
    \begin{center}
    \scalebox{0.67}{
    \begin{tabular}{l||c||ag|ag}
    \toprule
    & & \multicolumn{2}{c|}{asphalt night} & \multicolumn{2}{c}{snowstorm} \\
    \midrule
    & OoD & \multicolumn{2}{c|}{$N=30$} & \multicolumn{2}{c}{$N=55$} \\
    Method & data & \multicolumn{1}{c}{AuPRC} & \multicolumn{1}{c|}{$\overline{F_1}$} & \multicolumn{1}{c}{AuPRC} & \multicolumn{1}{c}{$\overline{F_1}$} \\
    \midrule
    Maximum softmax \cite{Hendrycks2017msp} & \xmark & 6.0 &                              2.5 &                         1.6 &                                  0.8 \\
    ODIN \cite{liang18odin} & \xmark & 8.0 &                              1.8 &                         6.7 &                                  4.6 \\
    Mahalanobis \cite{Lee2018mahala} & \xmark &                    14.2 &                              5.5 &                        21.2 &                                 13.2 \\
    MC dropout \cite{mukhoti2019bdl} & \xmark &                     4.2 &                              1.1 &                         0.5 &                                  0.6 \\
    Ensemble \cite{Lakshminarayanan17} & \xmark & 11.5 & 16.9 & 0.6 & 0.0 \\
\bottomrule
    \end{tabular}
    }
    \end{center}
    
    \caption{Effect of different of scenes in the RoadObstacle21 dataset. Here, $N$ denotes the number of images in a subset. As main evaluation metrics we consider the pixel-wise AuPRC and the component-wise $\overline{F_1}$.
    }
    \label{tab:table_scenes}
\end{table*}

\vspace{-2cm}
\begin{figure*}[ht]
    \centering
    \captionsetup[subfigure]{labelformat=empty, position=top}
    \subfloat[Image \& annotation]{\includegraphics[width=0.199\textwidth]{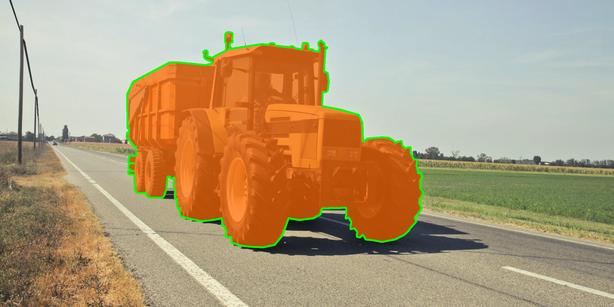}}
    \subfloat[Maximum softmax]{\includegraphics[width=0.199\textwidth]{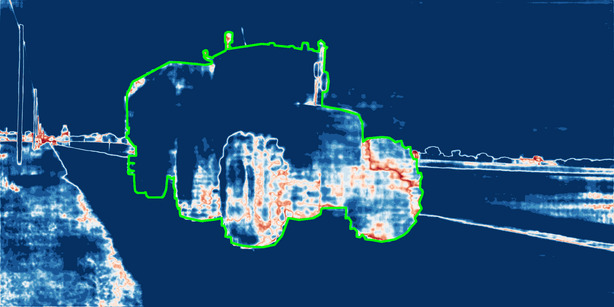}}
    \subfloat[ODIN]{\includegraphics[width=0.199\textwidth]{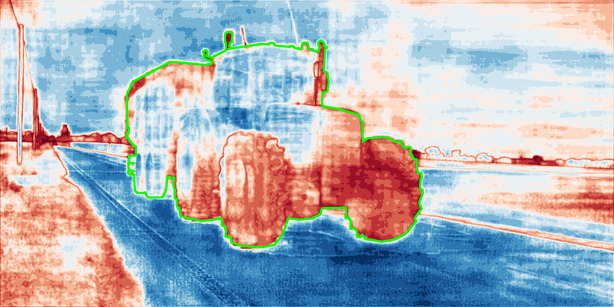}}
    \subfloat[Mahalanobis]{\includegraphics[width=0.199\textwidth]{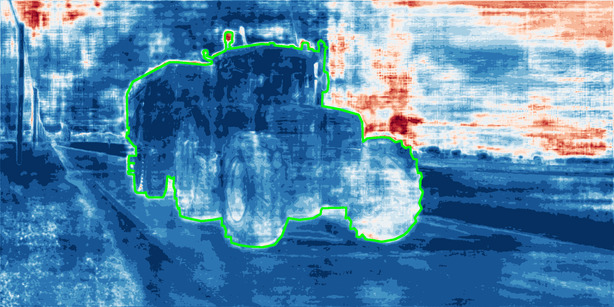}}
    \subfloat[MC dropout]{\includegraphics[width=0.199\textwidth]{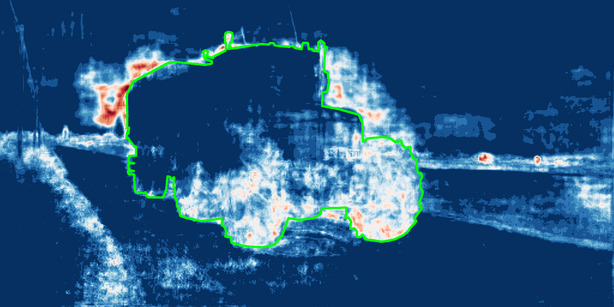}}\\
    \vspace{-1.07\baselineskip}
    \captionsetup[subfigure]{labelformat=empty, position=bottom}
    \subfloat[Void classifier]{\includegraphics[width=0.199\textwidth]{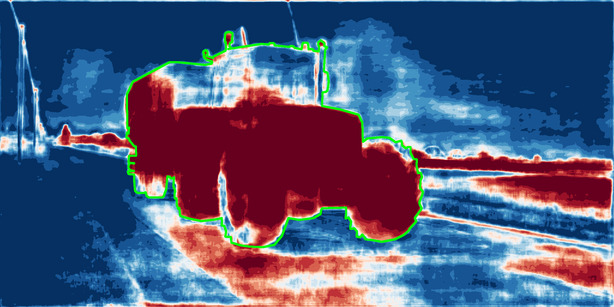}}
    \subfloat[Embedding density]{\includegraphics[width=0.199\textwidth]{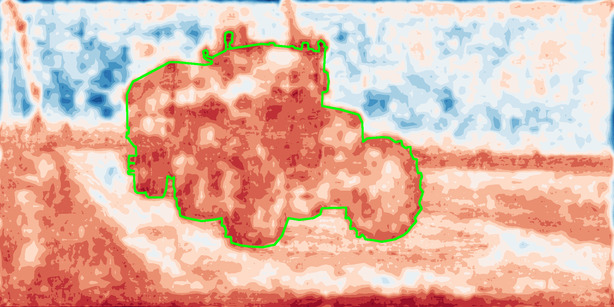}}
    \subfloat[Image resynthesis]{\includegraphics[width=0.199\textwidth]{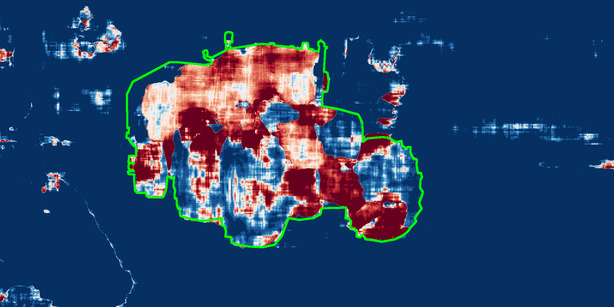}}
    \subfloat[SynBoost]{\includegraphics[width=0.199\textwidth]{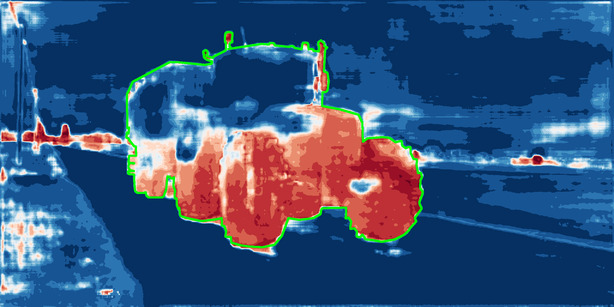}}
    \subfloat[Maximized entropy]{\includegraphics[width=0.199\textwidth]{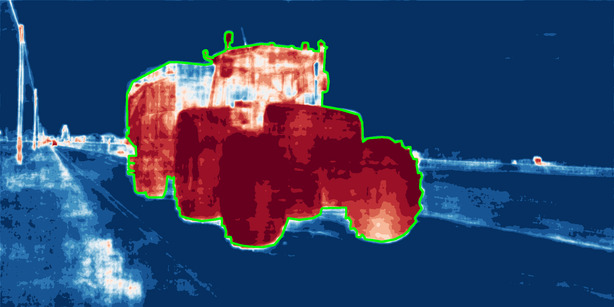}}
    \caption{Qualitative comparison of the methods introduced in \cref{sec:methods} and \cref{sec:methods-detail} on a sample from RoadAnomaly21. The scene shows a tractor which does not appear in Cityscapes. Green contours indicate the annotation of the anomaly.
    }
    \label{fig:example4}
\end{figure*}
\vspace{-2cm}
\vspace{-2cm}

\clearpage

\end{document}